\documentclass[a4paper,fleqn]{cas-sc}
\usepackage{bm}
\usepackage[ruled,vlined,linesnumbered]{algorithm2e}
\SetNlSty{textnormal}{}{}
\usepackage{booktabs}
\usepackage{multirow}
\usepackage{siunitx}
\usepackage{afterpage}
\usepackage{placeins}

\usepackage[authoryear,longnamesfirst,sort&compress,numbers]{natbib}

\def\tsc#1{\csdef{#1}{\textsc{\lowercase{#1}}\xspace}}
\tsc{WGM}
\tsc{QE}

\newtheorem{theorem}{Theorem}
\newtheorem{lemma}[theorem]{Lemma}
\newdefinition{rmk}{Remark}
\newproof{pf}{Proof}

\let\printorcid\relax

\let\oldnl\nl
\newcommand{\nlnonumber}{\renewcommand{\nl}{\let\nl\oldnl}}

\begin{document}
\let\WriteBookmarks\relax
   
\shortauthors{I. Corona Guevara, Y. Hu}  

\title [mode = title]{Structured Neural Chaos: An Adaptive Surrogate Modeling Framework for Functional Uncertainty Quantification and Global Sensitivity Analysis}  
\shorttitle{sNC: An Adaptive Surrogate Modeling Framework for Functional UQ and GSA}
%

\author[1]{Isabel {Corona Guevara}}

\cormark[1]

\ead{isabel.coronaguevara@ucdenver.edu}

\affiliation[1]{organization={Department of Mathematical and Statistical Sciences, University of Colorado Denver},
            city={Denver},
            postcode={80204}, 
            state={CO},
            country={USA}}

\author[2]{Yeping Hu}

\ead{hu25@llnl.gov}

\affiliation[2]{organization={Computational Engineering Division, Lawrence Livermore National Laboratory},
            city={Livermore},
            postcode={94550}, 
            state={CA},
            country={USA}}

\cortext[1]{Corresponding author}



\begin{abstract}
Variance-based global sensitivity analysis (GSA) plays an important role in uncertainty quantification by identifying the contributions of uncertain inputs to the variability of the model response. The repeated model evaluations required for these tasks are often prohibitively expensive; surrogate models provide an efficient alternative by constructing inexpensive approximations of the underlying system response. Constructing surrogate models that combine scalability and interpretability for systems with high-dimensional stochastic inputs and functional responses remains challenging, particularly when sensitivity estimates are required across spatial or temporal domains. Polynomial chaos expansion (PCE) provides an effective framework for uncertainty propagation and sensitivity analysis due to its orthogonal structure and direct relationship with variance-based sensitivity measures. However, PCE suffers from the curse of dimensionality, and the resulting computational burden becomes more pronounced for problems with functional responses. In this work, we introduce the Structured Neural Chaos (sNC) expansion as a surrogate modeling framework for variance-based GSA, inspired by the interpretability and orthogonal structure of PCE. The proposed framework is designed to retain the interpretability of structured decompositions while leveraging the expressive power of neural networks. The sNC expansion mirrors a truncated functional analysis of variance (ANOVA) decomposition, where each interaction component admits a separable low-rank approximation whose basis functions and coefficients are parameterized using neural networks. The expansion is constructed sequentially, adaptively identifying the dominant modes within each ANOVA subspace and determining the effective complexity of the representation. The resulting structure enables the extraction of statistical and sensitivity quantities directly from the coefficients of the sNC expansion at negligible cost. Numerical results demonstrate that the proposed framework achieves accurate surrogate predictions and highly accurate estimates of statistical and sensitivity quantities while substantially reducing computational cost relative to Monte Carlo simulation.
\end{abstract}


\maketitle

\section{Introduction}

Mathematical modeling plays a central role in modern engineering and applied sciences, particularly in the analysis of complex systems under uncertainty. Simulations of these systems provide insight into how different parameters influence complex physical phenomena, often revealing behaviors that are otherwise difficult to observe directly. Uncertainty is inherent in all modeling and simulation processes, whether arising from incomplete knowledge of system parameters, limitations in computational models, or intrinsic variability in the underlying physical processes~\cite{sullivan2015introduction, smith2024uncertainty}. Quantifying and understanding the impact of these uncertainties is essential for reliable prediction and decision-making.

Sensitivity analysis provides a systematic framework for quantifying how uncertainty in the model inputs influences the variability of the model response and is commonly categorized into local and global approaches. Local sensitivity analysis examines the effect of small perturbations around a nominal parameter configuration but may fail to capture nonlinear interactions or parameter importance across the full input space. In contrast, global sensitivity analysis (GSA) quantifies the influence of uncertain inputs by integrating over their full probability distributions, making it particularly well suited for high-dimensional stochastic systems. Several approaches to GSA have been developed in the literature, among which variance-based methods are especially attractive because they decompose the output variance into contributions associated with individual variables and their interactions~\cite{saltelli2008global}.

Variance-based GSA was originally introduced through the work of Cukier et al.~\cite{cukier1973study} in the early 1970s and later generalized by Sobol' through the functional ANOVA decomposition and the associated Sobol' sensitivity indices~\cite{sobol1993sensitivity}. Subsequently, several Monte Carlo-based estimation procedures were developed to compute these indices directly from model evaluations~\cite{homma1996importance, saltelli2002making}. Despite their broad applicability, these approaches can become computationally expensive when applied to high-dimensional stochastic systems, where each high-fidelity simulation may require substantial computational resources. The computational burden becomes even more pronounced in the context of functional uncertainty quantification (UQ) and sensitivity analysis, where the quantities of interest are not scalar-values but instead depend on spatial and temporal variables. In such settings, statistical and sensitivity quantities become functional quantities defined over the underlying physical domain. 

Surrogate modeling techniques provide an efficient alternative to repeated high-fidelity simulations by constructing inexpensive approximations of the underlying system response. As a result, surrogate models have become an important tool for reducing the computational cost associated with UQ and sensitivity analysis. A wide variety of surrogate modeling approaches have been developed, including Gaussian Process (GP) models, also referred to as kriging models, which provide both predictions and uncertainty estimates~\cite{rasmussen2006gaussian}; Radial Basis Function (RBF) models, which offer flexible interpolation and regression capabilities~\cite{buhmann2000radial}; Polynomial Chaos Expansion (PCE) models, which provide interpretable spectral representations through orthogonal polynomial bases~\cite{wiener1938}; and Neural Networks (NNs), which are highly expressive and capable of approximating complex high-dimensional systems~\cite{goodfellow2016deep}. Despite the success of these approaches, accurately approximating high-dimensional functional quantities while preserving interpretability and enabling efficient sensitivity analysis remains a significant challenge.

PCE is a widely used framework for uncertainty quantification due to its spectral representation of stochastic systems, which enables UQ in high-dimensional settings. In particular, for GSA, the orthogonality of the polynomial basis functions enables the direct computation of statistical moments and Sobol' sensitivity indices from the expansion coefficients~\cite{sudret2008global, novak2025fractional}. Furthermore, the tensor-product structure of the polynomial basis naturally yields an interpretable decomposition of the stochastic response into contributions associated with individual variables and their interactions. PCE has been employed as a machine learning method and combined with existing concepts such as physics-informed learning \cite{sharma2024physics}, Reduced Order Models (ROMs) \cite{sun2021non}, and manifold learning \cite{kontolati2022manifold}. Several approaches aim to improve the efficiency and scalability of functional PCE approximations for stochastic systems. For example, Sharma et al.~\cite{sharma2026polynomial} recently introduced a physics-constrained polynomial chaos framework that incorporates governing physical constraints directly into the stochastic surrogate construction process, improving robustness and accuracy in scientific machine learning and uncertainty quantification applications.

Despite its advantages, PCE suffers from the curse of dimensionality, where the rapid growth in the number of polynomial basis terms required for accurate approximation limits scalability in high-dimensional settings. These challenges become even more pronounced in functional settings, where the expansion coefficients themselves become varying functions. Consequently, the construction of functional PCE representations in high-dimensional stochastic systems can become computationally expensive and may lead to ill-conditioned regression problems. The accurate and computationally efficient approximation of high-dimensional functional stochastic systems remains a significant open challenge, particularly when interpretable sensitivity analysis and scalable surrogate representations are simultaneously desired.

Surrogate modeling approaches based on NNs have received significant attention due to their strong approximation capabilities in high-dimensional and nonlinear settings~\cite{tripathy2018, meng2020ppinn, lu2021, kontolati2022survey}. NNs have been successfully applied to uncertainty quantification problems because of their ability to learn complex functional relationships directly from data while avoiding many of the restrictions associated with fixed polynomial bases. For example, Bahmani et al.~\cite{bahmani2025neural} recently introduced a neural chaos method inspired by PCE, where the orthogonal polynomial basis functions are replaced by NNs. In particular, NN parameterizations provide substantial flexibility for representing localized features and strongly nonlinear behavior that may be difficult to capture accurately using classical spectral methods. 

Despite these advantages, many NN-based surrogate approaches do not intrinsically provide the interpretable variance decompositions required for efficient global sensitivity analysis. Consequently, statistical and sensitivity quantities are often computed through additional post-processing procedures such as Monte Carlo simulation (MCS), potentially introducing additional computational cost and approximation error. Furthermore, many existing NN surrogate approaches do not explicitly preserve interpretable structures or orthogonality properties that are fundamental to variance-based sensitivity analysis. These observations motivate the development of surrogate modeling approaches that combine the expressive approximation capabilities of NNs with structured decompositions suitable for uncertainty quantification and global sensitivity analysis.

In this work, we propose a Structured Neural Chaos (sNC) framework that combines functional ANOVA decompositions with low-rank separable representations and NN parameterizations to construct efficient surrogate models for high-dimensional functional stochastic systems. The proposed approach preserves the interpretable variance decomposition structure associated with classical PCE methods while additionally leveraging the expressive approximation capabilities of NNs to improve the approximation of localized and highly nonlinear functional quantities. The resulting framework enables the direct computation of functional statistical and sensitivity quantities from the learned sNC expansion coefficients. Furthermore, the proposed framework is entirely non-intrusive and data-driven, requiring only evaluations of the underlying model without modifications to the governing equations. By combining low-rank approximations, the functional analysis of variance (ANOVA) structure, and sparse representations, the proposed method aims to mitigate the curse of dimensionality while maintaining interpretability and computational efficiency.

The remainder of this paper is structured as follows. Section~\ref{sec:background} reviews the mathematical framework underlying the proposed method, including the functional ANOVA decomposition, variance-based GSA, and PCE. Section~\ref{sec:sNC_formulation} presents the sNC formulation, including the direct extraction of statistical and sensitivity indices directly from the learned NNs. Section~\ref{sec:Numerical_examples} demonstrates the performance of the proposed framework for surrogate modeling and GSA through five numerical examples, including a comparison study with a polynomial chaos-based approach. Finally, Section~\ref{sec:conclusion} concludes the study by summarizing the main findings and contributions of this work.
\section{Mathematical background}
\label{sec:background}

This section introduces the mathematical framework underlying the proposed method, including the functional ANOVA decomposition, variance-based sensitivity analysis, and PCE. Consider a general model response $u(\bm{\xi})$, where $\bm{\xi}=[\xi_1,\dots, \xi_d]$ is a $d$-dimensional vector of independent random variables defined on a probability space $(\Omega, \mu)$. Assume that $u(\bm{\xi})\in L^2(\Omega,\mu)$, the space of square-integrable functions. The space $L^2(\Omega, \mu)$ forms a Hilbert space~\cite{cameron1947} when equipped with the inner product
\begin{align}
\langle f,g \rangle = \int f(\bm{\xi})g(\bm{\xi})\, d\mu(\bm{\xi})=\mathbb{E}[f(\bm{\xi})g(\bm{\xi})], && f,g\in L^2.
\label{eq:inner_product_generic}
\end{align}


\subsection{Functional ANOVA decomposition}
\label{subsec:ANOVA}

The functional ANOVA decomposition plays a central role in the representation of square-integrable stochastic responses. Let $D=\{1,\dots, d\}$, and for any subset $\alpha\subseteq D$, define the subspace
\begin{align}
    \mathcal{H}_\alpha=\left\{u_\alpha(\bm{\xi}_\alpha)\in L^2(\Omega, \mu) \mid \mathbb{E}[u_\alpha(\bm{\xi}_\alpha)\mid \bm{\xi}_{\alpha\setminus \{i\} } ]=0,\quad \forall i\in \alpha \right\},
    \label{eq:H_alpha_set}
\end{align}
where the space associated with the empty set is defined as 
\begin{align}
    \mathcal{H}_\emptyset = \{\text{constants}\}.
\end{align}
Under the assumption that the random variables $\{\xi_i\}_{i=1}^d$ are independent, the subspaces $\{\mathcal{H}_\alpha\}_{\alpha\subseteq D}$ are mutually orthogonal and satisfy the orthogonal direct-sum decomposition
\begin{align}
    L^2(\Omega,\mu) = \bigoplus_{\alpha\subseteq D}\mathcal{H}_\alpha.
\end{align}
Consequently, every function $u\in L^2(\Omega,\mu)$ admits the unique decomposition
\begin{align}
    u(\bm{\xi}) = \sum_{\alpha\subseteq D}u_\alpha(\bm{\xi}_\alpha), \quad u_\alpha\in\mathcal{H}_\alpha,
    \label{eq:ANOVA_u}
\end{align}
commonly referred to as the functional ANOVA decomposition.

Since $L^2(\Omega, \mu)$ is a Hilbert space and the subspaces $\mathcal{H}_\alpha$ are closed, every $u\in L^2(\Omega,\mu)$ admits a unique orthogonal projection onto $\mathcal{H}_\alpha$. In particular, the orthogonal projection onto $\mathcal{H}_\alpha$ coincides with the associated ANOVA component,
\begin{align}
    P_{\mathcal{H}_\alpha}u = u_\alpha.
\end{align}
For example, the orthogonal projections onto $\mathcal{H}_{\{i\}}$ and $\mathcal{H}_{\{i,j\}}$ are given by
\begin{align}
    P_{\mathcal{H}_{\{i\}}}u=\mathbb{E}[u(\bm{\xi})\mid\xi_i] -\mathbb{E}[u(\bm{\xi})],
    \label{eq:projection_H_i}
\end{align}
and
\begin{align}
    P_{\mathcal{H}_{\{i,j\}}}u = \mathbb{E}[u(\bm{\xi})\mid \xi_i, \xi_j] - \mathbb{E}[u(\bm{\xi})\mid \xi_i] - \mathbb{E}[u(\bm{\xi})\mid \xi_j] + \mathbb{E}[u(\bm{\xi})],
    \label{eq:projection_H_ij}
\end{align}
respectively. The quantity $P_{\mathcal{H}_\alpha}u$ represents the contribution of the interaction subset $\alpha$ to the total stochastic response. 


\subsection{Variance-based global sensitivity analysis}
\label{subsec:GSA}
In many applications, we are interested in quantifying the overall uncertainty in a model response as well as identifying which uncertain inputs contribute most significantly to that uncertainty. Global sensitivity analysis addresses this by allocating the variability of the model response to its sources in the input space. Sobol'~\cite{sobol2001} introduced two widely used sensitivity indices: the first-order and total-order Sobol' indices. The first-order index quantifies the contribution of an individual input variable to the output variance, while the total-order index measures the total contribution of an input, including all interaction effects involving that variable. 

The Sobol' sensitivity indices arise naturally from the functional ANOVA decomposition of the stochastic response. In particular, the orthogonality of the ANOVA components implies the variance decomposition 
\begin{align}
        \text{Var}[u(\bm{\xi})] = \sum_{\substack{\alpha\subseteq D \\ \alpha\neq\emptyset}} \text{Var}[u_\alpha],
        \label{eq:ANOVA_var}
\end{align}
where $u_\alpha$ denotes the ANOVA component associated with the interaction subset $\alpha$. This decomposition enables the contribution of individual variables and their interactions to be quantified through variance-based sensitivity indices. The first-order Sobol' indices are defined as
\begin{align}
    S_i = \frac{\text{Var}[u_{\{i\}}]}{\text{Var}[u]},\quad i=1,\dots,d
    \label{eq:ANOVA_S_i}
\end{align}
and the total-order Sobol' indices are then defined as
\begin{align}
    S_{T_i}=\frac{V_{T_i}}{\text{Var}[u]},\quad i=1,\dots,d,
    \label{eq:ANOVA_S_Ti}
\end{align}
where
\begin{align}
    V_{T_i}=\sum_{i\in\alpha\subseteq D }\text{Var}[u_\alpha].
\end{align}

Direct computation of these indices is often challenging due to the high-dimensional integrals involved. Several Monte Carlo approaches have been proposed for their estimation. For example, Saltelli et al.~\cite{saltelli2010} introduced a widely adopted variance-based estimation procedure, and randomized quasi-Monte Carlo methods have since been developed to improve computational efficiency~\cite{liu2015global, liu2016accurate, okten2021randomized}. However, these approaches can become computationally prohibitive when evaluations of the underlying model are expensive. To alleviate the computational burden associated with repeated model evaluations, surrogate modeling techniques are often employed for efficient sensitivity analysis.


\subsection{Polynomial chaos expansion}
\label{subsec:PCE}
PCE is a spectral surrogate modeling approach that represents stochastic responses using orthogonal polynomial basis functions. It has been widely applied in uncertainty quantification, particularly for sensitivity analysis, reliability assessment, and surrogate modeling of complex engineering systems.

Let 
\begin{align}
    \mathcal{B}=\{\Psi_{\bm{\beta}}\mid \bm{\beta}\in\mathbb{N}^d\}
\end{align}
be a complete orthonormal basis for $L^2(\Omega,\mu)$, consisting of multivariate polynomials in $d$ variables that are orthonormal with respect to the inner product defined in \eqref{eq:inner_product_generic}. That is,
\begin{align}
\langle \Psi_{\bm{\beta}},\Psi_{\bm{\gamma}} \rangle = \int \Psi_{\bm{\beta}}(\bm{\xi})\Psi_{\bm{\gamma}}(\bm{\xi})\, d\mu(\bm{\xi}) =  \delta_{\bm{\beta\gamma}},\quad \bm{\beta},\bm{\gamma}\in\mathbb{N}^d ,\label{eq:inner_product}
\end{align}
where $\delta_{\bm{\beta\gamma}}$ is the Kronecker delta. Since the random variables $\xi_1,\dots,\xi_d$ are independent, the multivariate basis functions can be constructed through tensor products of univariate orthonormal polynomial bases:
\begin{align}
\Psi_{\bm{\beta}}(\bm{\xi}) = \prod_{i=1}^d \psi_{\beta_i}(\xi_i).\label{eq:prod_univariate}
\end{align}
Here, 
\begin{align}
    \{\psi_{\beta_i}\mid \beta_i\in\mathbb{N}\},\quad i=1,\dots,d
\end{align}
denotes a family of univariate orthonormal polynomials of degree $\beta_i$. In classical PCE, these polynomials are selected from the Wiener-Askey polynomial families~\cite{askey1985} so that they are orthogonal with respect to the probability distributions of the input variables.

Once the basis is defined, then $u\in L^2(\Omega,\mu)$ admits the expansion
\begin{align}
u(\bm{\xi}) = \sum_{\bm{\beta}\in\mathbb{N}^d}a_{\bm{\beta}}\Psi_{\bm{\beta}}(\bm{\xi}) \label{eq:PCE_infinite}
\end{align} 
where $a_{\bm{\beta}}$ are the expansion coefficients. Since the basis functions are orthonormal, the coefficients are formally given by
\begin{align}
a_{\bm{\beta}} = \langle u, \Psi_{\bm{\beta}} \rangle.
\end{align}
In principle, the coefficients can be computed exactly by this projection integral, but it is not feasible in cases where the model response $Y(\bm{\xi})$ is computationally expensive to run. As a result, numerical methods such as regression, quadrature, or sparse approximation are employed to efficiently estimate the coefficients.

In practice, the expansion is truncated at a maximum total polynomial degree $p\in\mathbb{N}$, yielding
\begin{align}
u(\bm{\xi}) = \sum_{i=0}^{n-1} a_{i}\Psi_i(\bm{\xi}) + \epsilon_p, \label{eq:PCE}
\end{align}
where $a_i$ denotes the coefficients associated with the $i$-th multi-index up to total degree $p$ and $\epsilon_p$ denotes the truncation error. The coefficients are relabeled using a single index $i$ purely for notational convenience. The number of terms in the expansion is given by
\begin{align}
n = \frac{(d+p)!}{d!p!}. \label{def:n}
\end{align}
While PCE provides an efficient and computationally inexpensive means of evaluating complex systems, it suffers from the curse of dimensionality: the number of terms in the expansion grows rapidly as the number of parameters increases, thereby increasing the computational burden.

\subsubsection{PCE-based global sensitivity analysis}
By exploiting the orthonormality of the polynomial basis, statistical quantities and Sobol' sensitivity indices can be obtained directly from the PCE coefficients, providing an efficient alternative to Monte Carlo-based methods. Consider the truncated expansion in Eq.~\eqref{eq:PCE}, and let 
\begin{align}
    \bm{\beta}^{(j)}=(\beta_1^{(j)}, \dots, \beta_d^{(j)})
\end{align}
denote the multi-index associated with the basis function $\Psi_j(\bm{\xi})$. Since the basis functions are orthonormal, the mean and variance of the stochastic response are approximated by
\begin{align}
&\mathbb{E}[u] = a_0,
\end{align}
and
\begin{align}
&\text{Var}[u] = \sum_{j=1}^{n-1} a_j^2,
\end{align}
respectively. 

The partial contribution of the $i$-th variable to the variance is given by
\begin{align}
\text{Var}[u_{\{i\}}] = \sum_{j\in\mathcal{A}_i} a^2_{j},
\end{align}
where 
\begin{align}
    \mathcal{A}_i=\{j\in\{0,\dots, n-1\}\mid \beta_i^{(j)}\not= 0 \text{ and } \beta_r^{(j)} = 0 \text{ for } r\not= i\}. \label{eq:A_i}
\end{align}
That is, the sum includes coefficients corresponding to basis functions that depend exclusively on $\xi_i$. Similarly, the total contribution of the $i$-th variable is given by
\begin{align}
V_{T_i} = \sum_{j\in\mathcal{A}_{T_i}} a_j^2,
\end{align}
where 
\begin{align}
    \mathcal{A}_{T_i} = \{j\in\{0,\dots,n-1\}\mid \beta_i^{(j)}\not= 0\}.
\end{align}

Consequently, the Sobol' sensitivity indices can be computed directly from the PCE coefficients without requiring additional Monte Carlo integration. This property is one of the primary advantages of PCE-based sensitivity analysis and partially motivates the proposed framework, which seeks to retain similar interpretable variance decompositions while replacing fixed polynomial bases with adaptive neural representations.
\section{Structured Neural Chaos formulation}
\label{sec:sNC_formulation}

\begin{figure}
  \center
  \includegraphics[width=5.6in]{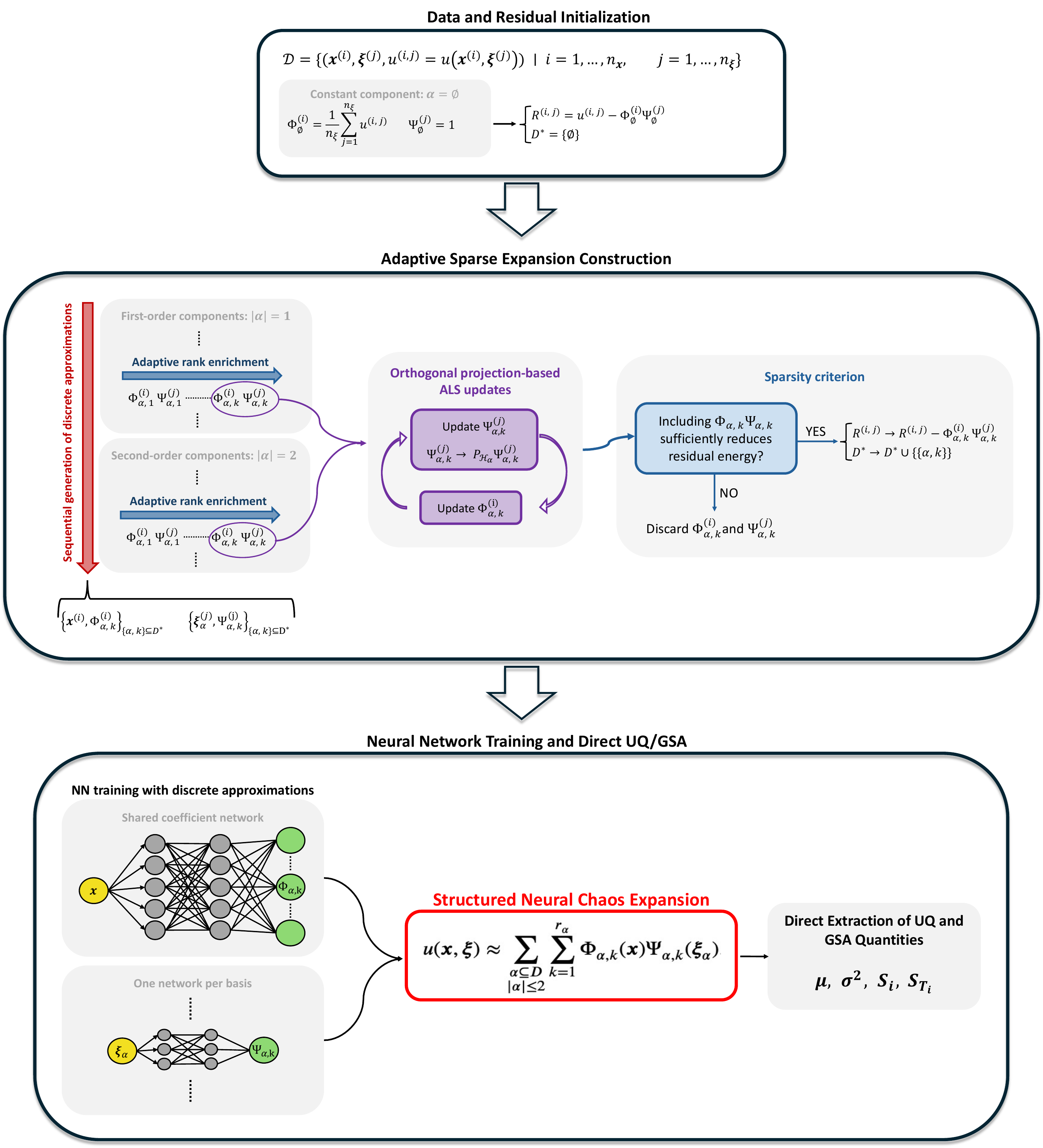}\\
  \vspace{0.2em}
  \caption{Schematic of the sNC framework. Discrete approximations of the coefficient and basis functions are first constructed through residual-based alternating least squares (ALS) together with orthogonal projection onto the corresponding ANOVA subspaces. Adaptive rank enrichment and sparsity are incorporated through a sparsity criterion that determines whether the inclusion of a candidate term yields sufficient reduction in the residual energy. The resulting discrete coefficient and basis data are used to train NN approximations which define the terms of the sNC expansion, providing a continuous functional representation suitable for uncertainty quantification and global sensitivity analysis.}
  \label{fig:sNC_schematic}
\end{figure}

The proposed framework combines a functional ANOVA decomposition with low-rank separable representations and NN parameterizations to construct efficient and interpretable surrogate models for high-dimensional functional stochastic systems. The proposed formulation is termed Structured Neural Chaos (sNC), where the terminology is inspired by the Neural Chaos formulation introduced in~\cite{bahmani2025neural}, which itself was named in homage to PCE. In contrast to the original Neural Chaos framework, the proposed method fundamentally differs through the incorporation of a structured ANOVA decomposition that enables interpretable low-rank interaction representations together with the direct computation of Sobol' sensitivity indices. The resulting framework combines the flexibility of NN parameterizations with ANOVA-based variance decomposition for functional uncertainty quantification and global sensitivity analysis. A schematic representation of the proposed method is shown in Fig.~\ref{fig:sNC_schematic}.

Consider a model response $u(\bm{x},\bm{\xi})$, where $\bm{x}$ denotes deterministic variables, which can include spatial and temporal coordinates, and $\bm{\xi}$ is a $d$-dimensional vector of independent random variables defined on a probability space $(\Omega, \mu)$. Assume that, for each fixed $\bm{x}$, the stochastic response $u(\bm{x},\cdot)$ belongs to $L^2(\Omega,\mu)$, allowing the concepts introduced in the previous section to be extended pointwise over the deterministic domain. Consequently, the response admits the unique ANOVA decomposition given in Eq.~\eqref{eq:ANOVA_u}. In the present work, this decomposition is truncated after second-order interaction terms:
\begin{align}
    u(\bm{x},\bm{\xi}) \approx \sum_{\substack{\alpha\subseteq D\\|\alpha|\leq 2}} u_\alpha(\bm{x},\bm{\xi}_\alpha) , \quad u_\alpha\in \mathcal{H}_\alpha,
\end{align}
or equivalently, 
\begin{align}
    u(\bm{x},\bm{\xi}) \approx u_\emptyset(\bm{x}) + \sum_{i=1}^d u_{\{i\}}(\bm{x}, \xi_i) + \sum_{1\leq i<j\leq d} u_{\{i,j\}}(\bm{x}, \xi_i,\xi_j).
\end{align}
This truncation is motivated by the observation that, in many practical systems, most variance is captured by low-order interactions, while higher-order Sobol terms often contribute negligibly. Moreover, restricting the decomposition to pairwise interactions substantially reduces the combinatorial complexity of the approximation. In particular, the number of terms in the truncated decomposition is reduced from $2^d$ to
\begin{align}
    n = 1 + d + \binom{d}{2} = 1 + \frac{d(d+1)}{2}.
    \label{eq:n}
\end{align}

To this end, we further assume that each ANOVA component $u_\alpha$, with $\alpha\not=\emptyset$, admits a low-rank separable approximation of the form
\begin{align}
    u_\alpha(\bm{x},\bm{\xi}_\alpha) \approx \sum_{k=1}^{r_\alpha}\Phi_{\alpha,k}(\bm{x})\Psi_{\alpha,k}(\bm{\xi}_\alpha).
\end{align}
In other words, within each Sobol space $\mathcal{H}_\alpha$, the approximation is restricted to a low-dimensional manifold of separable functions. Since the constant ANOVA component 
\begin{align}
    u_\emptyset(\bm{x})=\mathbb{E}_{\bm{\xi}}[u(\bm{x},\bm{\xi})]
\end{align}
is independent of the stochastic variables, it already admits the trivial separable representation 
\begin{align}
    u_\emptyset(\bm{x})=\Phi_\emptyset(\bm{x})\Psi_\emptyset(\bm{\xi}),\quad \Psi_\emptyset(\bm{\xi})=1.
\end{align}
Consequently, no additional rank enrichment is required for the constant component, and a rank-1 representation is sufficient. The resulting truncated sNC expansion takes the form
\begin{align}
    u(\bm{x},\bm{\xi}) \approx \Phi_\emptyset(\bm{x}) + \sum_{\substack{\alpha\subseteq D\\1\leq|\alpha|\leq 2}} \sum_{k=1}^{r_\alpha}\Phi_{\alpha,k}(\bm{x})\Psi_{\alpha,k}(\bm{\xi}_\alpha).
    \label{eq:sNC_truncated}
\end{align}
Following the terminology commonly used in the PCE literature, $\Phi_{\alpha,k}(\bm{x})$ are referred to as the coefficient functions of the expansion, while $\Psi_{\alpha,k}(\bm{\xi}_\alpha)$ are referred to as the basis functions of the expansion. This terminology will be adopted throughout the remainder of the paper. The total number of terms in the expansion is
\begin{align}
    n^*=1+\sum_{\substack{\alpha\subseteq D\\ 1\leq|\alpha|\leq 2}} r_\alpha
\end{align}
In particular, if each non-constant ANOVA component is approximated using a rank-$r$ representation, the total number of terms becomes
\begin{align}
    1+(n-1)r.
\end{align}
As will be shown, the proposed approach further reduces this number through adaptive rank truncation, thereby generating sparse low-rank representations. 

The subspaces $\{\mathcal{H}_\alpha\}_{\alpha\subseteq D}$ associated with the ANOVA decomposition are mutually orthogonal, and therefore the corresponding ANOVA components $u_\alpha$ are orthogonal to one another. In addition, we impose the requirement that, within each ANOVA component, the basis functions $\Psi_{\alpha,k}(\bm{\xi}_\alpha)$ are orthogonal with respect to the probability measure of $\bm{\xi}_\alpha$. The goal now is to parameterize the coefficient functions $\Phi_{\alpha,k}(\bm{x})$ and basis functions $\Psi_{\alpha,k}(\bm{\xi}_\alpha)$ using NNs while preserving the orthogonality structure in a fully data-driven manner. Furthermore, the basis functions will be normalized such that the resulting basis functions are orthonormal. These orthonormality conditions enable the direct extraction of statistical quantities and Sobol' sensitivity indices from the coefficient functions.

The proposed framework adopts a residual-based construction strategy, where the expansion is enriched sequentially through residual minimization together with orthogonal projection onto the corresponding ANOVA subspaces. The orthogonal projection enforces orthogonality between distinct ANOVA components $u_\alpha$, while the sequential residual minimization promotes orthogonality among the basis functions $\Psi_{\alpha,k}(\bm{\xi}_\alpha)$ associated with the same ANOVA component. In addition, minimizing the residual adaptively captures the dominant modes contributing to the remaining approximation error. The present work utilizes the residual-based approach introduced in \cite{bahmani2025neural} for stochastic processes, summarized in the following Lemma.

\begin{lemma}
\label{lemma:theory}
    Consider the model response $u(\bm{x},\bm{\xi})\in L^2(\Omega,\mu)$, where $\bm{x}$ denotes deterministic variables and $\bm{\xi}$ is a random vector. Let
    \begin{align}
        u^{(p)}(\bm{x}, \bm{\xi}) = \sum_{i=0}^p\Phi_i(\bm{x})\Psi_i(\bm{\xi}).
    \end{align}
    denote a separable stochastic approximation of $u$, and define the truncated residual
    \begin{align}
        R^{(p)}(\bm{x}, \bm{\xi})=u(\bm{x}, \bm{\xi})-u^{(p)}(\bm{x}, \bm{\xi}).
    \end{align}
    Assume that the basis functions $\{\Psi_{i}\}_{i=0}^p$ are orthonormal in $L^2(\Omega,\mu)$.
    \begin{enumerate}
        \item[(a)] The truncated residual $R^{(p)}(\bm{x}, \bm{\xi})$ is orthogonal to each basis function $\Psi_i$ for $i=0,\dots,p$, i.e., 
        \begin{align}
            \mathbb{E}\left[\Psi_i(\bm{\xi})R^{(p)}(\bm{x}, \bm{\xi})\right]=0, \quad i=0,\dots,p.
        \end{align}
        \item[(b)] Let $\Psi(\bm{\xi})$ and $\Phi(\bm{x})$ be nonzero functions. If
        \begin{align}
            \| R^{(p)}(\bm{x}, \bm{\xi})-\Phi(\bm{x})\Psi(\bm{\xi})\|^2 < \| R^{(p)}(\bm{x}, \bm{\xi})\|^2
        \end{align}
        then 
        \begin{align}
            \Psi(\bm{\xi})\notin \mathrm{span}\left\{\Psi_i(\bm{\xi})\right\}_{i=0}^p.
        \end{align}
    \end{enumerate}
\end{lemma}
\begin{lemma}
\label{lemma:ALS}
    Let $R(\bm{x},\bm{\xi})\in L^2(\Omega,\mu)$ denote a residual function. Consider the separable approximation problem
    \begin{align}
        \underset{\Phi, \Psi}{\min}\,
        \|R(\bm{x}, \bm{\xi})-\Phi(\bm{x})\Psi(\bm{\xi})\|^2.
    \end{align}
    Then, in an alternating minimization procedure, each factor admits a closed-form update when the other factor is fixed as follows.
    \begin{enumerate}
        \item[(a)] For fixed nonzero $\Psi(\bm{\xi})$, the optimal $\Phi(\bm{x})$ has the following closed-form solution:
        \begin{align}
            \Phi^*(\bm{x})
            =
            \underset{\Phi}{\arg\min}\,
            \mathbb{E}_{\bm{\xi}}
            \left[
            \left(\Phi(\bm{x})\Psi(\bm{\xi})-R(\bm{x},\bm{\xi})\right)^2
            \right]
            =
            \frac{
            \mathbb{E}_{\bm{\xi}}[\Psi(\bm{\xi})R(\bm{x}, \bm{\xi})]
            }{
            \mathbb{E}_{\bm{\xi}}[\Psi(\bm{\xi})^2]
            }.
            \label{eq:Phi_update}
        \end{align}

        \item[(b)] For fixed nonzero $\Phi(\bm{x})$, the optimal $\Psi(\bm{\xi})$ has the following closed-form solution:
        \begin{align}
            \Psi^*(\bm{\xi})
            =
            \underset{\Psi}{\arg\min}\,
            \mathbb{E}_{\bm{x}}
            \left[
            \left(\Phi(\bm{x})\Psi(\bm{\xi})-R(\bm{x},\bm{\xi})\right)^2
            \right]
            =
            \frac{
            \mathbb{E}_{\bm{x}}[\Phi(\bm{x})R(\bm{x}, \bm{\xi})]
            }{
            \mathbb{E}_{\bm{x}}[\Phi(\bm{x})^2]
            }.
            \label{eq:Psi_update}
        \end{align}
    \end{enumerate}
\end{lemma}
Consequently, sequential residual minimization adaptively enriches the basis functions with modes associated with the dominant remaining approximation error while promoting orthogonality with previously constructed basis functions.

\begin{algorithm}
\caption{Construction of the Structured Neural Chaos Expansion}\label{alg:sNC}
\SetKwInOut{Input}{Input}
\SetKwInOut{Output}{Output}
\SetKwInOut{Initialize}{Initialize}

\Input{$\mathcal{D}=\left\{\left(\bm{x}^{(i)}, \bm{\xi}^{(j)}, u^{(i,j)}=u(\bm{x}^{(i)}, \bm{\xi}^{(j)})\right)\mid i=1,\dots, n_{\bm{x}}, \quad j=1,\dots,n_{\bm{\xi}}\right\}$ (data), \\
$\delta$ (convergence tolerance),\\
$\tau$ (sparsity parameter), \\
$r_{\max}$ (maximum allowable rank for each interaction component) \\
\phantom{x}}

$\Psi^{(j)}_\emptyset=1$ \\
$\Phi_\emptyset^{(i)} = \frac{1}{n_{\bm{\xi}}}\sum_{j=1}^{n_{\bm{\xi}}}u^{(i,j)}$ \\
$R^{(i,j)}=u^{(i,j)}-\Phi^{(i)}_\emptyset\Psi^{(j)}_\emptyset$\\
\BlankLine \BlankLine
\nlnonumber
\textbf{Orthogonal Projection-Based ALS Iterations} \\
\For{$\alpha\subseteq D,\ \alpha\not=\emptyset$}{
    \For {$k=1,2,\dots, r_{\max}$}{
        Initialize $\Psi_{\alpha,k}^{(j)}=1$ \\
        \While{relative change in $\Phi$ and $\Psi$ is greater than $\delta$}{
            $\Phi_{\alpha,k}^{(i)}=\frac{\sum_{j=1}^{n_{\bm{\xi}}}\Psi_{\alpha,k}^{(j)}R^{(i,j)}}{\sum_{j=1}^{n_{\bm{\xi}}}\Psi_{\alpha,k}^{(j)}\Psi_{\alpha,k}^{(j)}}$ \\
            
            $\Psi_{\alpha,k}^{(j)}=\frac{\sum_{i=1}^{n_{\bm{x}}}\Phi_{\alpha,k}^{(i)}R^{(i,j)}}{\sum_{i=1}^{n_{\bm{x}}}\Phi_{\alpha,k}^{(i)}\Phi_{\alpha,k}^{(i)}}$ \\
            \If{$|\alpha|=1$}{
                $\Psi_{\alpha,k}^{(j)}=P_{\mathcal{H}_\alpha}\Psi_{\alpha,k}^{(j)}$ using Eq.~\eqref{eq:projection_H_i_pce}
            }
            \ElseIf{$|\alpha|=2$}{
                $\Psi_{\alpha,k}^{(j)}=P_{\mathcal{H}_\alpha}\Psi_{\alpha,k}^{(j)}$  using Eq.~\eqref{eq:projection_H_ij_pce}
            }
            Let $s_{\alpha,k}=\left(\frac{1}{n_{\bm{\xi}}}
            \sum_{m=1}^{n_{\bm{\xi}}}
            \left(\Psi_{\alpha,k}^{(m)}\right)^2\right)^{1/2}$ \\
            
            $\Psi_{\alpha,k}^{(j)}=\Psi_{\alpha,k}^{(j)}/s_{\alpha,k}$ \\
            
            $\Phi_{\alpha,k}^{(i)}=\Phi_{\alpha,k}^{(i)}\cdot s_{\alpha,k}$
        }
        
        \If{relative reduction in the residual energy is less than $\tau$}{
            Omit $\Phi_{\alpha,k}$ and $\Psi_{\alpha,k}$ from expansion.\\
            Set $r_\alpha=k-1$ and exit $k$ loop.
        }
        \Else{
        $R^{(i,j)} = R^{(i,j)} - \Phi_{\alpha,k}^{(i)}\Psi_{\alpha,k}^{(j)}$
        }
        
    }
}
\BlankLine \BlankLine
\nlnonumber
\textbf{Neural Network Training} \\
Train one basis NN $\Psi_{\alpha,k}(\bm{\xi}_\alpha)$ for each retained basis using
$\left\{\bm{\xi}_\alpha^{(j)},\Psi_{\alpha,k}^{(j)}\right\}_{j=1,\dots, n_{\bm{\xi}}}$\\

Train a single multi-output coefficient NN using
$\left\{\bm{x}^{(i)},\{\Phi_{\alpha,k}^{(i)}\}\right\}_{i=1,\dots,n_{\bm{x}}}$\\
\BlankLine \BlankLine 

\Output{
Structured Neural Chaos expansion:
\begin{align*}
u(\bm{x},\bm{\xi}) \approx \sum_{\substack{\alpha\subseteq D \\|\alpha|\leq 2}}\sum_{k=1}^{r_\alpha}
\Phi_{\alpha,k}(\bm{x})\Psi_{\alpha,k}(\bm{\xi}_\alpha).
\end{align*}
}
\end{algorithm}

The construction of the sNC expansion is summarized in Algorithm~\ref{alg:sNC} and consists of two main stages. The first stage sequentially generates discrete approximations of the coefficient and basis functions through residual-based alternating least squares (ALS) together with orthogonal projection onto the corresponding ANOVA subspaces. The second stage consists of training NNs using the generated discrete data. 

In the current setting, the data consist of $n_{\bm{x}}$ deterministic grid points and $n_{\bm{\xi}}$ realizations of the stochastic variables,
\begin{align}
    \mathcal{D}=\left\{\left(\bm{x}^{(i)}, \bm{\xi}^{(j)}, u^{(i,j)}=u(\bm{x}^{(i)}, \bm{\xi}^{(j)})\right)\mid i=1,\dots, n_{\bm{x}}, \quad j=1,\dots,n_{\bm{\xi}}\right\}.
\end{align}
Discrete approximations of the coefficients and basis functions are denoted by 
\begin{align}
    \Phi_{\alpha,k}^{(i)}=\Phi_{\alpha,k}(\bm{x}^{(i)}),\ \quad \Psi_{\alpha,k}^{(j)}=\Psi_{\alpha,k}(\bm{\xi}_\alpha^{(j)}),
\end{align} 
respectively. The constant ANOVA component $u_\emptyset=\mathbb{E}_{\bm{\xi}}[u(\bm{x}, \bm{\xi})]$ is approximated at each deterministic grid point using the trivial separable representation
\begin{align}
    \Phi^{(i)}_\emptyset = \frac{1}{n_{\bm{\xi}}}\sum_{j=1}^{n_{\bm{\xi}}} u^{(i,j)}.
\end{align}
The residual is then initialized as 
\begin{align}
    R^{(i,j)}=u^{(i,j)}-\Phi_\emptyset^{(i)}.
\end{align}
The first-order ANOVA components are constructed sequentially, followed by the second-order interaction components. Within each ANOVA component, a rank-$r_\alpha$ approximation is permitted, where a user-prescribed maximum rank may be specified for practical purposes. However, the proposed algorithm adaptively determines the effective rank directly from the data, allowing the prescribed maximum rank to be chosen conservatively without imposing a rigid approximation structure.

The coefficient and basis functions are obtained through ALS using Eq.~\eqref{eq:Phi_update} and Eq.~\eqref{eq:Psi_update}, which are approximated from the discrete data as
\begin{align}
    \Phi_{\alpha,k}^{(i)} = \frac{\mathbb{E}_{\bm{\xi}}[\Psi_{\alpha,k}(\bm{\xi})R(\bm{x}^{(i)}, \bm{\xi})]}{\mathbb{E}_{\bm{\xi}}[\Psi_{\alpha,k}(\bm{\xi})^2]} 
    \approx
    \frac{\sum_{j=1}^{n_{\bm{\xi}}}\Psi_{\alpha,k}^{(j)}R^{(i,j)}}{\sum_{j=1}^{n_{\bm{\xi}}}\Psi_{\alpha,k}^{(j)}\Psi_{\alpha,k}^{(j)}}
\end{align}
for every deterministic grid point $\bm{x}^{(i)}$, and
\begin{align}
    \Psi_{\alpha,k}^{(j)} = \frac{\mathbb{E}_{\bm{x}}[\Phi_{\alpha,k}(\bm{x})R(\bm{x}, \bm{\xi}^{(j)})]}{\mathbb{E}_{\bm{x}}[\Phi_{\alpha,k}(\bm{x})^2]}
    \approx
    \frac{\sum_{i=1}^{n_{\bm{x}}}\Phi_{\alpha,k}^{(i)}R^{(i,j)}}{\sum_{i=1}^{n_{\bm{x}}}\Phi_{\alpha,k}^{(i)}\Phi_{\alpha,k}^{(i)}},
\end{align}
for every stochastic realization $\bm{\xi}^{(j)}$, respectively. Within each ALS iteration, the basis function is projected onto the corresponding ANOVA subspace $\mathcal{H}_\alpha$ using Eq.~\eqref{eq:projection_H_i} or Eq.~\eqref{eq:projection_H_ij}. The projection is applied only to the basis function since the ANOVA decomposition and associated subspaces are defined with respect to the stochastic variables, whereas the coefficient functions remain unconstrained. After projection, the basis function is normalized and the corresponding coefficient function is scaled accordingly in order to preserve the separable representation. The orthogonal projected ALS iterations are repeated until both the coefficient and basis functions satisfy a prescribed convergence tolerance $\delta$. In the present work, $\delta=10^{-6}$ is used for all problems.  

In practice, these orthogonal projections cannot be computed exactly from a finite amount of data. To approximate the projection operators, low-order PCEs are employed. In particular, for first-order ANOVA components, the projection onto $\mathcal{H}_{\{i\}}$ is approximated using a PCE with maximum total degree $p$ in the single random variable $\xi_i$:
\begin{align}
    P_{\mathcal{H}_{\{i\}}}\Psi_{\{i\},k} &= \mathbb{E}[\Psi_{\{i\},k}(\bm{\xi})\mid \xi_i] -\mathbb{E}[\Psi_{\{i\},k}(\bm{\xi})] \\
    &\approx
    \sum_{\ell=1}^{p}
    c_{\ell} \psi_{\ell}(\xi_i),
    \label{eq:projection_H_i_pce}
\end{align}
where $\{\psi_{\ell}\}_{\ell=1}^{p}$ denotes a family of orthonormal polynomial basis functions associated with the probability distribution of $\xi_i$. The subtraction of the constant polynomial mode removes the mean contribution of the conditional expectation, thereby enforcing the zero-mean property required for membership in the ANOVA subspace $\mathcal{H}_{\{i\}}$. For second-order ANOVA components, the projection onto $\mathcal{H}_{\{i,j\}}$ is approximated using a restricted tensor-product PCE with maximum total degree $p$ in the two random variables $\xi_i$ and $\xi_j$:
\begin{align}
    P_{\mathcal{H}_{\{i,j\}}}\Psi_{\{i,j\},k}&=\mathbb{E}[\Psi_{\{i,j\},k}(\bm{\xi})\mid \xi_i,\xi_j]
    -\mathbb{E}[\Psi_{\{i,j\},k}(\bm{\xi})\mid \xi_i]
    -\mathbb{E}[\Psi_{\{i,j\},k}(\bm{\xi})\mid \xi_j]
    +\mathbb{E}[\Psi_{\{i,j\},k}(\bm{\xi})] \\
    &\approx
    \sum_{\substack{\ell_i,\ell_j\geq 1 \\ \ell_i+\ell_j\leq p}}
    c_{\ell_i,\ell_j}
    \psi_{\ell_i}(\xi_i)
    \psi_{\ell_j}(\xi_j).
    \label{eq:projection_H_ij_pce}
\end{align}
Here, $\{\psi_{\ell_i}\}_{\ell_i=1}^{p}$ and $\{\psi_{\ell_j}\}_{\ell_j=1}^{p}$ denote families of orthonormal polynomial basis functions associated with the probability distributions of $\xi_i$ and $\xi_j$, respectively. The exclusion of the constant polynomial mode together with all single-variable polynomial modes ensures that only interaction terms between $\xi_i$ and $\xi_j$ are retained, thereby enforcing membership in the ANOVA subspace $\mathcal{H}_{\{i,j\}}$.

The polynomial basis functions are selected according to the probability distributions of the random variables using the classical Wiener-Askey polynomial families \cite{askey1985}. Alternatively, if the probability distributions are not known explicitly, data-driven approaches may be employed to construct orthonormal polynomial basis functions numerically~\cite{gautschi1982,oladyshkin2012}. In the present work, a maximum total polynomial degree of $p=3$ is used for all problems considered in the present work. This results in projection regression problems involving only 4 unknown coefficients for first-order ANOVA components and 10 unknown coefficients for second-order interaction components. Consequently, the projection step remains low-dimensional and can be efficiently solved using standard linear regression techniques without introducing significant computational complexity. 

To incorporate sparsity into the expansion, the significance of each candidate separable component is assessed after the orthogonal projection-based ALS construction of the discrete coefficient and basis function data $\{\Phi^{(i)}_{\alpha,k}\}_{i=1}^{n_{\bm{x}}}$ and $\{\Psi^{(j)}_{\alpha,k}\}_{j=1}^{n_{\bm{\xi}}}$. In particular, a component is retained only if its inclusion produces a sufficiently large relative reduction in the residual energy. Let 
\begin{align}
    R_{\alpha,k}(\bm{x},\bm{\xi})=R(\bm{x},\bm{\xi})-\Phi_{\alpha,k}(\bm{x})\Psi_{\alpha,k}(\bm{\xi}_\alpha)
\end{align}
denote the updated residual after adding the candidate component. The candidate component is discarded if
\begin{align}
    \frac{\|R(\bm{x},\bm{\xi})\|^2-\|R_{\alpha,k}(\bm{x},\bm{\xi})\|^2}{\|R(\bm{x},\bm{\xi})\|^2}< \tau
    \label{eq:tau_criteria}
\end{align}
where $\tau$ is a prescribed sparsity tolerance parameter controlling the level of sparsity in the expansion. The parameter $\tau$ additionally acts as a natural stopping criterion for the rank enrichment procedure within each ANOVA component. As will be demonstrated, the choice of $\tau$ must balance sparsity and approximation expressiveness in the resulting expansion. The relative reduction in the residual energy is evaluated directly using the discrete data in order to avoid training unnecessary NNs and thereby reduce computational cost. Accordingly, the criterion in Eq.~\eqref{eq:tau_criteria} is approximated as
\begin{align}
    \frac{\|R(\bm{x},\bm{\xi})\|^2-\|R_{\alpha,k}(\bm{x},\bm{\xi})\|^2}{\|R(\bm{x},\bm{\xi})\|^2}
    \approx
    \frac{
    \sum_{i=1}^{n_{\bm{x}}}\sum_{j=1}^{n_{\bm{\xi}}}\left(R^{(i,j)}\right)^2
    -\sum_{i=1}^{n_{\bm{x}}}\sum_{j=1}^{n_{\bm{\xi}}}\left(R_{\alpha,k}^{(i,j)}\right)^2
    }
    {
    \sum_{i=1}^{n_{\bm{x}}}\sum_{j=1}^{n_{\bm{\xi}}}\left(R^{(i,j)}\right)^2
    }<\tau.
\end{align}

\begin{rmk}
    The proposed sparsity mechanism is flexible and can be modified to enforce application-specific structural constraints on the expansion. In the present work, in order to demonstrate that the proposed method can accurately capture even small first-order Sobol' indices, the rank-1 approximation associated with each first-order ANOVA component is always retained. No such restriction is imposed on higher-rank terms or on second-order interaction components.
\end{rmk}

After the orthogonal projection-based ALS construction and sparsity selection procedure are completed, the collected discrete basis and coefficient data are used to train the NNs. In particular, separate basis NNs are trained for each retained basis function, while the complete collection of discrete coefficient data is used to train a single multi-output coefficient NN. For numerical stability, the discrete coefficient data are standardized prior to training and subsequently rescaled to their original scale before constructing the final sNC expansion. Employing a single multi-output network avoids the need to train separate coefficient NNs for each component while additionally mitigating potential ordering bias introduced by the sequential enrichment procedure.

The resulting algorithm produces the following functional approximation of the stochastic process $u(\bm{x},\bm{\xi})$:
\begin{align}
u(\bm{x},\bm{\xi}) \approx \sum_{\substack{\alpha\subseteq D,\\ |\alpha|\leq 2}}\sum_{k=1}^{r_\alpha}
\Phi_{\alpha,k}(\bm{x})\Psi_{\alpha,k}(\bm{\xi}_\alpha)
\end{align}
where $\Phi_{\alpha,k}(\bm{x})$ and $\Psi_{\alpha,k}(\bm{\xi}_\alpha)$ are NN approximations of the coefficient and basis functions, respectively, and $r_\alpha$ denotes the number of retained separable terms associated with the ANOVA component $u_\alpha$. Components with $r_\alpha=0$ are omitted from the expansion.

\begin{rmk}
    The proposed method employs an ANOVA decomposition truncated after second-order interaction terms. This truncation is not an inherent limitation of the proposed framework itself, and higher-order interaction spaces may be incorporated directly using the same residual-driven construction strategy whenever richer interaction structure is required. However, the inclusion of higher-order interaction terms increases the number of expansion components and may require higher polynomial degrees for accurate approximation of the associated projection operators. The development and investigation of higher-order sNC expansions are left for future work.
\end{rmk}

\subsection{sNC-based global sensitivity analysis}

The proposed sNC expansion enables the direct computation of functional statistical quantities and Sobol' sensitivity indices. The final sNC approximation is given by
\begin{align}
u(x,\bm{\xi}) \approx \sum_{\substack{\alpha\subseteq D \\ |\alpha|\leq 2}}\sum_{k=1}^{r_\alpha}
\Phi_{\alpha,k}(\bm{x})\Psi_{\alpha,k}(\bm{\xi}_\alpha),
\end{align}
where $r_\alpha$ denotes the number of retained separable terms associated with the ANOVA component $u_\alpha$, with components satisfying $r_\alpha=0$ omitted from the expansion. 

The functional mean is given by
\begin{align}
    \mathbb{E}[u(\bm{x},\bm{\xi})]\approx \Phi_\emptyset(\bm{x}).
\end{align}
Since the basis functions $\Psi_{\alpha,k}$ are orthonormal and satisfy the ANOVA zero-mean conditions, the total variance in Eq.~\ref{eq:ANOVA_var} is given by
\begin{align}
    \text{Var}[u(\bm{x},\bm{\xi})] \approx \sum_{\substack{\alpha\subseteq D \\ 1\leq|\alpha|\leq 2}}\sum_{k=1}^{r_\alpha} \Phi_{\alpha,k}(\bm{x})^2.
\end{align}

Similarly, the partial variance contribution of the $i$-th variable to the total variance, $\text{Var}[u_i(\bm{x},\xi_i)]$, required for the computation of the first-order Sobol' index for the $i$-th variable in Eq.~\eqref{eq:ANOVA_S_i} is given by
\begin{align}
    \text{Var}[u_{\{i\}}(\bm{x},\xi_i)] \approx \sum_{k=1}^{r_{\{i\}}}\Phi_{\{i\},k}(\bm{x})^2,
    \label{eq:NN_V_i}
\end{align}
and the numerator of the total-order Sobol' index for the $i$-th variable in Eq.~\eqref{eq:ANOVA_S_Ti} is given by
\begin{align}
    V_{T_i} \approx \sum_{\substack{i\in\alpha\subseteq D \\ |\alpha|\leq 2}}\sum_{k=1}^{r_\alpha} \Phi_{\alpha,k}(\bm{x})^2.
    \label{eq:NN_V_Ti}
\end{align}

Consequently, all statistical and sensitivity quantities are obtained directly from the coefficient functions as a post-processing step requiring negligible additional computational cost after training. Importantly, the resulting quantities are themselves functional representations over the deterministic variables $\bm{x}$, which constitutes one of the primary advantages of the proposed framework.
\section{Numerical examples}
\label{sec:Numerical_examples}

In this section, five numerical examples are presented to demonstrate the capability of the proposed sNC framework. The test problems span increasing levels of complexity, from smooth parametric ordinary differential equations to nonlinear partial differential equations exhibiting transport-dominated behavior and shock formation. To validate the GSA results, Monte Carlo simulations (MCS) are used to compute reference estimates of the mean, variance and the first-order and total-order Sobol' indices. Training data are generated using Sobol sequences with sample sizes chosen as powers of two, while each example employs an independently generated randomly sampled test set containing 5000 samples to ensure consistent and robust performance evaluation.

To assess training robustness, each sNC expansion is trained independently ten times for every numerical example. The reported mean, variance, and Sobol' index estimates correspond to averages over these independent training trials. Across all considered examples, the variability between runs was consistently negligible, indicating that the learned decompositions and associated statistical and sensitivity estimates are stable with respect
to network initialization and stochastic optimization. For clarity, this variability is omitted from the figures. 

\begin{table}
\centering
\begin{tabular}{lcccccc}
\toprule
Problem & $d$ & $N$ & Grid resolution & $n^*$ & $\varepsilon_u$ & Training time (s) \\
\midrule
1D nonlinear ODE & 2 & 512 & 20 & 5 & $3.01\times10^{-2}$ & 61.18 \\
1D Euler-Bernoulli beam  & 7 & 4096 & 51 & 24 & $1.53\times10^{-3}$ & 533.77 \\
2D nonlinear elliptic equation & 13 & 8192 & $20\times 20$ & 94 & $1.37\times10^{-3}$ & 3088.77 \\
1D advection--diffusion equation & 6 & 4096 & $50\times50$ & 12 & $8.05\times10^{-4}$ & 10270.86 \\
1D Burgers' equation & 6 & 8192 & $50\times50$ & 22 & $2.75\times10^{-4}$ & 10364.66 \\
\bottomrule
\end{tabular}

\caption{
Surrogate modeling characteristics and training performance of the proposed sNC framework across all benchmark problems. Here, $d$ denotes the stochastic dimension, $N$ denotes the number of stochastic training samples, and $n^*$ denotes the number of retained terms in the final sNC expansion.
}
\label{tab:summary_results1}
\end{table}

\begin{table}
\centering
\begin{tabular}{lcccccc}
\toprule
& \multicolumn{4}{c}{Relative $L^2$ errors} & \multicolumn{2}{c}{Computational time (s)} \\
\cmidrule(lr){2-5} \cmidrule(lr){6-7}
Problem & $\varepsilon_\mu$ & $\varepsilon_{\sigma^2}$ & $\varepsilon_{S_i}$ & $\varepsilon_{S_{T_i}}$ & MCS & \textbf{sNC} \\
\midrule
1D nonlinear ODE 
& $2.48\times 10^{-4}$ & $4.90\times 10^{-3}$ & $1.67\times10^{-3}$ & $1.60\times10^{-3}$ 
& $95.58$ & $\boldsymbol{3.93\times10^{-4}}$ \\

1D Euler-Bernoulli beam 
& $2.00\times10^{-4}$ & $1.89\times10^{-3}$ & $7.87\times10^{-4}$ & $6.40\times10^{-4}$ 
& $1.51\times 10^3$ & $\boldsymbol{2.27\times10^{-3}}$  \\

2D nonlinear elliptic equation
& $3.20\times 10^{-4}$ & $2.52\times 10^{-3}$ & $1.06\times 10^{-2}$ & $6.77\times 10^{-3}$ 
& $4.31\times10^{5}$ & $\boldsymbol{1.92\times10^{-3}}$  \\

1D advection--diffusion equation
& $1.04\times 10^{-4}$ & $9.23\times 10^{-4}$ & $1.11\times 10^{-3}$ & $1.11\times 10^{-3}$ 
& $5.13\times 10^4$ & $\boldsymbol{8.04\times10^{-2}}$  \\

1D Burgers' equation
& $1.33\times 10^{-4}$ & $3.72\times 10^{-3}$ & $6.13\times 10^{-3}$ & $2.68\times 10^{-3}$ 
& $5.15\times 10^4$ & $\boldsymbol{1.43\times10^{-3}}$  \\
\bottomrule
\end{tabular}
\caption{
Summary of uncertainty quantification, global sensitivity analysis, and computational cost for the proposed sNC framework across all benchmark problems. Computational times are compared against MCS estimates computed using $2^{17}$ samples for the 2D nonlinear elliptic equation and $2^{18}$ samples for all other problems. 
}
\label{tab:summary_results2}
\end{table}

Tables~\ref{tab:summary_results1} and~\ref{tab:summary_results2} summarizes the surrogate modeling, uncertainty quantification, global sensitivity analysis, and computational cost results obtained using the proposed sNC framework across all five numerical examples. To enable consistent comparisons across problems, predictive accuracy is quantified using the relative $L^2$ error defined as
\begin{align}
\varepsilon_u = \frac{\|u-\hat{u}\|_2}{\|u\|_2},
\label{eq:varepsilon_u}
\end{align}
where $u$ denotes the reference model response and $\hat{u}$ denotes the corresponding sNC approximation. Similarly, the relative $L^2$ errors associated with the estimated mean and variance are defined as
\begin{align}
    \varepsilon_\mu = \frac{\|\mu-\hat{\mu}\|_2}{\|\mu\|_2},\quad 
    \varepsilon_{\sigma^2} = \frac{\|\sigma^2-\hat{\sigma}^2\|_2}{\|\sigma^2\|_2},
\end{align}
where $\mu$ and $\sigma^2$ denote the reference mean and variance fields obtained through MCS. To assess the overall accuracy of the first-order and total-order Sobol' index estimates, a global relative $L^2$ error aggregated across all Sobol' fields is employed:
\begin{align}
    \varepsilon_{S_i} = \frac{\left( \sum_{i=1}^d\|S_i-\hat{S_i}\|^2_2 \right)^{1/2}}{\left(\sum_{i=1}^d\|S_i\|_2^2 \right)^{1/2}},
    \quad
    \varepsilon_{S_{T_i}} = \frac{\left( \sum_{i=1}^d\|S_{T_i}-\hat{S}_{T_i}\|^2_2 \right)^{1/2}}{\left(\sum_{i=1}^d\|S_{T_i}\|_2^2 \right)^{1/2}}.
\end{align}
where $S_i$ and $S_{T_i}$ denote the reference first-order and total-order Sobol' sensitivity fields obtained through MCS. All computations were performed on CPU-only hardware within a high-performance computing cluster. The sNC framework was implemented using PyTorch, while MCS-based Sobol' estimates were computed using the SALib Python library.

As observed, the proposed sNC achieves excellent numerical accuracy across all examples. As expected, the total training time increases with problem complexity, with the one-dimensional Burgers' equation requiring the longest training time. In terms of the statistical and sensitivity quantities, the proposed framework consistently achieves small relative errors ranging on the order of $10^{-2}$ to $10^{-5}$ across all examples. 

These accurate approximations are obtained while substantially reducing the computational cost required to estimate the statistical and sensitivity quantities relative to the MCS estimates. Although the cost of evaluating the trained sNC surrogate is negligible, even when accounting for training, data generation, and post-processing costs, the proposed framework remains computationally more efficient than the corresponding MCS. This computational advantage becomes increasingly pronounced as the complexity of the underlying model grows. Furthermore, once trained, the sNC framework provides Sobol' indices as continuous functions of the deterministic variables rather than only at fixed grid locations. In contrast, MCS-based estimates of functional Sobol' fields are generally restricted to the discrete computational grid used during simulation.

\subsection{Example 1: One-dimensional nonlinear ODE with 2 random variables}
\label{subsec:1D2RV}

We begin by considering a first-order stochastic nonlinear ODE:
\begin{align}
    &\frac{du}{dx} = -\left( 2x\xi_1 + 3x^2\xi_2 + 4x^3\xi_1\xi_2 \right) \exp\left( x^2\xi_1+x^3\xi_2+x^4 \xi_1 \xi_2 \right), \quad u(0,\bm{\xi})=0,
\end{align}
where $\xi_1$ and $\xi_2$ are independent random variables uniformly distributed on $[0,1]$. Despite its relative simple structure, the problem contains both non-negligible first-order effects and interaction contributions, making it suitable baseline example for assessing the ability of the proposed framework to recover global sensitivity measures and interaction structure. The problem admits a closed-form solution given by
\begin{align}
    u(x,\bm{\xi})=1-\exp\left( x^2\xi_1+x^3\xi_2+x^4 \xi_1 \xi_2 \right).
\end{align}
A total of 512 training samples are generated and the corresponding solutions are evaluated at 20 equidistant points over the spatial domain $x\in[0,1]$.

\subsubsection{Surrogate accuracy and structure}
\begin{figure}
  \center
  \includegraphics[width=4.5in]{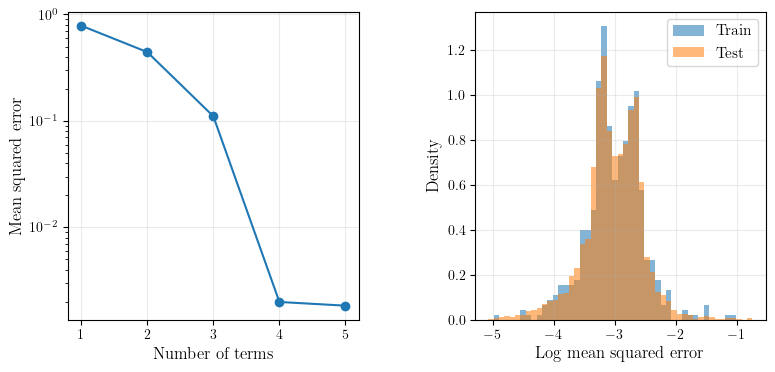} \\
  \makebox[\linewidth][l]{\hspace{2in} \small (a) \hspace{2.1in} (b)} \\
  \vspace{1em}
  \includegraphics[width=\textwidth]{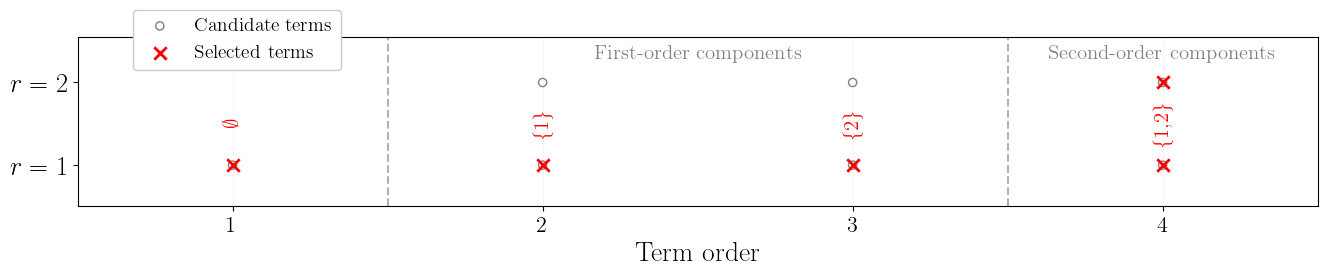}
  \makebox[\linewidth][c]{\small (c)}\\
  \vspace{0.2em}
  \caption{One-dimensional nonlinear ODE. (a)~Mean squared error of the sNC expansion with respect to the number of retained terms. (b)~Distribution of training and testing errors; the training and testing sets consist of 512 and 5000 stochastic samples, respectively. (c)~Candidate terms grouped by interaction structure, together with the associated approximation rank $r$. The index sets of the selected terms in the final expansion are highlighted in red.}
  \label{fig:1D2RV_errors}
\end{figure}

The mean squared error of the sNC expansion with respect to the number of retained terms is shown in Fig.~\ref{fig:1D2RV_errors}(a), where the error decreases as additional terms are incorporated with the exception of the final term. The error distribution between the model predictions and the ground truth is shown in Fig.~\ref{fig:1D2RV_errors}(b). The close agreement between training and testing errors indicates strong generalization of the sNC expansion. Fig.~\ref{fig:1D2RV_errors}(c) shows that all four candidate interaction terms are retained, with the single-variable terms requiring only a rank-1 approximation and the two-variable term admitting a rank-2 approximation. That is, for this problem, the final sNC expansion is
\begin{align}
    u(x,\bm{\xi})\approx \Phi_\emptyset(x)+\Phi_{\{1\}}(x)\Psi_{\{1\}}(\xi_1)+\Phi_{\{2\}}(x)\Psi_{\{2\}}(\xi_2) + \sum_{k=1}^2\Phi_{\{1,2\},k}(x)\Psi_{\{1,2\},k}(\xi_1, \xi_2).
\end{align}
For clarity, the subscript $k$ is omitted from the coefficient and basis functions associated with rank-1 approximations. This convention will be adopted throughout the remainder of the paper. 

Although a rank-2 approximation is identified for the interaction term, it provides only a marginal reduction in error, indicating that the interaction is dominated by a single mode with a weaker secondary component. In this example, the expansion is not truncated, as all candidate interaction terms are retained, corresponding to the full Sobol decomposition; the primary reduction in complexity arises from the low-rank approximation of each term. 

\begin{figure}
  \center
  \includegraphics[width=2.3in]{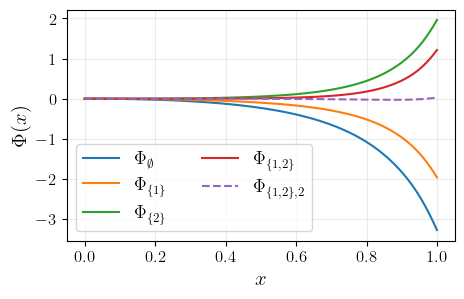}\\
  \makebox[\linewidth][c]{\small (a)}\\
  \vspace{1em}
  \includegraphics[width=\textwidth]{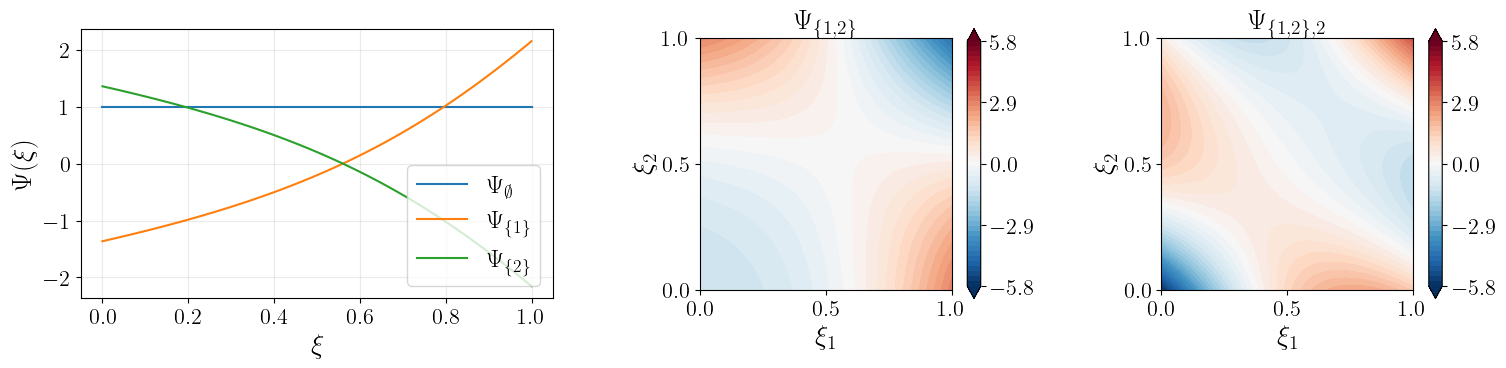}\\
  \makebox[\linewidth][l]{\hspace{1.25in} \small (b) \hspace{1.9in} (c) \hspace{1.55in} (d) } \\
  \vspace{0.2em}
  \caption{One-dimensional nonlinear ODE. (a)~coefficient functions corresponding to the retained terms in the sNC expansion. (b)~Basis functions associated with single-variable terms. (c)–(d)~Components of the rank-2 stochastic approximation for the two-variable interaction term.}
  \label{fig:1D2RV_coeff_basis}
\end{figure}

Fig.~\ref{fig:1D2RV_coeff_basis} shows the learned coefficient functions and basis functions. The coefficient functions in Fig.~\ref{fig:1D2RV_coeff_basis}(a) are smooth and exhibit similar spatial trends with varying magnitudes across terms, and the single-variable basis functions in Fig.~\ref{fig:1D2RV_coeff_basis}(b) exhibit simple, low-complexity structure. The interaction basis functions in Fig.~\ref{fig:1D2RV_coeff_basis}(c) and Fig.~\ref{fig:1D2RV_coeff_basis}(d) capture the joint dependence on $\xi_1$ and $\xi_2$, with the rank-2 approximation introducing a secondary mode. Although the secondary component exhibits a similar magnitude, its contribution to the solution is small due to the associated coefficient, indicating that it represents a weak but identifiable structural feature of the interaction and remains informative for characterizing the underlying structure. 

\subsubsection{Statistical and sensitivity estimates}

\begin{figure}
  \center
  \includegraphics[width=\textwidth]{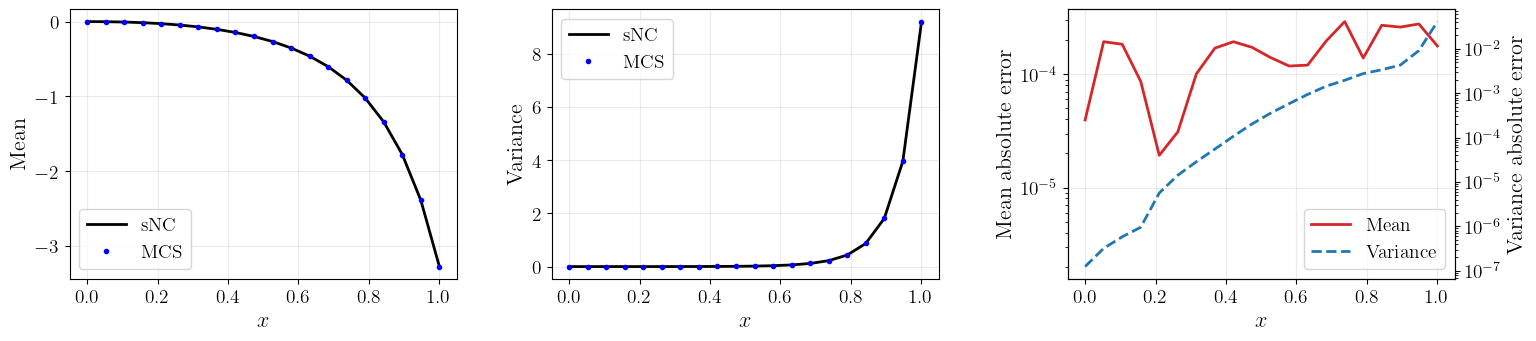} \\
  \makebox[\linewidth][l]{\hspace{1.1in} \small (a) \hspace{1.6in} (b) \hspace{1.85in} (c)} \\
  \vspace{0.2em}
  \caption{One-dimensional nonlinear ODE. (a)~Mean and (b)~variance estimates computed directly from the coefficient functions of the sNC expansion, compared with reference values estimated via MCS using $2^{18}$ realizations. (c)~Absolute errors in the mean and variance.}
  \label{fig:1D2RV_mean_var}
\end{figure}

Fig.~\ref{fig:1D2RV_mean_var} compares the mean and variance computed directly from the coefficient functions of the sNC expansion with reference values estimated via MCS with $2^{18}$ simulations; the corresponding absolute errors across the spatial domain are also shown. The sNC estimates closely match the MCS results, and the corresponding errors remain small throughout the domain for both the mean and variance estimates.

\begin{figure}
  \center
  \includegraphics[width=\textwidth]{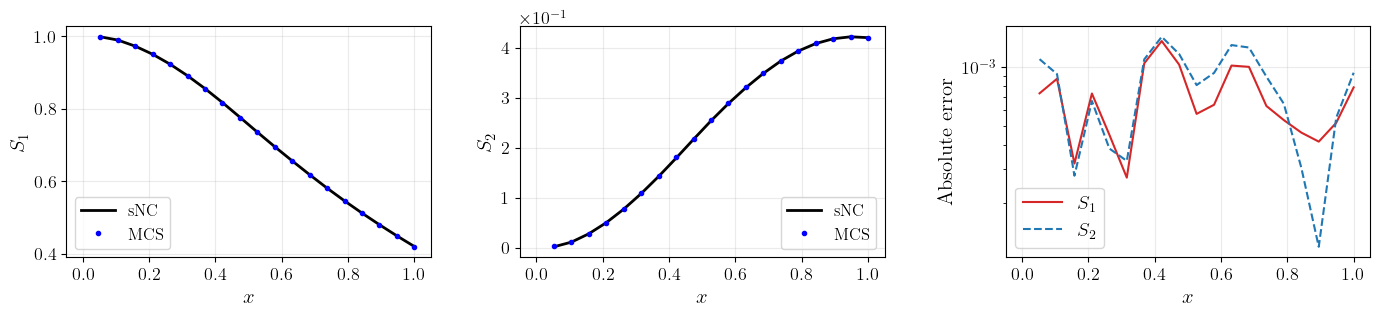} \\
  \makebox[\linewidth][l]{\hspace{1.1in} \small (a) \hspace{1.8in} (b) \hspace{1.8in} (c)} \\
  \vspace{1em}
  \includegraphics[width=\textwidth]{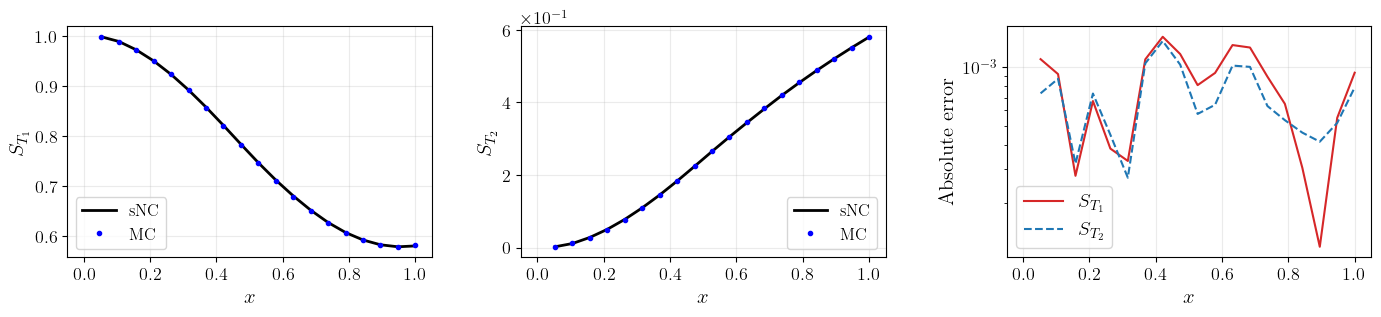}\\
  \makebox[\linewidth][l]{\hspace{1.1in} \small (d) \hspace{1.8in} (e) \hspace{1.8in} (f)} \\
  \vspace{0.2em}
  \caption{One-dimensional nonlinear ODE. (a)-(b)~First-order Sobol' indices and (c)~their absolute error. (d)-(e)~Total-order Sobol' indices and (f)~their absolute error. The Sobol' indices are computed directly from the coefficient functions of the sNC expansion and compared with reference values estimated via MCS using $2^{18}$ realizations. Values at locations where the total variance vanishes are omitted, as the Sobol' indices are undefined there.}
  \label{fig:1D2RV_GSA}
\end{figure}

A primary focus of the proposed framework is the efficient estimation of global sensitivity measures directly from the learned sNC expansion. The first-order and total-order Sobol' indices computed from the coefficient functions of the sNC expansion, along with the reference MCS estimates, are shown in Fig.~\ref{fig:1D2RV_GSA}. The sNC estimates closely match the MCS results across the spatial domain for both the first-order and total-order indices, and the corresponding absolute errors remain small throughout the domain.

To assess training robustness, the model was trained independently ten times, with negligible variability observed across the resulting mean, variance, and Sobol' index estimates; for clarity, this variability is not shown in Fig.~\ref{fig:1D2RV_mean_var} and Fig.~\ref{fig:1D2RV_GSA}. This consistency indicates that the learned decomposition and associated estimates are stable with respect to network initialization and stochastic optimization. In particular, the stable behavior observed in regions of small variance is a consequence of the training strategy adopted in the present work, as discussed in Appendix~\ref{sec:Appendix_NN_hyperparameters}. Finally, the combined cost of constructing the sNC surrogate and subsequently evaluating the statistical quantities and sensitivity indices remains lower than the cost of directly estimating the same quantities through MCS (see Tables~\ref{tab:summary_results1} and~\ref{tab:summary_results2}). Although the computational savings are naturally modest for this simple problem with an available closed-form solution, the benefit is expected to become more pronounced for higher-dimensional and computationally expensive problems.

\subsection{Example 2: One-dimensional Euler-Bernoulli beam with 7 random variables}
\label{subsec:1D7RV}

In this example, we consider a stochastic Euler-Bernoulli beam with seven random input variables following the setup in \cite{bahmani2025neural}. The bending behavior of a beam of length $L=\SI{10}{\meter}$ subjected to a uniformly distributed load $q=\SI{-5}{\kilo\newton\per\meter}$ is governed by
\begin{align}
    \frac{d^2}{dx^2} \left( K(x,\bm{\xi})\frac{d^2 u}{dx^2} \right) = -0.005, \quad x\in[0,10].
\end{align}
Here, $u(x,\bm{\xi})$ denotes the beam deflection and $K(x,\bm{\xi})$ is the bending stiffness. The beam is assumed to be simply supported, which yields the boundary conditions
\begin{align}
    u(0,\bm{\xi})=u(10,\bm{\xi})=0, \quad M(0,\bm{\xi})=M(10,\bm{\xi})=0,
\end{align}
where the bending moment is defined as
\begin{align}
    M(x,\bm{\xi}) = -K(x, \bm{\xi})\frac{d^2 u}{dx^2}.
\end{align}
The uncertainty in the deflection field is assumed to arise solely from the stiffness random field $K(x, \bm{\xi})$, which is modeled as a one-dimensional Gaussian random field with mean $\mu_K=\SI{8}{\giga\pascal}$ and covariance kernel
\begin{align}
    \text{Cov}(x_1, x_2) = \exp\left( -\frac{(x_1-x_2)^2}{2l_c^2} \right),
\end{align}
where $l_c=\SI{2}{\meter}$ is the correlation length. The random field is approximated using a truncated Karhunen-Lo\`eve (KL) expansion \cite{ghanem2003stochastic} with 7 modes, resulting in a seven-dimensional stochastic problem parameterized by independent standard Gaussian random variables $\{\xi_i\}_{i=1}^7$. This problem provides a suitable setting for assessing the proposed framework in a higher dimensional stochastic setting, while also examining the influence of the global selection tolerance $\tau$ on the learned decomposition and the role of higher-rank approximations in accurately recovering the Sobol' sensitivity structure. A total of 4096 realizations of the random field are generated and the governing equation is solved numerically on a uniform grid of 51 equidistant points over the spatial domain $x\in[0,10]$.

\subsubsection{Surrogate accuracy and structure}
\begin{figure}
  \center
  \includegraphics[width=4.5in]{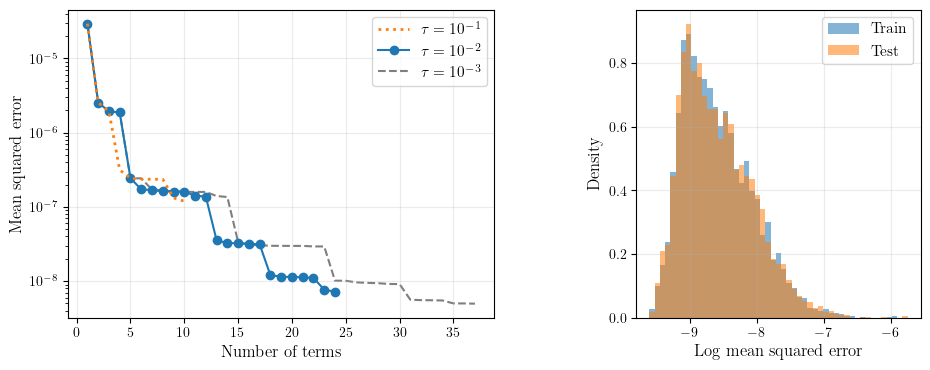} \\
  \makebox[\linewidth][c]{\small (a) \hspace{2in} (b)} \\
  \vspace{1em}
  \includegraphics[width=\textwidth]{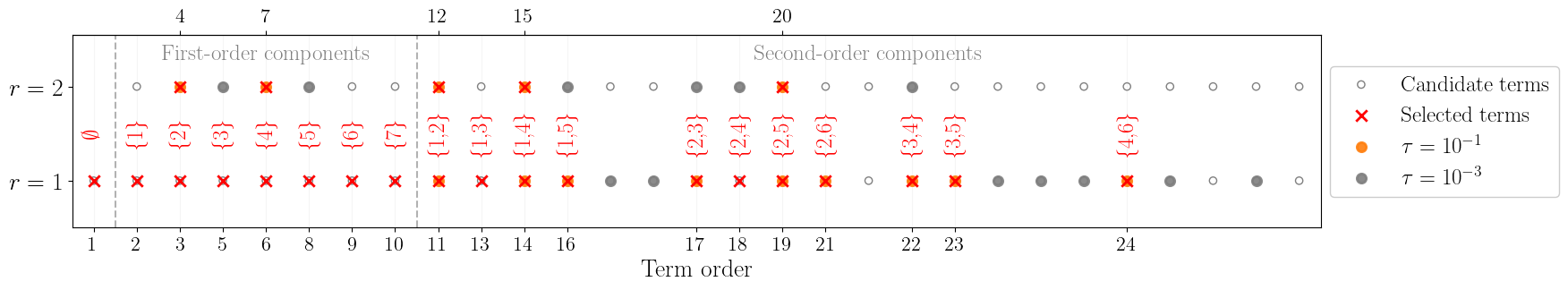}
  \makebox[\linewidth][c]{\small (c)} \\
  \vspace{0.2em}
  \caption{One-dimensional Euler-Bernoulli beam. (a)~Mean squared error of the sNC expansion with respect to the number of retained terms for different values of the global selection tolerance $\tau$. (b)~Distribution of training and testing errors; the training and testing sets consist of 4096 and 5000 stochastic samples, respectively. (c)~Candidate terms grouped by interaction structure, together with the associated approximation rank $r$. The index sets of the selected terms in the final expansion are highlighted in red, while orange and gray markers indicate terms removed under aggressive pruning $(\tau=10^{-1})$ and additional terms retained under relaxed selection $(\tau=10^{-3})$, respectively.}
  \label{fig:1D7RV_errors}
\end{figure}

The mean squared error of the sNC expansion with respect to the number of retained terms is shown in Fig.~\ref{fig:1D7RV_errors}(a), where the final expansion trained with $\tau=10^{-2}$ is shown in blue. The orange dotted curve corresponds to a large global selection tolerance ($\tau=10^{-1}$), resulting in aggressive pruning, while the gray dashed curve corresponds to a smaller tolerance ($\tau=10^{-3}$), which yields a more relaxed selection strategy. All three expansions exhibit an overall decreasing trend as additional terms are retained. While aggressive pruning reduces the complexity of the expansion, the resulting approximation does not fully realize the predictive accuracy achievable with more relaxed selection strategies. Conversely, overly relaxed selection increases the complexity of the expansion while providing only diminishing reductions in the approximation error.

The error distribution between the model predictions and the ground truth for the final sNC expansion containing 24 terms is shown in Fig.~\ref{fig:1D7RV_errors}(b), where the close agreement between training and testing errors indicates strong generalization of the learned expansion. Fig.~\ref{fig:1D7RV_errors}(c) shows the retained terms in the final sNC expansion highlighted in red. Most terms adopt rank-1 approximations, while two single-variable terms and three pairwise interaction terms acquire rank-2 approximations. According to Eq.~\eqref{eq:n}, the total number of terms associated with a full rank-$r$ approximation is $1+28 r$; therefore, the final sNC expansion containing $n^*=24$ terms remains sparse in comparison. The orange and gray markers further indicate terms removed under aggressive pruning and additional terms retained under relaxed selection, respectively.

\begin{figure}
  \center
  \includegraphics[width=\textwidth]{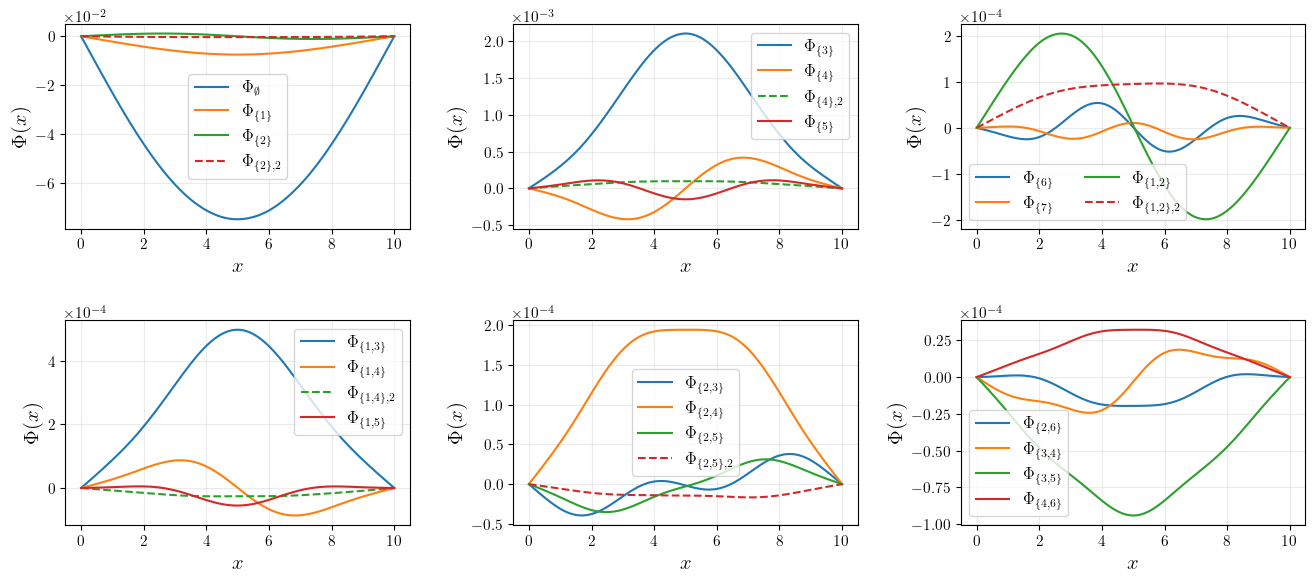}\\
  \vspace{0.2em}
  \caption{One-dimensional Euler-Bernoulli beam. Coefficient functions corresponding to the retained terms in the sNC expansion ($\tau=10^{-2}$). Dashed curves denote the secondary components associated with rank-2 approximations.}
  \label{fig:1D_7RV_coeff}
\end{figure}

\begin{figure}
  \center
  \includegraphics[width=5.5in]{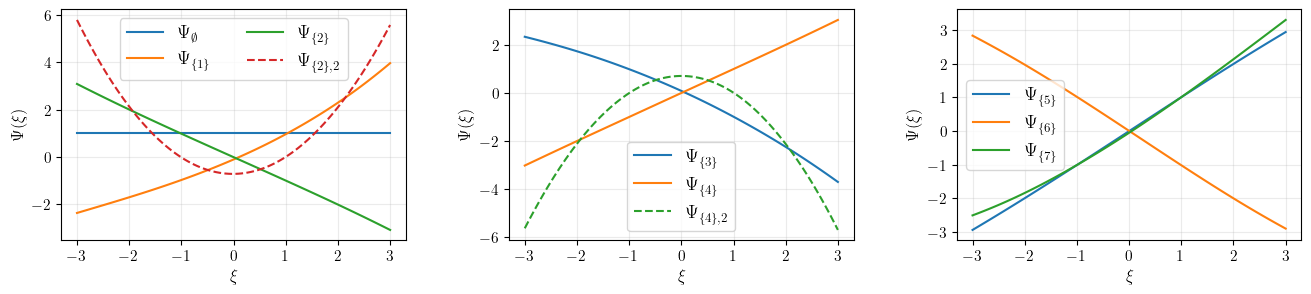} \\
  \makebox[\linewidth][c]{\small (a)}\\
  \vspace{1em}
  \includegraphics[width=\textwidth]{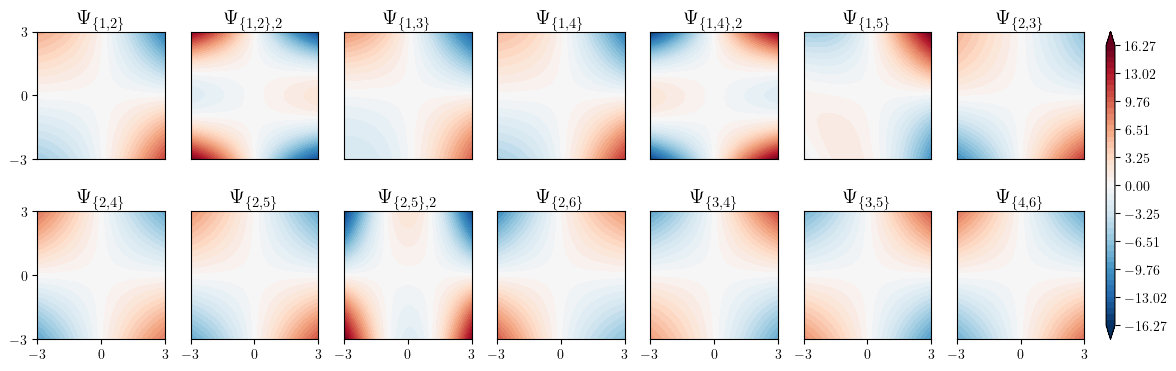}\\
  \makebox[\linewidth][c]{\small (b)} \\
  \vspace{0.2em}
  \caption{One-dimensional Euler-Bernoulli beam. (a)~Basis functions corresponding to the retained single-variable terms in the sNC expansion ($\tau=10^{-2}$). Dashed curves denote the secondary components associated with rank-2 approximations. (b)~Basis functions associated with the retained pairwise interaction terms. The secondary components corresponding to rank-2 approximations are explicitly labeled.}
  \label{fig:1D7RV_basis}
\end{figure}

Fig.~\ref{fig:1D_7RV_coeff} shows the learned coefficient functions, which remain smooth and exhibit a clear spatial structure across the domain. The magnitudes of the retained coefficient functions generally decrease with the ordering of the associated KL modes, reflecting the hierarchical structure induced by the eigenvalue decay of the stochastic stiffness field. The learned basis functions are shown in Fig.~\ref{fig:1D7RV_basis}. The single-variable basis functions in Fig.~\ref{fig:1D7RV_basis}(a) exhibit a relatively simple low-order structure, primarily resembling linear and quadratic forms. Similarly, the pairwise interaction basis functions shown in Fig.~\ref{fig:1D7RV_basis}(b) remain smooth and structured, while the secondary rank-2 components exhibit additional curvature and sign-changing behavior indicative of increased representational complexity. These observations suggest that the proposed framework adaptively identifies compact stochastic representations, introducing additional representational complexity only when required to accurately capture the underlying stochastic behavior. 

\subsubsection{Statistical and sensitivity estimates}
\begin{figure}
  \center
  \includegraphics[width=\textwidth]{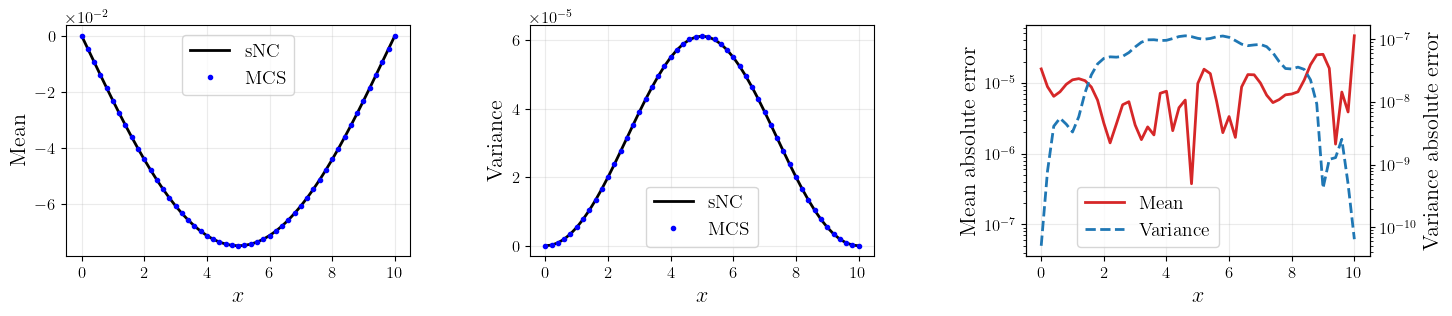} \\
  \makebox[\linewidth][l]{\hspace{1.1in} \small (a) \hspace{1.7in} (b) \hspace{1.7in} (c)}\\
  \vspace{0.2em}
  \caption{One-dimensional Euler-Bernoulli beam. (a)~Mean and (b)~variance estimates computed directly from the coefficient functions of the sNC expansion ($\tau=10^{-2}$), compared with reference values estimated via MCS using $2^{18}$ realizations. (c)~Absolute errors in the mean and variance.}
  \label{fig:1D7RV_mean_var}
\end{figure}

Fig.~\ref{fig:1D7RV_mean_var} compares the mean and variance computed directly from the coefficient functions of the sNC expansion with reference values estimated via MCS with $2^{18}$ simulations; the corresponding absolute errors across the spatial domain are also shown. The close agreement between the sNC estimates and the MCS results is evident in this example as well, with small absolute errors observed throughout the domain for both the mean and variance. The mean and variance estimates obtained from the expansions trained with $\tau=10^{-1}$ and $\tau=10^{-3}$ are nearly indistinguishable from those of the final expansion and are therefore omitted from the figure for clarity. 

\begin{figure}
  \center
  \includegraphics[width=\textwidth]{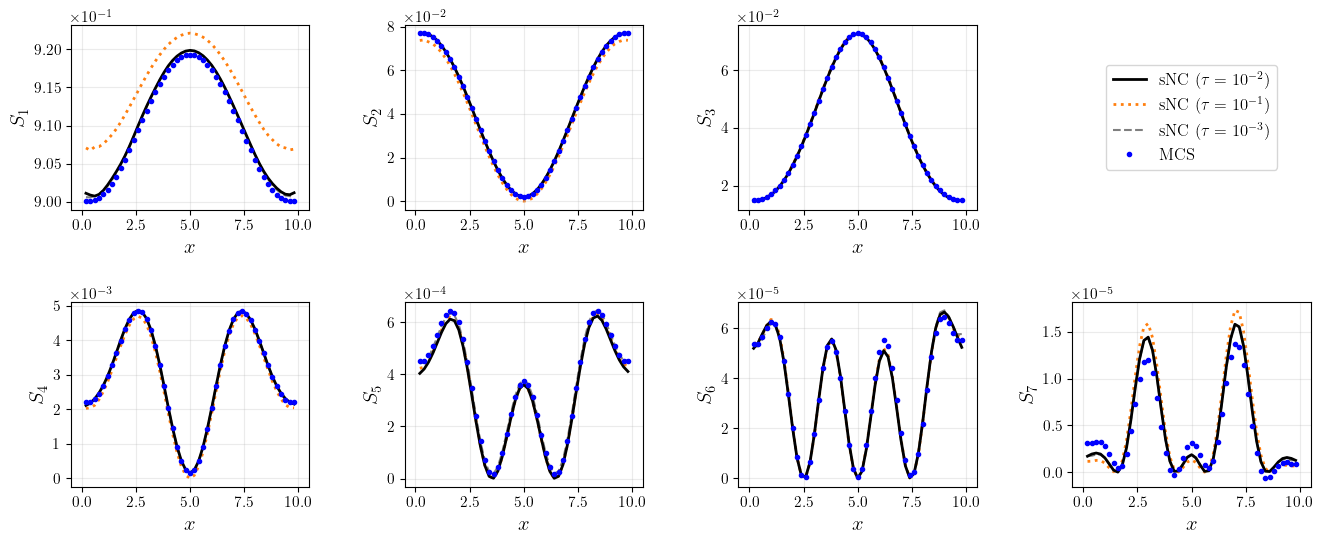}\\
  \makebox[\linewidth][c]{\small (a)} \\
  \vspace{1em}
  \includegraphics[width=\textwidth]{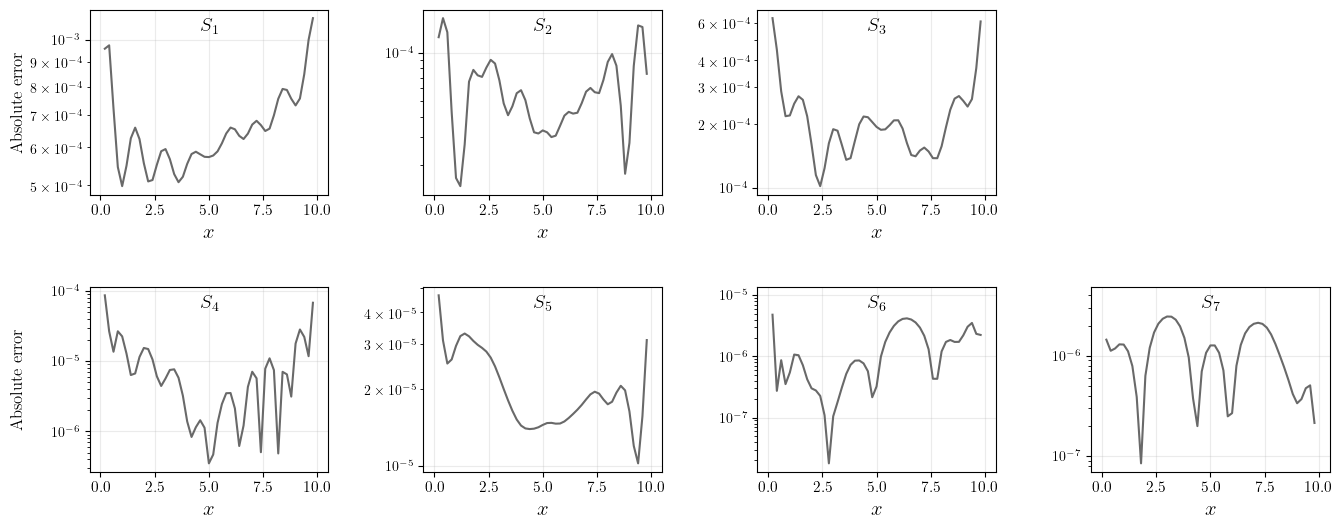}
  \makebox[\linewidth][c]{\small (b)} \\
  \vspace{0.2em}
  \caption{One-dimensional Euler-Bernoulli beam. (a)~First-order Sobol' indices computed directly from the coefficient functions of the sNC expansion for different values of the global selection tolerance $\tau$, compared with reference values estimated via MCS using $2^{18}$ realizations. (b)~Absolute error of the sNC ($\tau=10^{-2}$) Sobol' index estimates relative to the MCS estimates. Values at locations where the total variance vanishes are omitted, as the Sobol' indices are undefined there.}
  \label{fig:1D7RV_GSA_first}
\end{figure}

The first-order Sobol' indices computed directly from the coefficients of the sNC expansion, along with the reference MCS estimates and the corresponding absolute errors, are shown in Fig.~\ref{fig:1D7RV_GSA_first}. The solid black curves correspond to the main results, the sNC expansion with $\tau=10^{-2}$, while the orange dotted and gray dashed curves correspond to the aggressively pruned ($\tau=10^{-1}$) and relaxed selection ($\tau=10^{-3}$) expansions, respectively. The relaxed-selection results are nearly identical to those of the reference expansion, indicating that the additional retained terms contribute only marginally to the resulting sensitivity estimates. In contrast, noticeable discrepancies are observed for the aggressively pruned expansion, particularly for $S_1$. In this case, aggressive pruning removes important variance contributions, leading to an underestimation of the total variance and consequently an overestimation of the normalized first-order Sobol' index. 

Importantly, these discrepancies are primarily associated with the omission of crucial rank-2 approximation components, as a purely rank-1 approximation exhibits similar overestimation behavior. Nevertheless, all three expansions correctly recover the hierarchical ordering and spatial structure of the dominant sensitivity contributions. The absolute errors associated with the main model, the sNC expansion with $\tau=10^{-2}$, relative to the MCS estimates using $2^{18}$ realizations remain small throughout the domain. The errors become slightly elevated near locations where the total variance is small, since the Sobol' indices become increasingly sensitive to small inaccuracies in the variance estimates in these regions.

\begin{figure}
  \center
  \includegraphics[width=\textwidth]{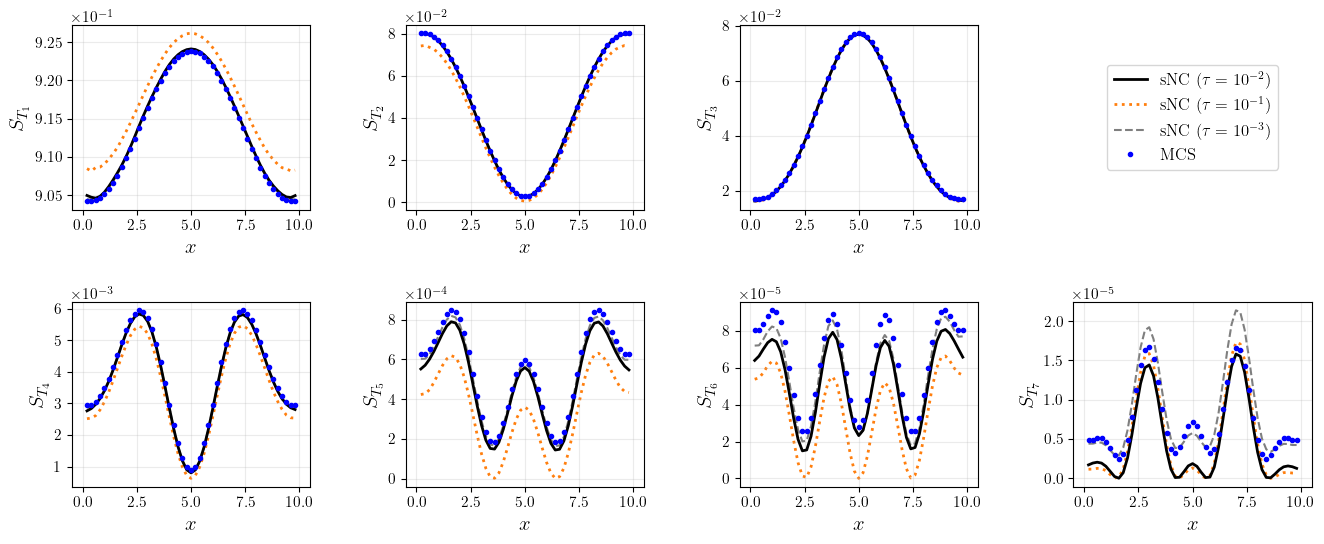}\\
  \makebox[\linewidth][c]{\small (a)} \\
  \vspace{1em}
  \includegraphics[width=\textwidth]{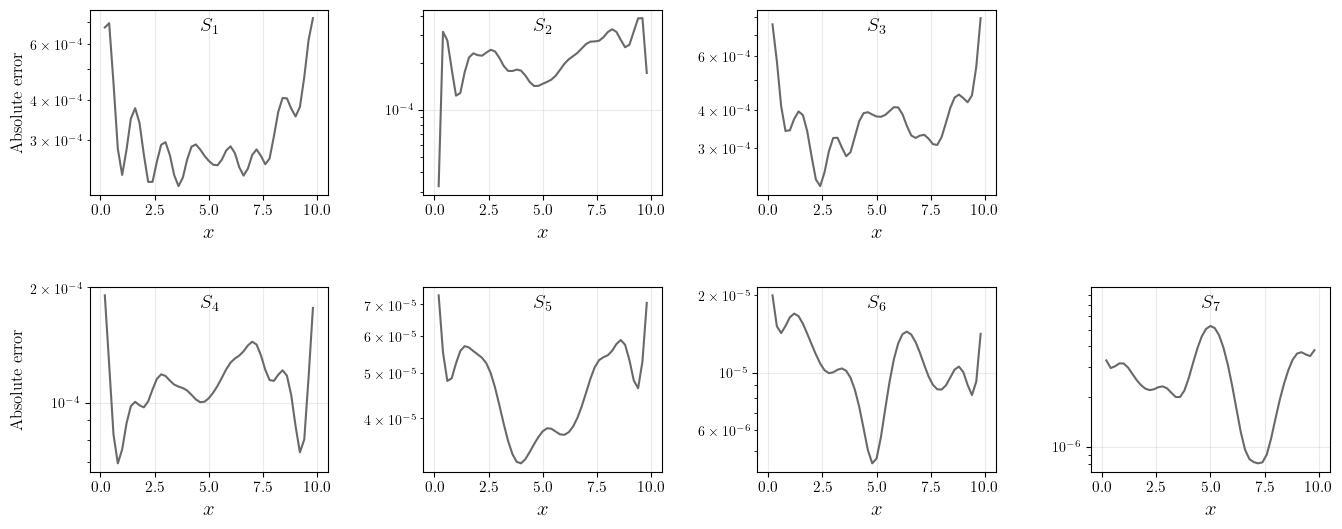} \\
  \makebox[\linewidth][c]{\small (b)} \\
  \vspace{0.2em}
  \caption{One-dimensional Euler-Bernoulli beam. (a)~Total-order Sobol' indices computed directly from the coefficient functions of the sNC expansion for different values of the global selection tolerance $\tau$, compared with reference values estimated via MCS using $2^{18}$ realizations. (b)~Absolute error of the sNC ($\tau=10^{-2}$) Sobol' index estimates relative to the MCS estimates. Values at locations where the total variance vanishes are omitted, as the Sobol' indices are undefined there.}
  \label{fig:1D7RV_GSA_total}
\end{figure}

The total-order Sobol' indices computed directly from the coefficient functions of the sNC expansion, along with the reference MCS estimates and the corresponding absolute errors, are shown in Fig.~\ref{fig:1D7RV_GSA_total}. Similar behavior to that observed for the first-order indices is evident for $S_{T_1}$, where the aggressively pruned expansion ($\tau=10^{-1}$) produces a noticeable overestimation. Additionally, aggressive pruning leads to an underestimation of several of the remaining total-order indices due to the omission of interaction terms necessary to accurately capture the associated variance contributions. In the present example, the interaction contributions remain relatively weak, such that the omission of several pairwise interaction terms has only a limited impact on the overall sensitivity characterization; however, more pronounced degradation would be expected for problems exhibiting stronger interaction effects.

Differences between the reference and relaxed-selection expansion are primarily visible only for the smallest total-order indices, where the additional retained terms improvements in the recovered sensitivity estimates. Nevertheless, the dominant sensitivity hierarchy and qualitative interpretation of the problem remain unchanged across all three expansions, indicating that the additional complexity introduced by the relaxed-selection strategy has only limited practical impact for this example. As with the first-order indices, the absolute errors become slightly elevated near locations where the total variance is small.

Collectively, the results for this example highlight the importance of appropriately selecting the global selection tolerance to balance sparsity and predictive accuracy while preserving reliable recovery of the underlying sensitivity structure. While several retained terms produce only marginal reductions in the global approximation error, these terms are nevertheless retained by the adaptive selection procedure because they remain important for accurately recovering the Sobol' index estimates, which constitute the primary focus of the proposed framework.

\subsection{Example 3: Two-dimensional nonlinear elliptic equation with 13 random variables}
\label{subsec:2D13RV}

For the next problem, we consider a nonlinear elliptic equation in a two-dimensional spatial domain. Specifically, a heat conduction problem with homogeneous Dirichlet boundary conditions:
\begin{align}
    \nabla\cdot\left( -\left( 1+\frac{u^2}{2} \right) \nabla u \right) = 100\sin(5x_1)\sin(4x_2) + f(\bm{x}, \bm{\xi})^2, \quad \bm{x}\in [0,1]^2.
\end{align}
subject to $u(\bm{x}, \bm{\xi})=0$ on the boundary, where $u(\bm{x}, \bm{\xi})$ denotes the temperature field. The heat source $f(\bm{x},\bm{\xi})$ is modeled as a zero-mean, two-dimensional Gaussian random field with covariance kernel
\begin{align}
    \text{Cov}(\bm{x}_i, \bm{x}_j) = \exp\left( -\frac{\lVert \bm{x}_i - \bm{x}_j \rVert_2^2}{2l_c^2} \right),
\end{align}
where the correlation length is set to $l_c=\SI{2}{\meter}$. This problem follows the setup in \cite{bahmani2025neural}, where the source term is squared to introduce stronger interaction effects while preserving the underlying physical structure of the problem. The random field is approximated using a truncated KL expansion with 13 modes, resulting in a 13-dimensional stochastic problem parameterized by independent standard Gaussian random variables $\{\xi_i\}_{i=1}^{13}$. This problem provides a challenging benchmark for assessing the proposed framework in a high-dimensional nonlinear setting with significant interaction effects and a broad range of influential variables contributing to the global sensitivity structure. A total of 8192 realizations of the random field are generated and the governing equation is solved numerically on a uniform $20\times 20$ spatial grid.

\subsubsection{Surrogate accuracy and structure}

\begin{figure}
  \center
  \includegraphics[width=\textwidth]{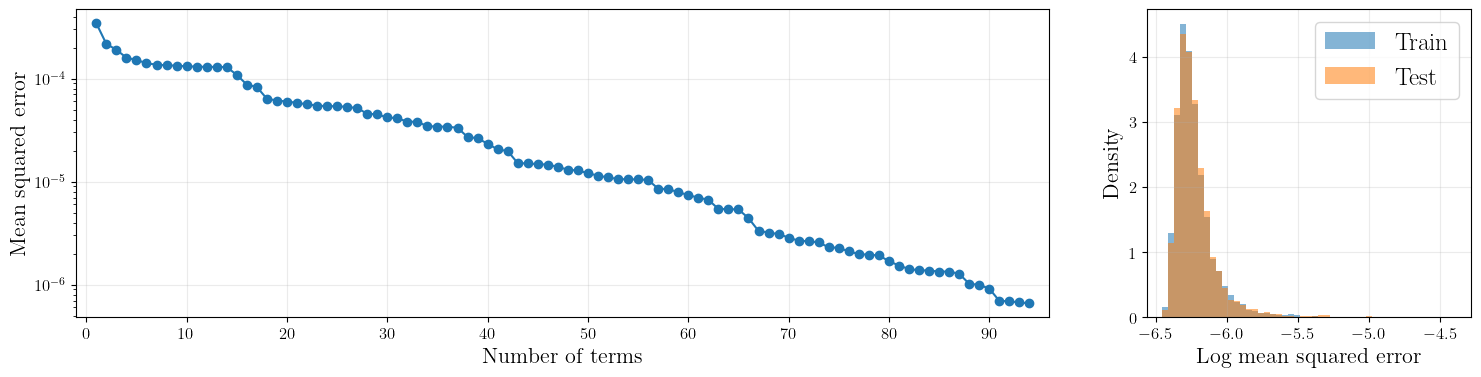} \\
  \makebox[\linewidth][c]{\small \hspace{1.5in} (a) \hspace{2.55in} (b)} \\
  \vspace{1em}
  \includegraphics[width=\textwidth]{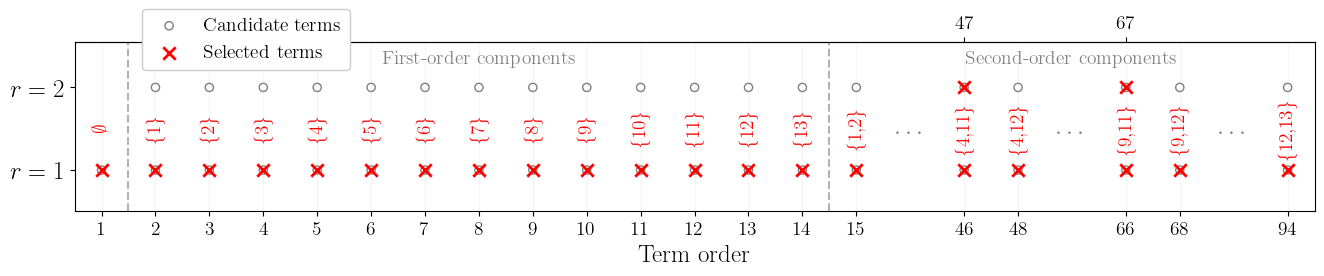}
  \makebox[\linewidth][c]{\small (c)} \\
  \vspace{0.2em}
  \caption{Two-dimensional nonlinear elliptic equation. (a)~Mean squared error of the sNC expansion with respect to the number of retained terms. (b)~Distribution of training and testing errors; the training and testing sets consist of 8192 and 5000 stochastic samples, respectively. (c)~Candidate terms grouped by interaction structure, together with the associated approximation rank $r$. The index sets of the selected terms in the final expansion are highlighted in red.}
  \label{fig:2D13RV_errors}
\end{figure}

The mean squared error of the sNC expansion with respect to the number of retained terms is shown in Fig.~\ref{fig:2D13RV_errors}(a), where the error decreases progressively as additional terms are incorporated into the expansion. The error distribution between the model predictions and the ground truth is shown in Fig.~\ref{fig:2D13RV_errors}(b). The distributions remain strongly concentrated around low error values, with a small tail corresponding to a limited number of more challenging stochastic realizations. The close agreement between the training and testing distributions indicates strong generalization of the learned expansion. Fig.~\ref{fig:2D13RV_errors}(c) shows the retained terms in the final sNC expansion highlighted in red. In contrast to the previous example, all candidate terms are retained in the final expansion; all terms adopt a rank-1 approximation except for two terms which adopt a rank-2 approximation, resulting in a total of $n^*=94$ retained terms. 

\begin{figure}
  \centering
  \includegraphics[width=5.89in]{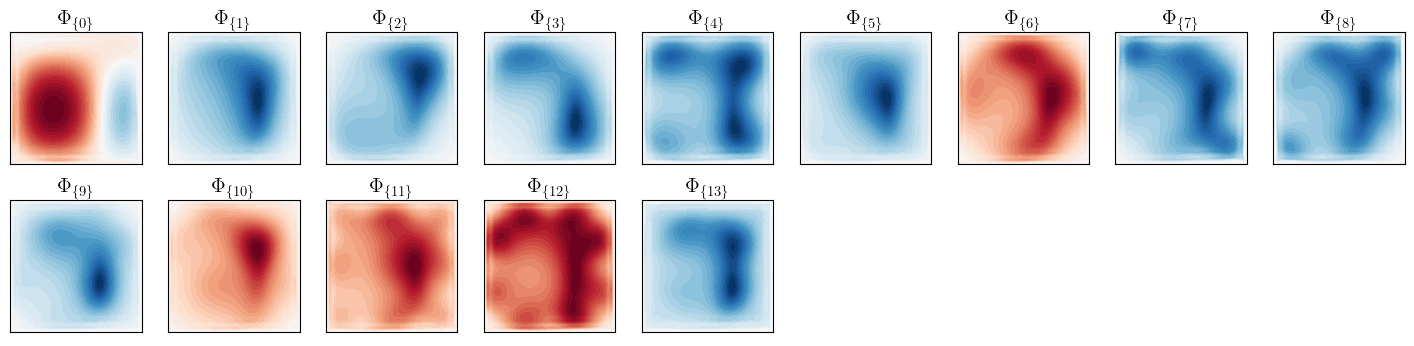}
  \makebox[\linewidth][c]{\small (a)} \\
  \vspace{1em}
  \includegraphics[width=5.89in]{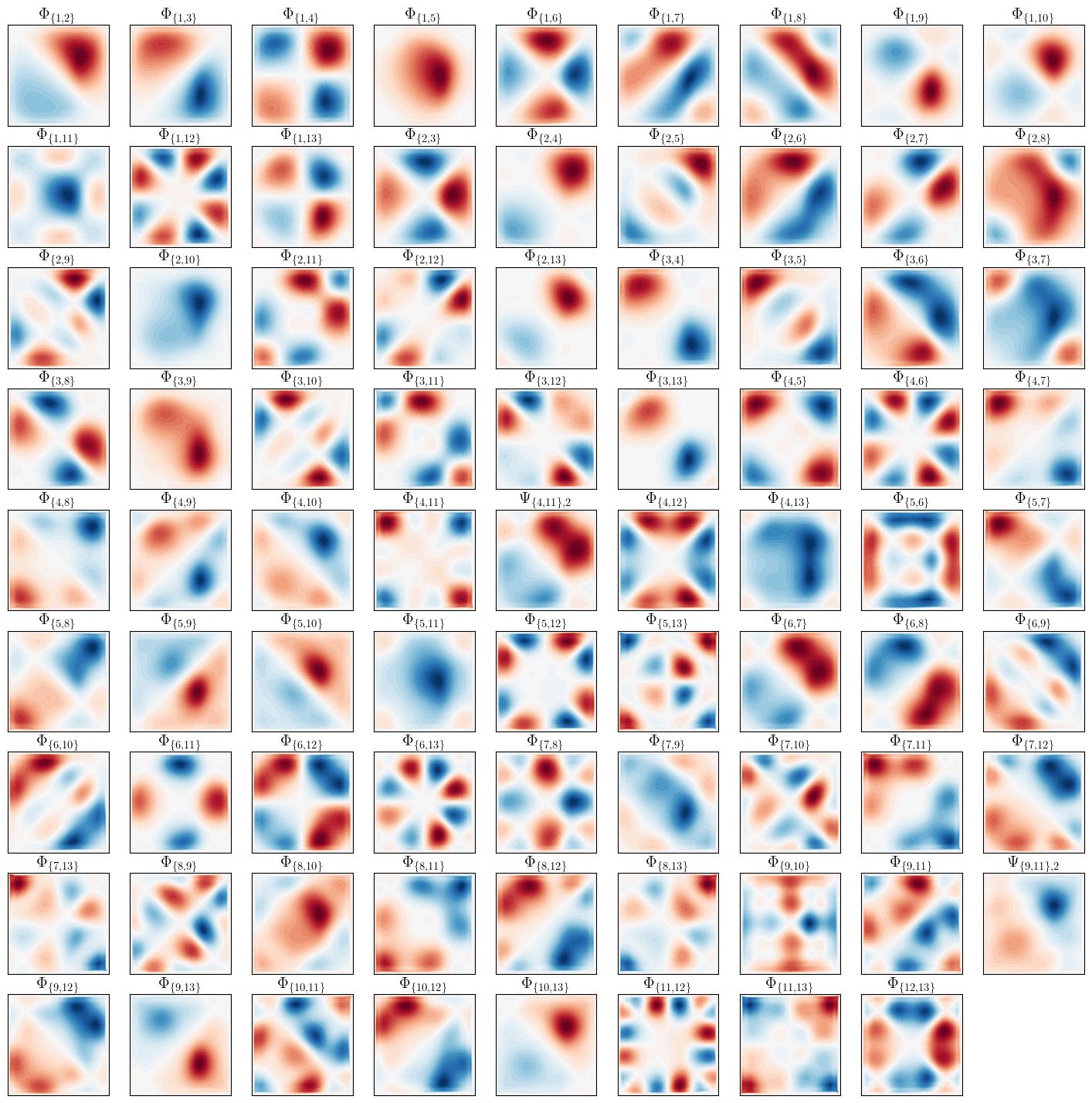}\\
  \makebox[\linewidth][c]{\small (b)} \\
  \vspace{0.2em}
  \caption{Two-dimensional nonlinear elliptic equation. Coefficient functions corresponding to the retained terms in the sNC expansion for (a) the constant and single-variable terms and (b) pairwise interaction terms.}
  \label{fig:2D13RV_coeff}
\end{figure}

Fig.~\ref{fig:2D13RV_coeff} shows the learned coefficient functions. The coefficient functions remain smooth and exhibit coherent spatial structure throughout the domain. The single-variable contributions shown in Fig.~\ref{fig:2D13RV_coeff}(a) are characterized by relatively simple spatial variation, whereas the pairwise interaction terms in Fig.~\ref{fig:2D13RV_coeff}(b) exhibit richer and localized spatial structures associated with nonlinear interaction effects.

\begin{figure}
  \center
  \includegraphics[width=\textwidth]{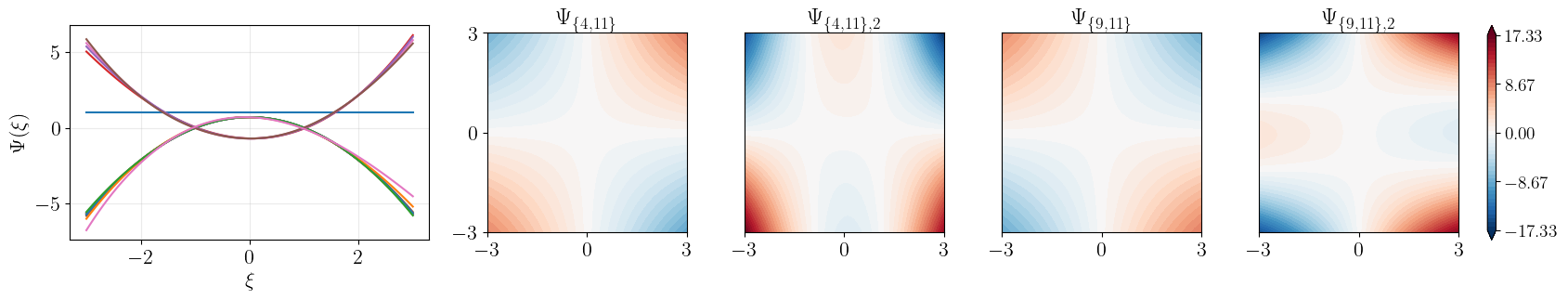}
    \makebox[\linewidth][l]{\hspace{0.9in} \small (a) \hspace{1.12in} (b) \hspace{0.82in} (c) \hspace{0.82in} (d) \hspace{0.82in} (e)} \\
  \vspace{0.2em}
  \caption{Two-dimensional nonlinear elliptic equation. (a) Basis functions corresponding to the retained single-variable terms in the sNC expansion. The single-variable basis functions exhibit similar structure and are shown without individual labels for clarity. (b)-(e) Basis functions associated with the retained pairwise interaction terms admitting a rank-2 approximation; second components are explicitly labeled.}
  \label{fig:2D13RV_basis}
\end{figure}

Fig.~\ref{fig:2D13RV_basis} shows the learned basis functions. In contrast to the previous example, the single-variable basis functions shown in Fig.~\ref{fig:2D13RV_basis}(a) exhibit nonlinear structure rather than simple linear trends, reflecting the increased complexity of the present problem and the stronger nonlinear interaction effects. Nevertheless, the learned basis functions remain smooth and low-dimensional, indicating that the proposed framework adaptively identifies the simplest functional representation required within each ANOVA subspace. The pairwise interaction basis functions similarly exhibit simple and structured behavior despite the increase complexity of the problem. Fig.~\ref{fig:2D13RV_basis}(b)-(e) shows four interaction basis functions associated with the two terms that admit a rank-2 approximation. These rank-1 components demonstrate the common structure observed across all interaction basis functions in the sNC expansion. As in the previous example, the interaction modes exhibit progressively richer stochastic structure, indicating that the proposed decomposition is capable of identifying distinct interaction patterns beyond the dominant rank-1 component, while remaining relatively simple. 

\subsubsection{Statistical and sensitivity estimates}
\begin{figure}
  \center
  \includegraphics[width=\textwidth]{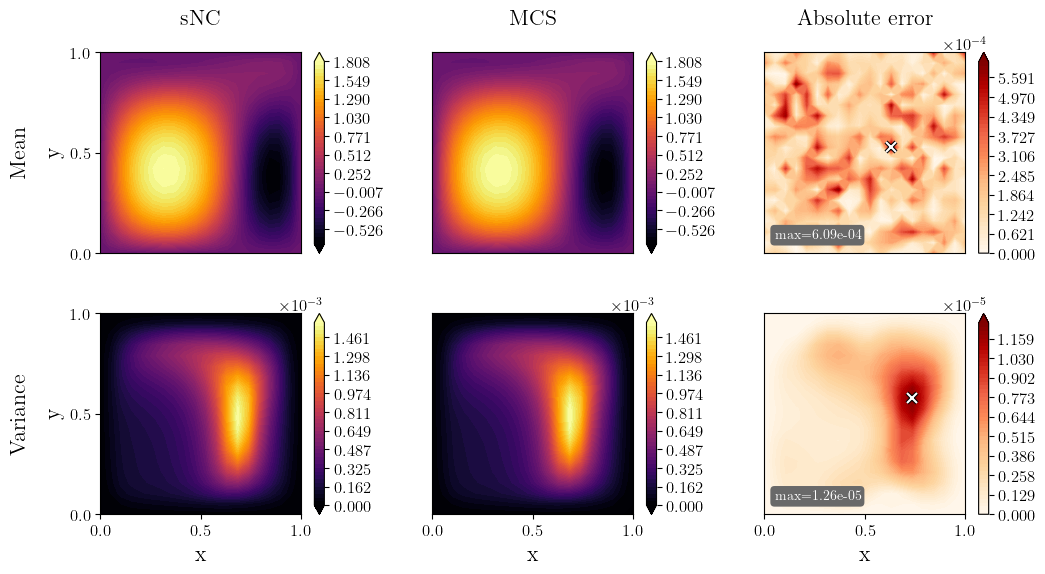}\\
  \vspace{0.2em}
  \caption{Two-dimensional nonlinear elliptic equation. Mean and variance fields computed directly from the coefficient functions of the sNC expansion, compared with reference values estimated via MCS using $2^{17}$ realizations. The corresponding absolute error fields are also shown, with the maximum absolute error across the spatial domain highlighted in each case.}
  \label{fig:2D13RV_mean_var}
\end{figure}

Fig.~\ref{fig:2D13RV_mean_var} compares the mean and variance fields computed directly from the learned coefficient functions of the sNC expansion with reference fields estimated via MCS with $2^{17}$ simulations; the corresponding absolute error fields with the maximum error highlighted are also shown. Excellent agreement between the sNC estimates and the MCS reference solutions is observed for both the mean and variance throughout the spatial domain. The error fields remain uniformly small, further demonstrating the ability of the proposed framework to accurately recover the dominant statistical structure of the stochastic solution field.

\begin{figure}
  \center
  \includegraphics[width=4.5in]{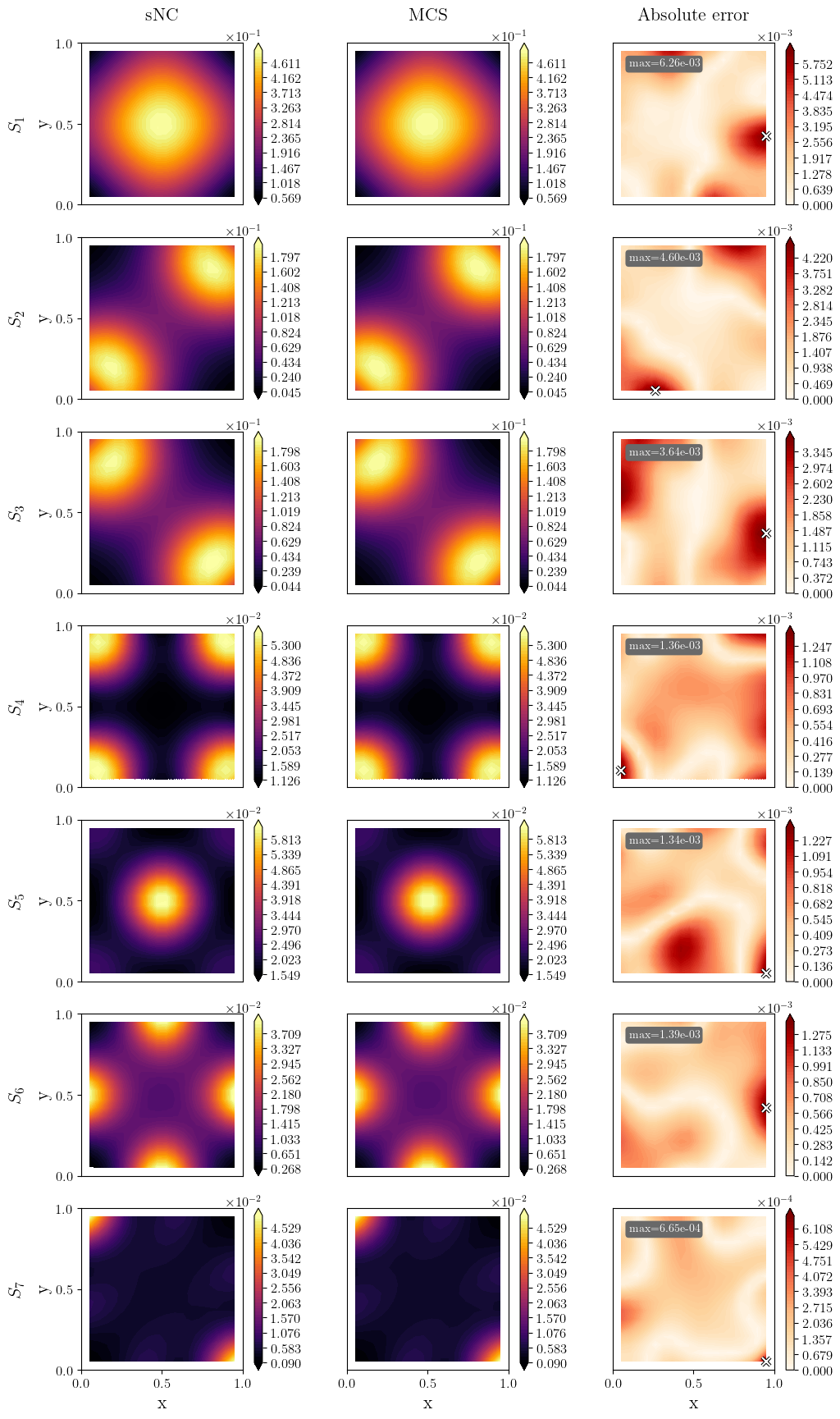}\\
  \vspace{0.2em}
  \caption{Two-dimensional nonlinear elliptic equation. First-order Sobol' indices $S_1,\dots,S_7$ computed directly from the coefficient functions of the sNC expansion, compared with reference values estimated via MCS using $2^{17}$ realizations. The corresponding absolute error fields are also shown, with the maximum absolute error across the spatial domain highlighted in each case.}
  \label{fig:2D13RV_GSA_first_part1}
\end{figure}

\begin{figure}
  \center
  \includegraphics[width=4.5in]{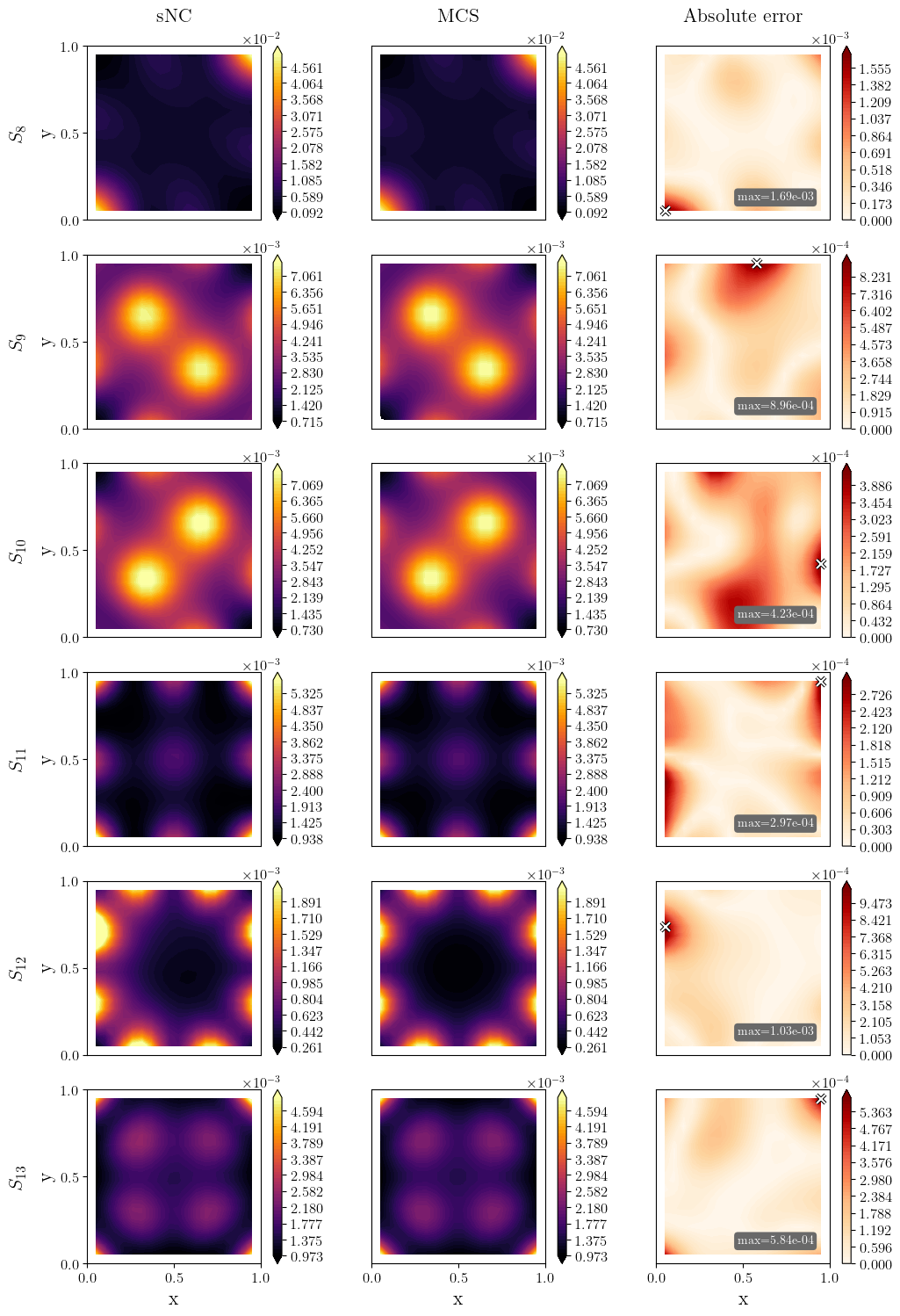}\\
  \vspace{0.2em}
  \caption{Two-dimensional nonlinear elliptic equation. First-order Sobol' indices $S_8,\dots,S_{13}$ computed directly from the coefficient functions of the sNC expansion, compared with reference values estimated via MCS using $2^{17}$ realizations. The corresponding absolute error fields are also shown, with the maximum absolute error across the spatial domain highlighted in each case.}
  \label{fig:2D13RV_GSA_first_part2}
\end{figure}

The first-order Sobol' indices computed directly from the coefficients of the sNC expansion are shown in Fig.~\ref{fig:2D13RV_GSA_first_part1} and Fig.~\ref{fig:2D13RV_GSA_first_part2}. The figures also show the reference MCS estimates and the corresponding absolute errors. Excellent agreement between the sNC and MCS estimates is observed across all Sobol' indices, indicating that the proposed framework accurately recovers the underlying global sensitivity structure throughout the spatial domain. The first-order Sobol' indices exhibit a clear hierarchy of influence, with groups of variables contributing at distinct orders of magnitude. The dominant indices are recovered with particularly high accuracy, while the weaker sensitivity contributions are also captured consistently despite their smaller magnitudes. The corresponding absolute error fields remain small throughout the spatial domain, with the maximum errors are generally localized near the domain boundaries where the total variance becomes small and the Sobol' indices are consequently more sensitive to localized approximation errors. A noticeable discrepancy is observed for $S_{12}$; however, the overall spatial structure and magnitude remain in good agreement with the MCS reference solution. 

\begin{figure}
  \center
  \includegraphics[width=4.5in]{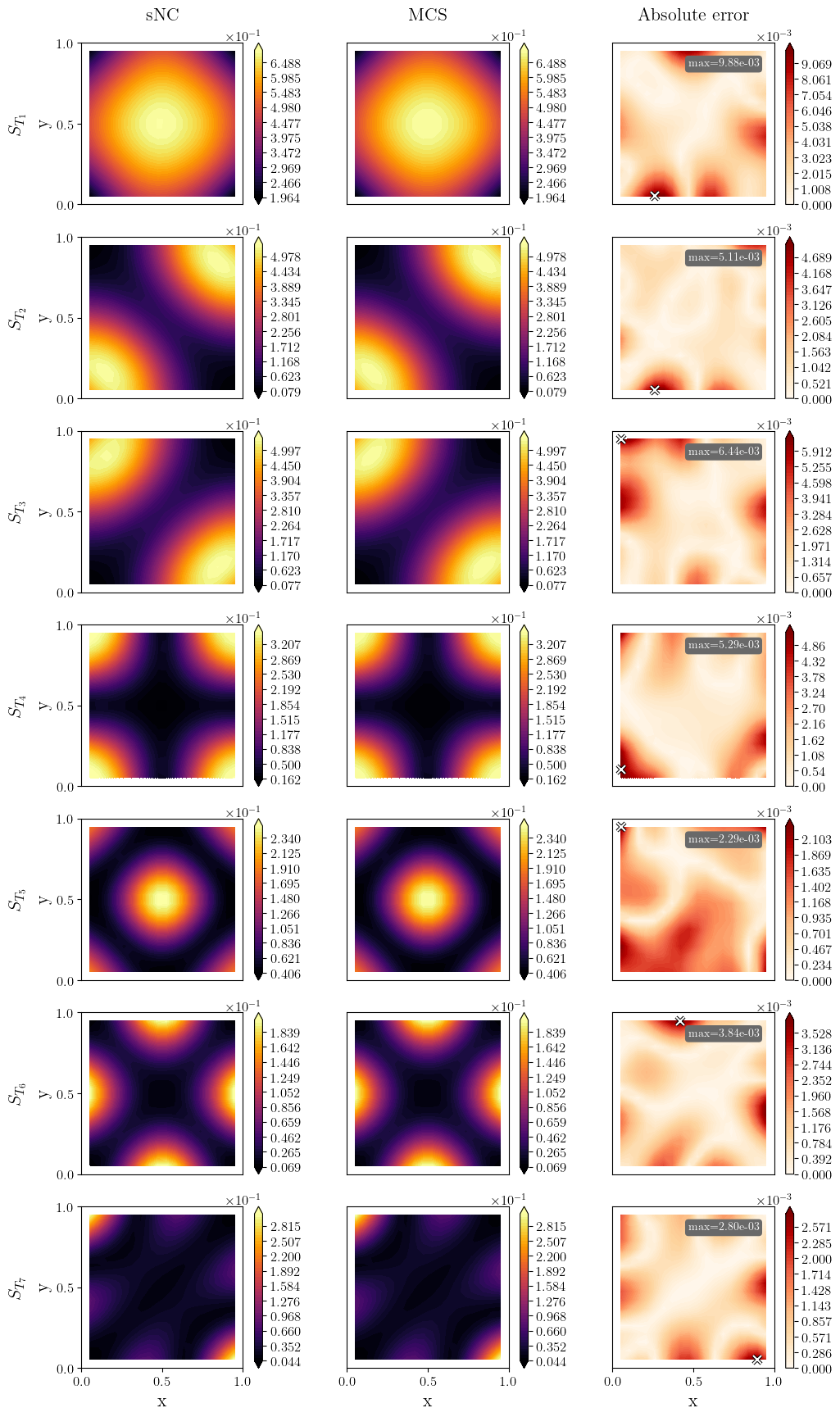}\\
  \vspace{0.2em}
  \caption{Two-dimensional nonlinear elliptic equation. Total-order Sobol' indices $S_{T_1},\dots,S_{T_7}$ computed directly from the coefficient functions of the sNC expansion, compared with reference values estimated via MCS using $2^{17}$ realizations. The corresponding absolute error fields are also shown, with the maximum absolute error across the spatial domain highlighted in each case.}
  \label{fig:2D13RV_GSA_total_part1}
\end{figure}

\begin{figure}
  \center
  \includegraphics[width=4.5in]{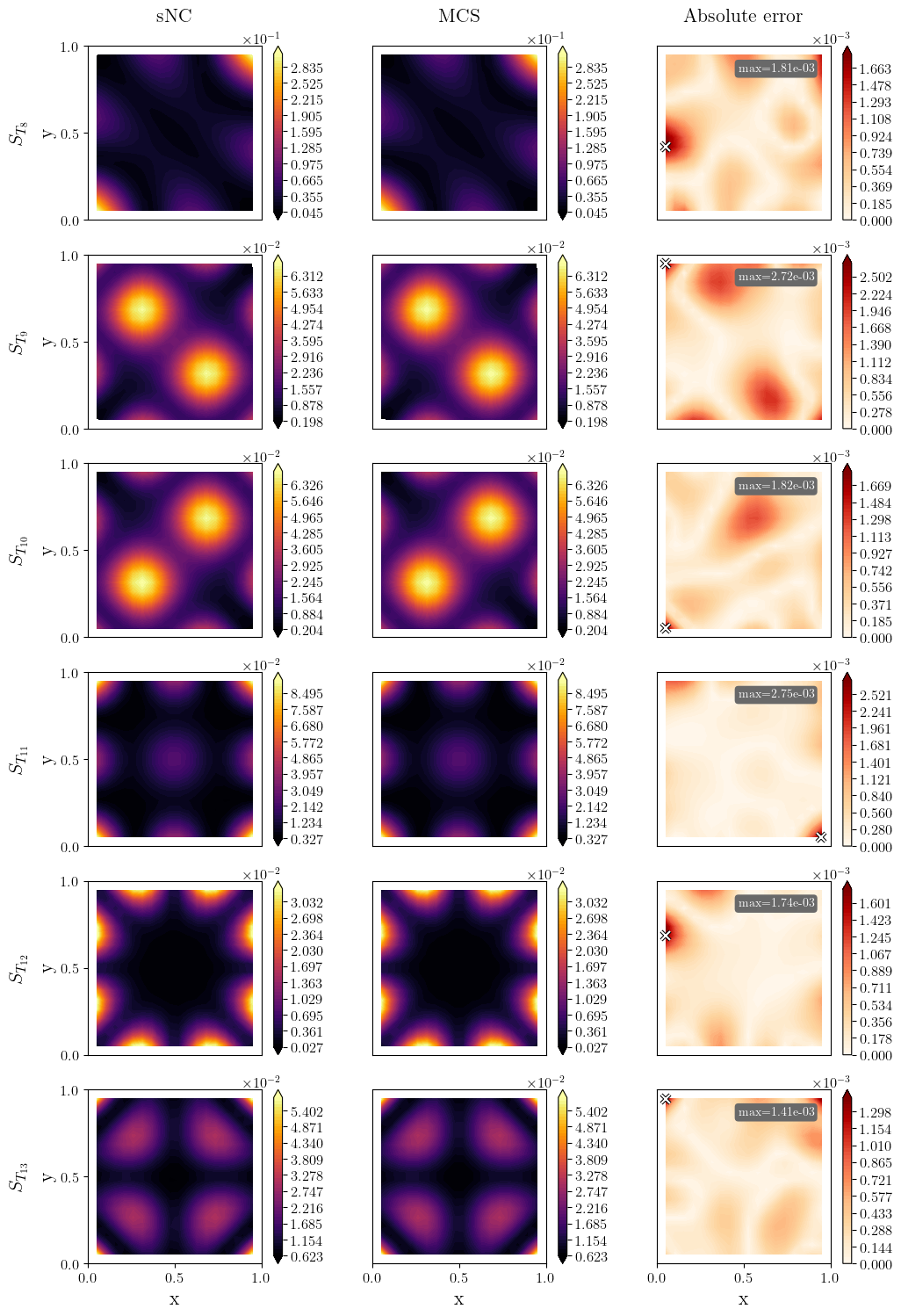}\\
  \vspace{0.2em}
  \caption{Two-dimensional nonlinear elliptic equation. Total-order Sobol' indices $S_{T_8},\dots,S_{T_{13}}$ computed directly from the coefficient functions of the sNC expansion, compared with reference values estimated via MCS using $2^{17}$ realizations. The corresponding absolute error fields are also shown, with the maximum absolute error across the spatial domain highlighted in each case.}
  \label{fig:2D13RV_GSA_total_part2}
\end{figure}

The total-order Sobol' indices computed directly from the coefficients of the sNC expansion are shown in Fig.~\ref{fig:2D13RV_GSA_total_part1} and Fig.~\ref{fig:2D13RV_GSA_total_part2}, together with the corresponding MCS estimates and absolute error fields. Overall, the behavior is similar to that observed for the first-order indices, with excellent agreement between the sNC and MCS estimates in both magnitude and spatial structure across all variables. The consistently larger magnitudes of the total-order indices compared with the corresponding first-order indices clearly demonstrate the presence of strong interaction effects in the present problem. The corresponding absolute error fields remain uniformly small throughout the spatial domain, with the maximum absolute errors consistently occurring near the domain boundaries where the total variance becomes small and the sensitivity indices become increasingly difficult to resolve accurately.

Collectively, the results of this example demonstrate that the proposed framework remains capable of accurately recovering the statistical and sensitivity structure of high-dimensional nonlinear stochastic systems with strong interaction effects. Despite the increased complexity of the present problem and the large number of retained interaction terms, the learned decomposition remains well-structured and low-rank, while maintaining excellent agreement with the reference MCS estimates across the mean, variance, and Sobol' index fields. The larger relative error (see Table~\ref{tab:summary_results2}) associated with the first-order indices is consistent with the stronger interaction effects present in the problem, where isolating purely first-order contributions becomes increasingly challenging.

\subsection{Example 4: One-dimensional advection-diffusion equation with 6 random variables}
\label{subsec:1D6RV_adv_diff}

In this example, we consider a time-dependent advection-diffusion equation in a one-dimensional spatial domain with a spatially varying advection velocity. The problem follows the setup in \cite{sharma2026polynomial}. The governing equation is given by
\begin{align}
    \frac{\partial s(x,t)}{\partial t} + v(x)\frac{\partial s(x,t)}{\partial x} = D\frac{\partial^2 s(x,t)}{\partial x^2}, \quad x\in[0,1], \quad t\in[0,1],
\end{align}
with boundary conditions
\begin{align}
    s(0,t)=s(1,t)=0,
\end{align}
and initial condition $s(x,0)=\sin(\pi x)$. Here, $s(x,t)$ denotes the transported scalar field, and the diffusion coefficient is set to $D=0.1$. The advection velocity $v(x)$ is modeled as a one-dimensional Gaussian random field with mean $\mu_v=1$ and covariance kernel
\begin{align}
    \text{Cov}(x_1, x_2) = \sigma^2\exp\left(-\frac{(x_1-x_2)^2}{2l_c^2}\right),
\end{align}
where the correlation length is $l_c=0.2$ and the standard deviation is $\sigma=0.05$. The random field is approximated using a truncated KL expansion with six modes, resulting in a six-dimensional stochastic problem parameterized by independent standard Gaussian random variables $\{\xi_i\}_{i=1}^6$. This problem provides a suitable benchmark for assessing the ability of the proposed framework to recover evolving statistical and sensitivity structure in time-dependent stochastic systems. A total of 4096 realizations of the random field are generated and the governing equation is solved numerically on a uniform $50\times 50$ spatial grid.

\subsubsection{Surrogate accuracy and structure}
\begin{figure}
  \center
  \includegraphics[width=4.5in]{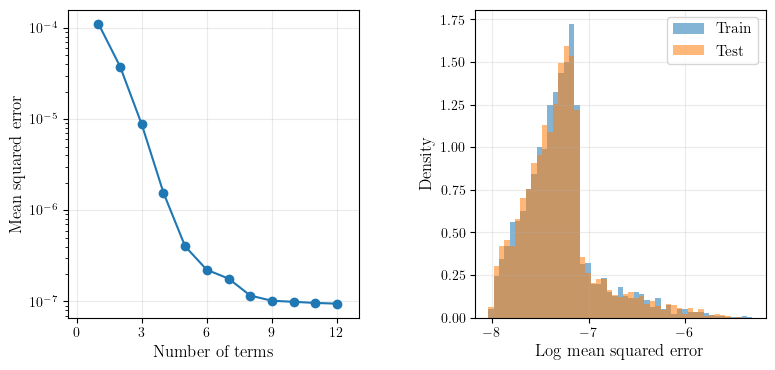} \\
  \makebox[\linewidth][c]{\small (a) \hspace{2in} (b)} \\
  \vspace{1em}
  \includegraphics[width=\textwidth]{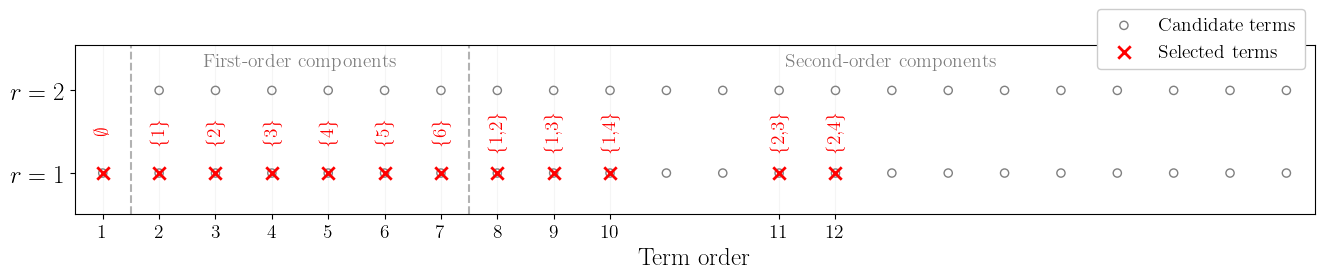}
  \makebox[\linewidth][c]{\small (c)}\\
  \vspace{0.2em}
  \caption{One-dimensional advection-diffusion equation. (a)~Mean squared error of the sNC expansion with respect to the number of retained terms. (b)~Distribution of training and testing errors; the training and testing sets consist of 4096 and 5000 stochastic samples, respectively. (c)~Candidate terms grouped by interaction structure, together with the associated approximation rank $r$. The index sets of the selected terms in the final expansion are highlighted in red.}
  \label{fig:1D6RV_adv_diff_errors}
\end{figure}

The mean squared error of the sNC expansion with respect to the number of retained terms is shown in Fig.~\ref{fig:1D6RV_adv_diff_errors}(a), where the error decreases rapidly as additional terms are incorporated before reaching a plateau. The error distribution between the model predictions and the ground truth is shown in Fig.~\ref{fig:1D6RV_adv_diff_errors}(b). The close agreement between the training and testing distribution indicates strong generalization of the learned sNC expansion. Fig.~\ref{fig:1D6RV_adv_diff_errors}(c) shows the retained terms in the final sNC expansion highlighted in red; all retained terms adopt rank-1 approximations. According to Eq.~\ref{eq:n}, a full rank-$r$ approximation would contain $1+21r$ terms, such that the final sNC expansion containing $n^*=12$ terms remains comparatively sparse.

\begin{figure}
  \centering
  \includegraphics[width=\textwidth]{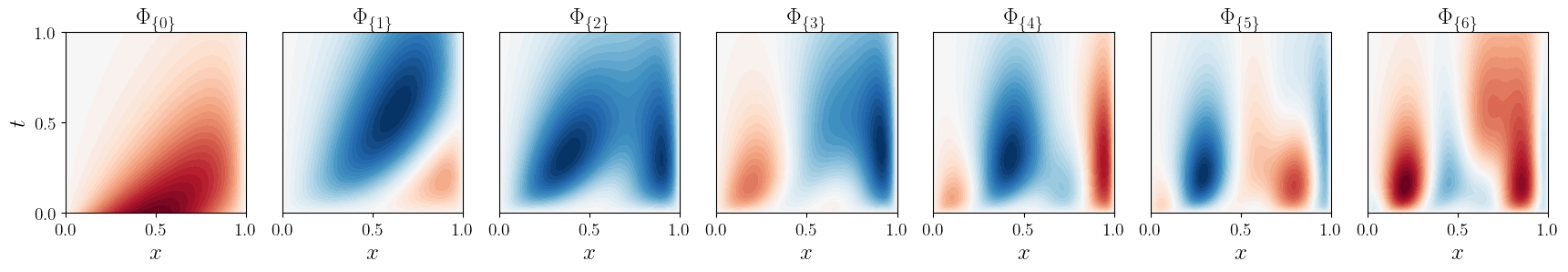}\\
  \makebox[\linewidth][c]{\small (a)} \\
  \vspace{1em}
  \phantom{xxx.}\includegraphics[width=6.18in]{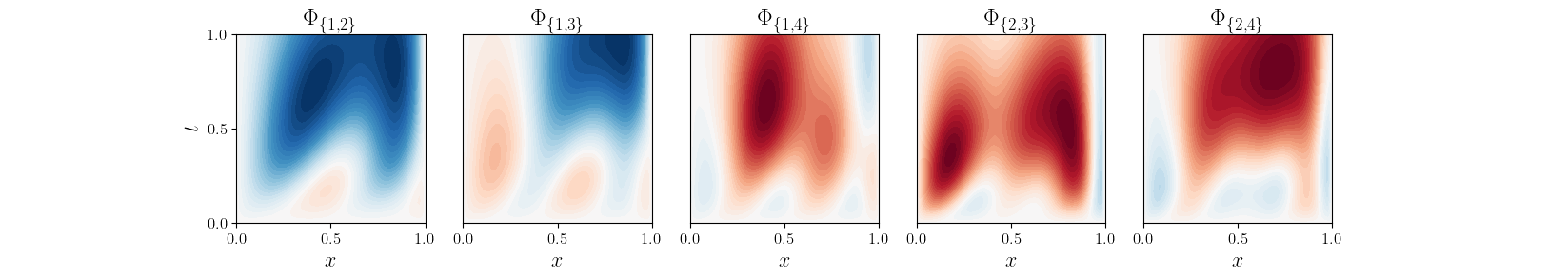}\\
  \makebox[\linewidth][c]{\small (b)} \\
  \vspace{0.2em}
  \caption{One-dimensional advection-diffusion equation. Coefficient functions corresponding to the retained terms in the sNC expansion for (a) the constant and single-variable terms and (b) pairwise interaction terms.}
  \label{fig:1D6RV_adv_diff_coeff}
\end{figure}

Fig.~\ref{fig:1D6RV_adv_diff_coeff} shows the learned coefficient functions. The coefficient functions exhibit smooth and highly structured behavior across the spatial-temporal domain. The dominant single-variable coefficient functions display progressively richer spatial variation for higher-index variables while remaining temporally coherent throughout the evolution. The retained pairwise interaction coefficient functions similarly exhibit organized low-complexity structure, indicating that the interaction effects remain well-resolved despite the increased complexity introduced by the time-dependent dynamics.

\begin{figure}
  \center
  \includegraphics[width=\textwidth]{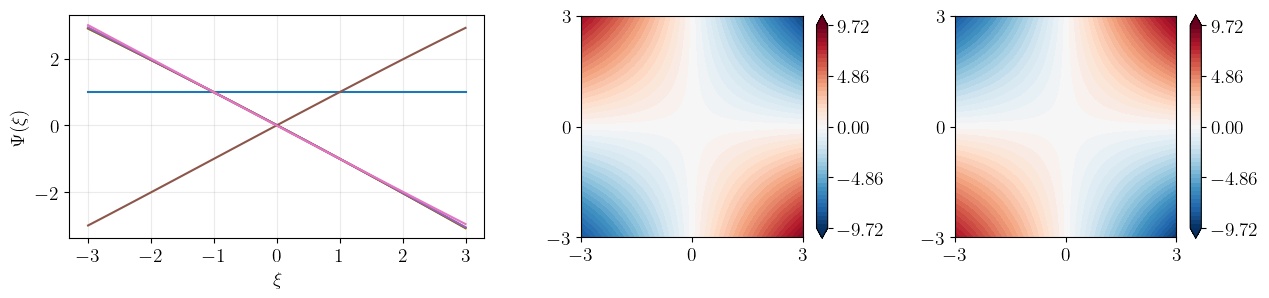}
    \makebox[\linewidth][l]{\hspace{1.25in} \small (a) \hspace{1.9in} (b) \hspace{1.55in} (c) } \\
  \vspace{0.2em}
  \caption{One-dimensional advection-diffusion equation. (a) Basis functions corresponding to the retained single-variable terms in the sNC expansion. The single-variable basis functions exhibit similar structure and are shown without individual labels for clarity. (b)-(c) Representative basis functions associated with the retained pairwise interaction terms, illustrating the common structure observed across all the interaction basis functions.}
  \label{fig:1D6RV_adv_diff_basis}
\end{figure}

Fig.~\ref{fig:1D6RV_adv_diff_basis} shows the learned basis functions. The retained single-variable basis functions shown in Fig.~\ref{fig:1D6RV_adv_diff_basis}(a) all exhibit simple linear structure, indicating that the dominant stochastic behavior within each Sobol space is captured using low-complexity representations. The pairwise interaction basis functions similarly exhibit simple and highly structured behavior. Two representative interaction basis functions are shown in Fig.~\ref{fig:1D6RV_adv_diff_basis}(b)-(c) to illustrate the common bilinear structure observed across all the retained interaction basis functions.

\subsubsection{Statistical and sensitivity estimates}
\begin{figure}
  \center
  \includegraphics[width=\textwidth]{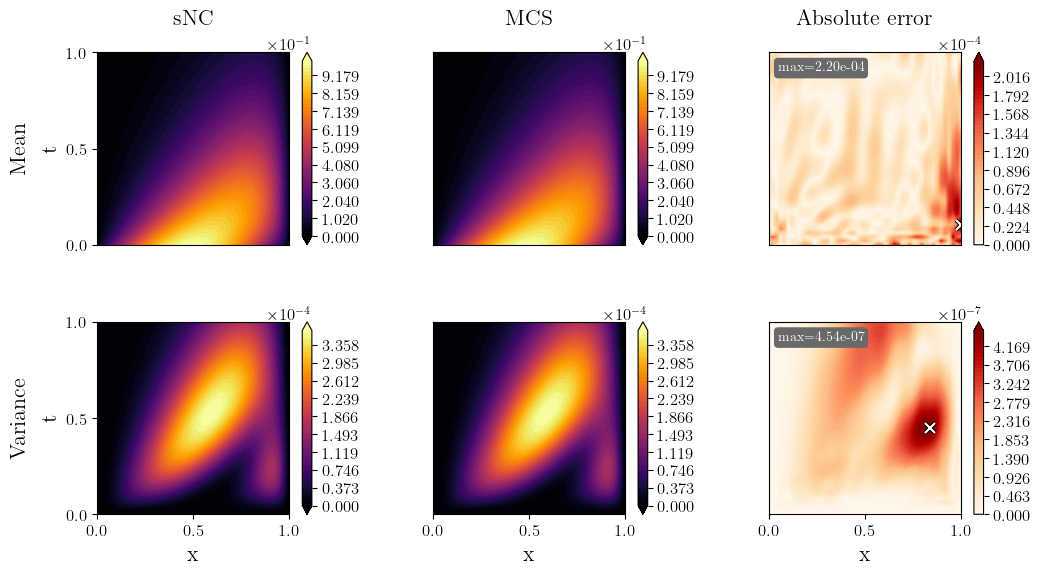}\\
  \vspace{0.2em}
  \caption{One-dimensional advection-diffusion equation. Mean and variance fields computed directly from the coefficient functions of the sNC expansion, compared with reference values estimated via MCS using $2^{18}$ realizations. The corresponding absolute error fields are also shown, with the maximum absolute error across the spatial-temporal domain highlighted in each case.}
  \label{fig:1D6RV_adv_diff_mean_var}
\end{figure}

Fig.~\ref{fig:1D6RV_adv_diff_mean_var} compares the mean and variance fields computed directly from the learned coefficient functions of the sNC expansion with reference fields estimated via MCS with $2^{18}$ simulations; the corresponding absolute error fields with the maximum error highlighted are also shown. Excellent agreement is observed between the sNC estimates and the MCS reference solutions for both the mean and variance throughout the spatial-temporal domain. The error fields remain uniformly small, with maximum absolute errors $2.31\times10^{-4}$ and $4.56\times 10^{-7}$ for the mean and variance, respectively, corresponding to errors that are several orders of magnitude smaller than the values of the underlying field. The relative $L^2$ errors are $1.04\times10^{-4}$ for the mean and $9.23\times 10^{-4}$ for the variance, further demonstrating the accuracy of the recovered statistical structure.

\begin{figure}
  \center
  \includegraphics[width=4.5in]{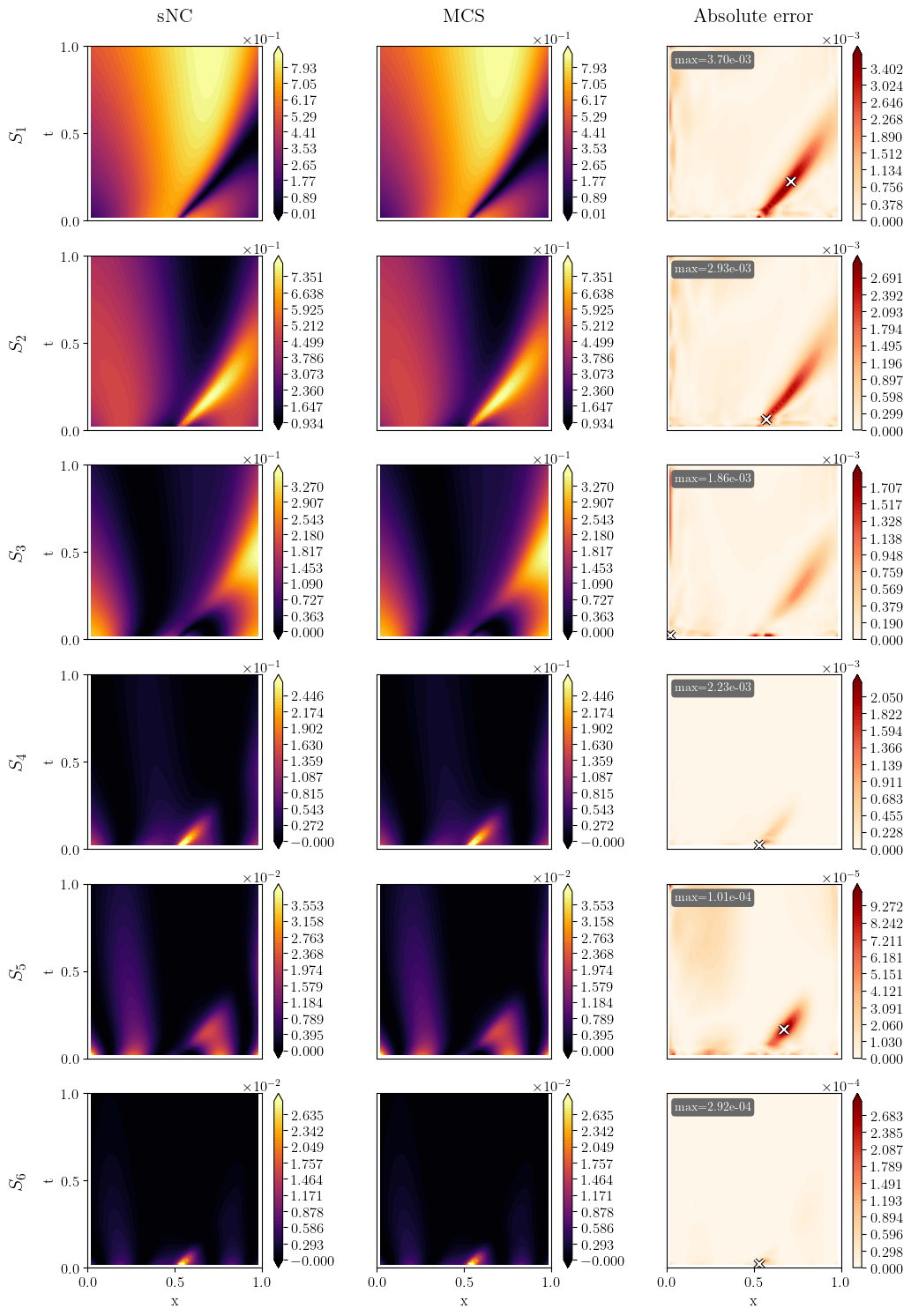}\\
  \vspace{0.2em}
  \caption{One-dimensional advection-diffusion equation. First-order Sobol' indices computed directly from the coefficient functions of the sNC expansion, compared with reference values estimated via MCS using $2^{18}$ realizations. The corresponding absolute error fields are also shown, with the maximum absolute error across the spatial domain highlighted in each case.}
  \label{fig:1D6RV_adv_diff_GSA_first}
\end{figure}

The first-order Sobol' indices computed directly from the coefficients of the sNC expansion are shown in Fig.~\ref{fig:1D6RV_adv_diff_GSA_first}. The figures also show the reference MCS estimates and the corresponding absolute errors. Excellent agreement between the sNC and MCS estimates is observed across all first-order indices, indicating that the proposed framework accurately recovers both the magnitude and spatial-temporal structure of the dominant sensitivity contributions. The first-order indices exhibit increasingly localized regions of influence throughout the spatial-temporal domain, while the overall magnitude of the sensitivity contributions decreases progressively for higher-index variables. The corresponding absolute error fields remain uniformly small, with the largest discrepancies localized near the sharply varying sensitivity features.

\begin{figure}
  \center
  \includegraphics[width=4.5in]{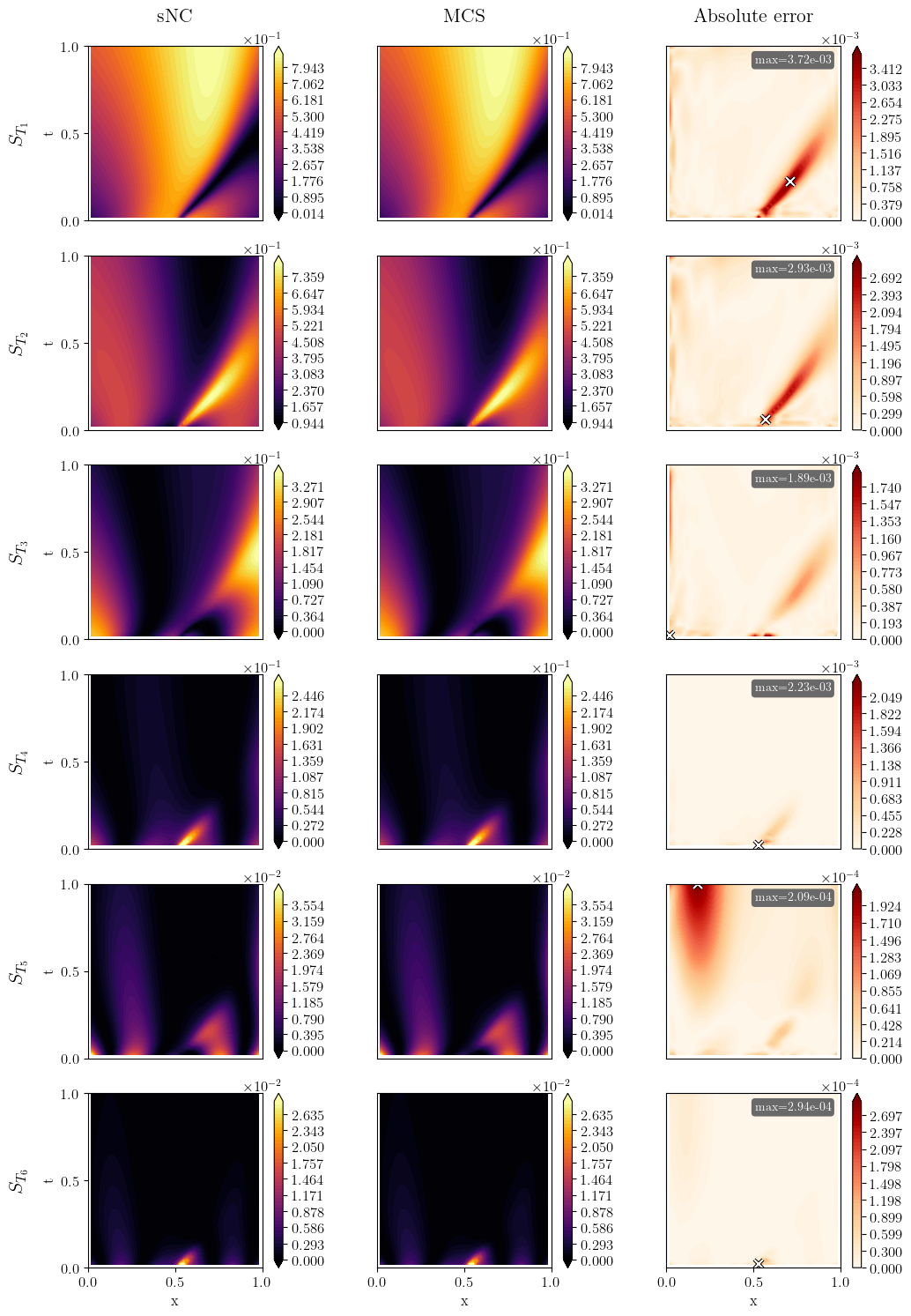}\\
  \vspace{0.2em}
  \caption{One-dimensional advection-diffusion equation. Total-order Sobol' indices computed directly from the coefficient functions of the sNC expansion, compared with reference values estimated via MCS using $2^{18}$ realizations. The corresponding absolute error fields are also shown, with the maximum absolute error across the spatial domain highlighted in each case.}
  \label{fig:1D6RV_adv_diff_GSA_total}
\end{figure}

The total-order Sobol' indices computed directly from the coefficients of the sNC expansion are shown in Fig.~\ref{fig:1D6RV_adv_diff_GSA_total}. Excellent agreement between the sNC and MCS estimates is observed across all total-order indices. In contrast to the previous example, the total-order indices remain very similar to their corresponding first-order indices, indicating that higher-order interaction effects are comparatively weak in the present problem and consistent with the sparse interaction structure identified by the adaptive selection procedure. The corresponding absolute error fields remain uniformly small throughout the spatial-temporal domain as with the first-order indices. The nearly identical error fields obtained for the first-order and total-order indices are consistent with the comparatively weak higher-order interaction effects present in the problem.

\subsection{Example 5: One-dimensional Burgers’ equation with 6 random variables}
\label{subsec:1D6RV_Burgers}

The final example considers a one-dimensional viscous Burgers' equation with spatially varying source term. The problem follows the setup in \cite{sharma2026polynomial}, where the source term is squared to introduce stronger interaction effects while preserving the underlying physical structure of the problem. The governing equation is given by
\begin{align}
    \frac{\partial s(x,t)}{\partial t} + s(x,t)\frac{\partial s(x,t)}{\partial x} - \nu\frac{\partial^2 s(x,t)}{\partial x^2}=f(x)^2, \quad x\in[0,1], \quad t\in[0,0.3],
\end{align}
with boundary conditions
\begin{align}
    s(0,t)=s(1,t)=0,
\end{align}
and initial condition $s(x,0)=\sin(\pi x)$. Here, $s(x,t)$ represents the state (or velocity field) of the system, and the viscosity is set to $\nu=0.001$. Burgers' equation is well known for developing sharp nonlinear solution features due to the interplay between nonlinear advection and diffusion, making it a challenging benchmark for stochastic approximation methods. The source term $f(x)$ is modeled as a one-dimensional zero-mean Gaussian random field with covariance kernel
\begin{align}
    \text{Cov}(x_1, x_2) = \sigma^2\exp\left(-\frac{(x_1-x_2)^2}{2l_c^2}\right),
\end{align}
where the correlation length is $l_c=0.2$ and the standard deviation is $\sigma=0.1$. The random field is approximated using a truncated KL expansion with six modes, resulting in a six-dimensional stochastic problem parameterized by independent standard Gaussian random variables $\{\xi_i\}_{i=1}^6$. A total of 8192 realizations of the random field are generated and the governing equation is solved numerically on a uniform $50\times 50$ space-time grid.

\subsubsection{Surrogate accuracy and structure}
\begin{figure}
  \center
  \includegraphics[width=4.5in]{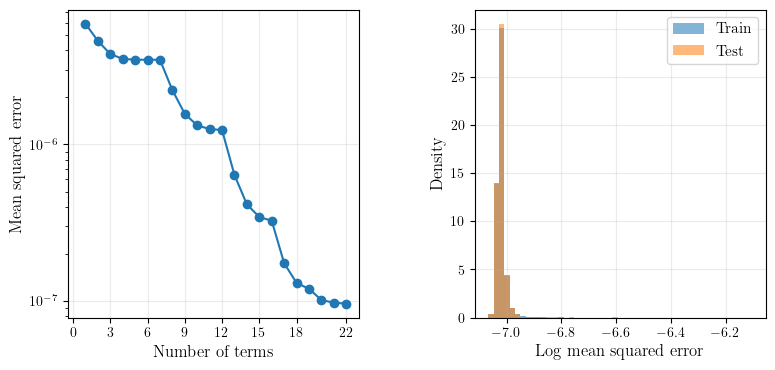} \\
  \makebox[\linewidth][c]{\small (a) \hspace{2in} (b)} \\
  \vspace{1em}
  \includegraphics[width=\textwidth]{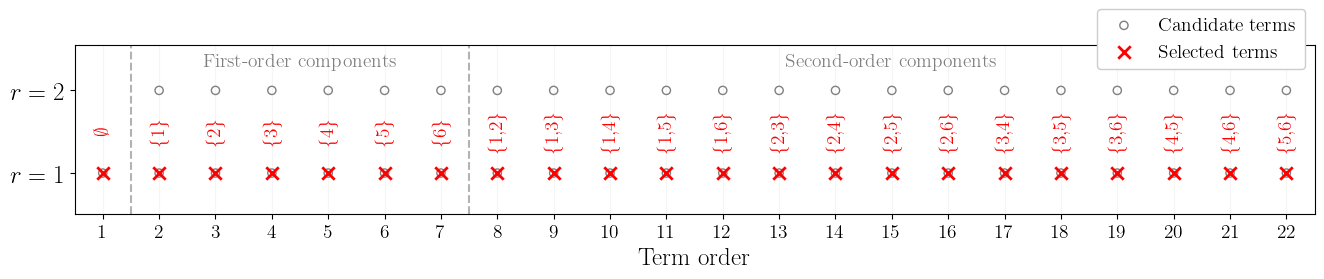}
  \makebox[\linewidth][c]{\small (c)}\\
  \vspace{0.2em}
  \caption{One-dimensional Burgers’ equation. (a)~Mean squared error of the sNC expansion with respect to the number of retained terms. (b)~Distribution of training and testing errors; the training and testing sets consist of 8192 and 5000 stochastic samples, respectively. (c)~Candidate terms grouped by interaction structure, together with the associated approximation rank $r$. The index sets of the selected terms in the final expansion are highlighted in red.}
  \label{fig:1D6RV_burgers_errors}
\end{figure}

The mean squared error of the sNC expansion with respect to the number of retained terms is shown in Fig.~\ref{fig:1D6RV_burgers_errors}(a), where the error decreases progressively as additional terms are incorporated into the expansion. The error distribution between the model predictions and the ground truth is shown in Fig.~\ref{fig:1D6RV_burgers_errors}(b). The error distributions remain sharply concentrated around a narrow range of low error values, with only a very small tail corresponding to a limited number of more challenging realizations, indicating highly consistent predictive accuracy across stochastic realizations. The close overlap between the training and testing distributions further demonstrates strong generalization of the learned expansion. Fig.~\ref{fig:1D6RV_burgers_errors}(c) shows the retained terms in the final sNC expansion highlighted in red. All candidate terms are retained in the final expansion, hinting at the increased complexity of this problem compared to the previous example. Each term adopts only a rank-1 approximation, resulting in a total of $n^*=22$ retained terms.  

\begin{figure}
  \center
  \includegraphics[width=\textwidth]{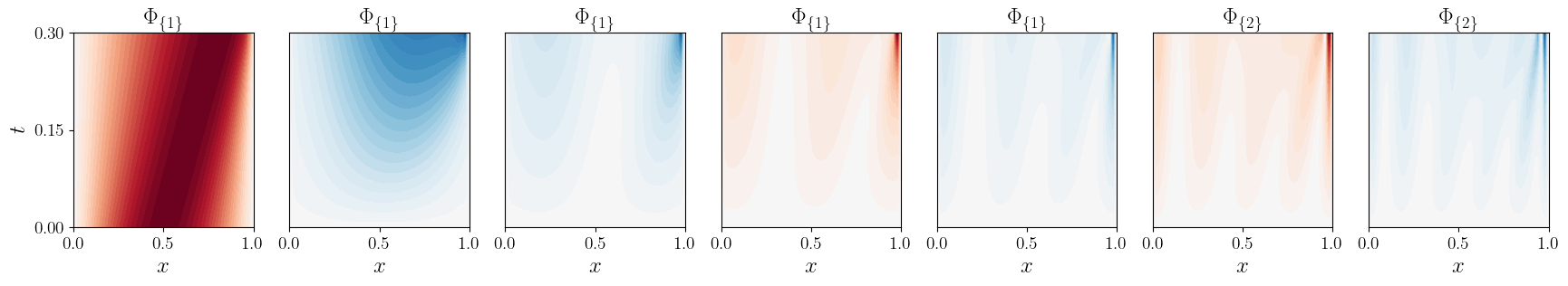}\\
  \makebox[\linewidth][c]{\small (a)} \\
  \vspace{1em}
  \phantom{xxx.}\includegraphics[width=6.18in]{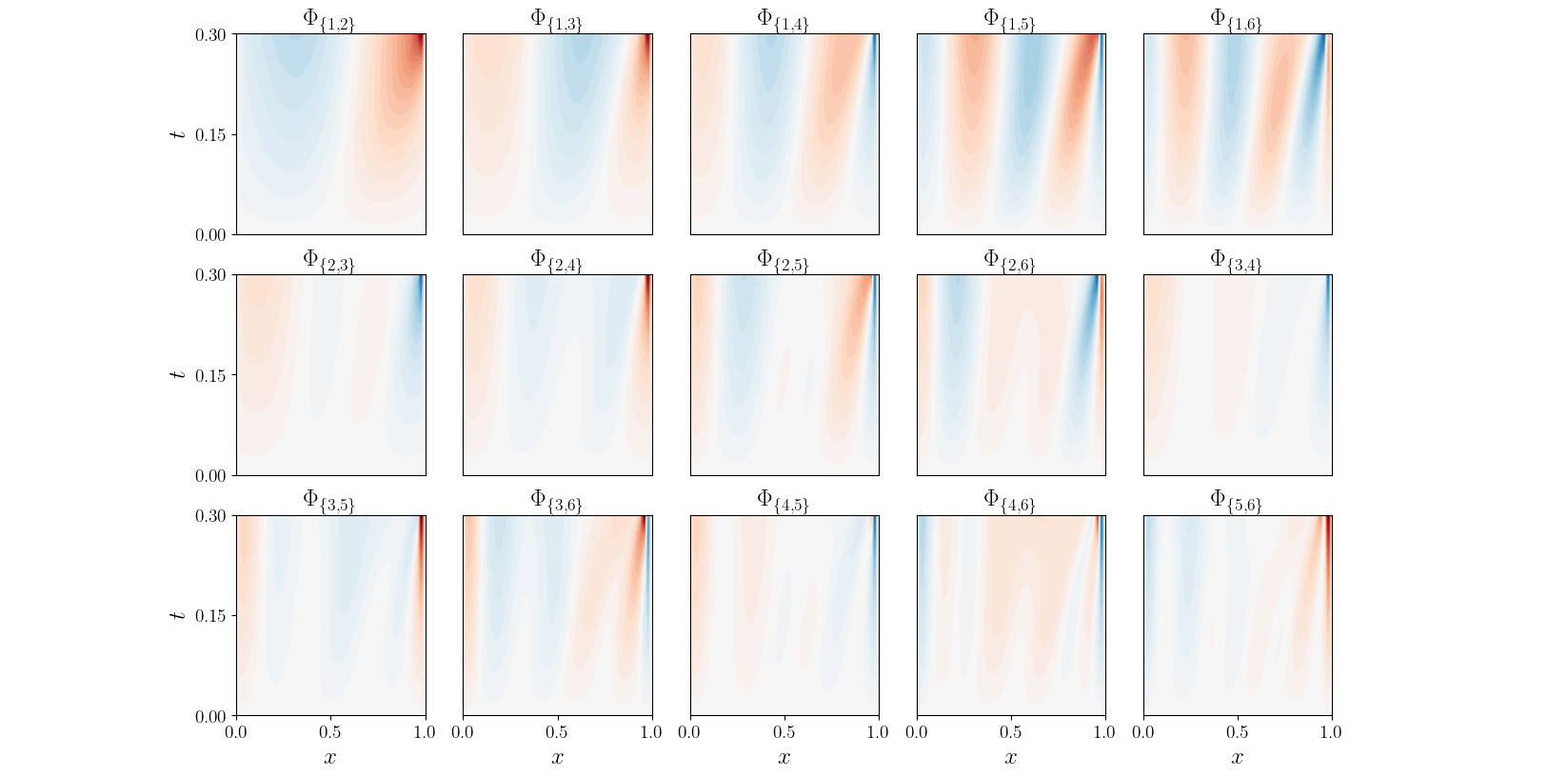}\\
  \makebox[\linewidth][c]{\small (b)} \\
  \vspace{0.2em}
  \caption{One-dimensional Burgers’ equation. Coefficient functions corresponding to the retained terms in the sNC expansion for (a) the constant and single-variable terms and (b) pairwise interaction terms.}
  \label{fig:1D6RV_burgers_coeff}
\end{figure}

Fig.~\ref{fig:1D6RV_burgers_coeff} shows the learned coefficient functions. In comparison with the previous examples, the coefficient functions exhibit increasingly localized and sharply varying spatial-temporal structure, reflecting the stronger nonlinear dynamics and enhanced interaction effects present in the Burgers' equation with squared stochastic forcing. The higher-index single-variable coefficient functions display progressively more concentrated features near the right boundary of the spatial domain, while the retained pairwise interaction coefficient functions exhibit similarly structured localized behavior. Despite the increased complexity of the problem, the learned coefficient functions remain smooth and highly organized, indicating that the proposed framework continues to recover interpretable low-complexity stochastic structure even in strongly nonlinear regimes.

\begin{figure}
  \center
  \includegraphics[width=\textwidth]{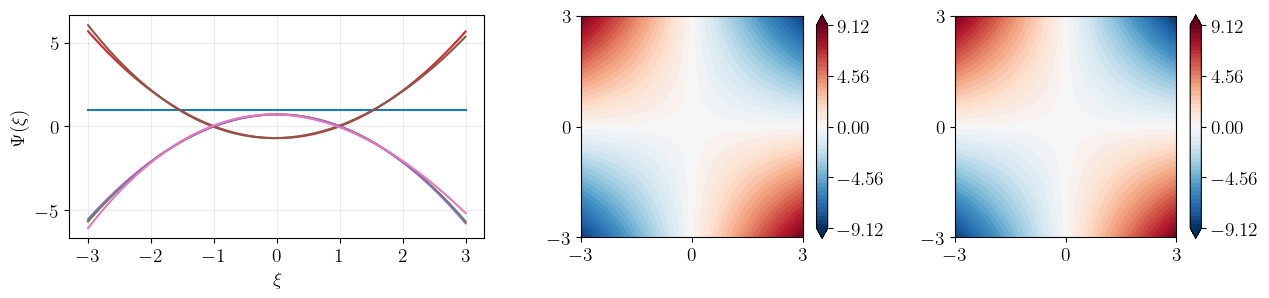}
    \makebox[\linewidth][l]{\hspace{1.25in} \small (a) \hspace{1.9in} (b) \hspace{1.55in} (c) } \\
  \vspace{0.2em}
  \caption{One-dimensional Burgers’ equation. (a) Basis functions corresponding to the retained single-variable terms in the sNC expansion. The single-variable basis functions exhibit similar structure and are shown without individual labels for clarity. (b)-(c) Representative basis functions associated with the retained pairwise interaction terms, illustrating the common structure observed across all the interaction basis functions.}
  \label{fig:1D6RV_burgers_basis}
\end{figure}

Fig.~\ref{fig:1D6RV_burgers_basis} shows the learned basis functions. Unlike the previous example, the retained single-variable basis functions shown in Fig.~\ref{fig:1D6RV_burgers_basis}(a) all exhibit simple quadratic structure, reflecting the increased complexity of the present problem and the stronger nonlinear interaction effects. This behavior mirrors the results observed in the two-dimensional nonlinear elliptic equation in Section \ref{subsec:2D13RV}, where the underlying equations were highly nonlinear. The pairwise interaction basis functions similarly exhibit simple and structured behavior despite the increase complexity of the problem. Two representative interaction basis functions are shown in Fig.~\ref{fig:1D6RV_burgers_basis}(b)-(c), demonstrating the common structure observed across the interaction basis functions for this example. In fact, similar structures were observed for the pairwise interaction rank-1 approximations throughout all the examples in this paper.

\subsubsection{Statistical and sensitivity estimates}
\begin{figure}
  \center
  \includegraphics[width=\textwidth]{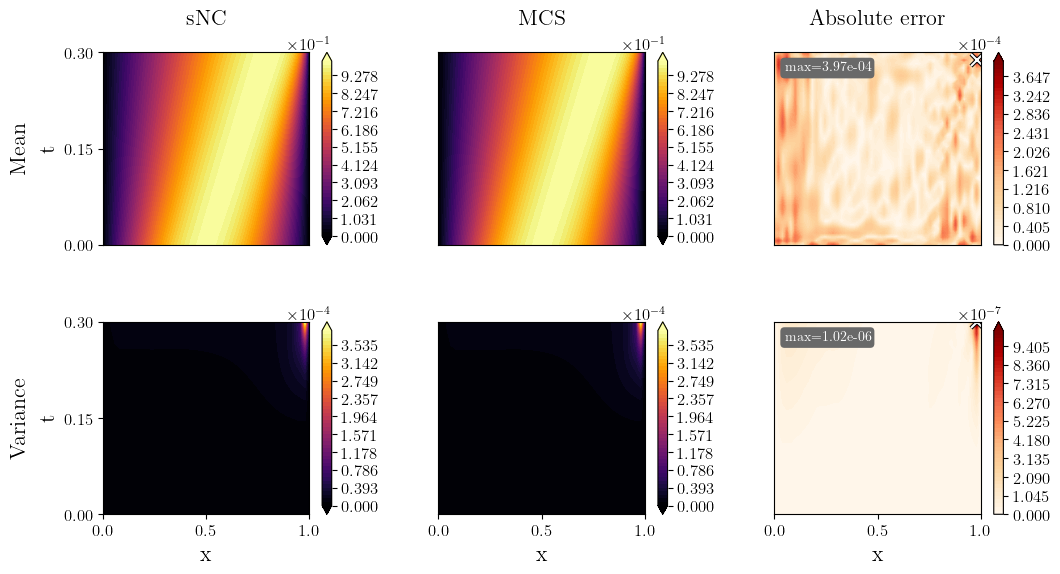}\\
  \vspace{0.2em}
  \caption{One-dimensional Burgers’ equation. Mean and variance fields computed directly from the coefficient functions of the sNC expansion, compared with reference values estimated via MCS using $2^{18}$ realizations. The corresponding absolute error fields are also shown, with the maximum absolute error across the spatial-temporal domain highlighted in each case.}
  \label{fig:1D6RV_burgers_mean_var}
\end{figure}

Fig.~\ref{fig:1D6RV_burgers_mean_var} compares the mean and variance fields computed directly from the learned coefficient functions of the sNC expansion with reference fields estimated via MCS with $2^{18}$ simulations; the corresponding absolute error fields with the maximum error highlighted are also shown. Excellent agreement between the sNC estimates and the MCS reference solutions is observed for both the mean and variance throughout the spatial-temporal domain. The variance field exhibits strong localized uncertainty concentrated near the right boundary at later times, reflecting the localized stochastic response induced by the nonlinear Burgers' dynamics and squared stochastic forcing. Despite the small magnitude and highly localized nature of the variance field, the proposed framework accurately captures both its magnitude and spatial-temporal structure.

\begin{figure}
  \center
  \includegraphics[width=4.5in]{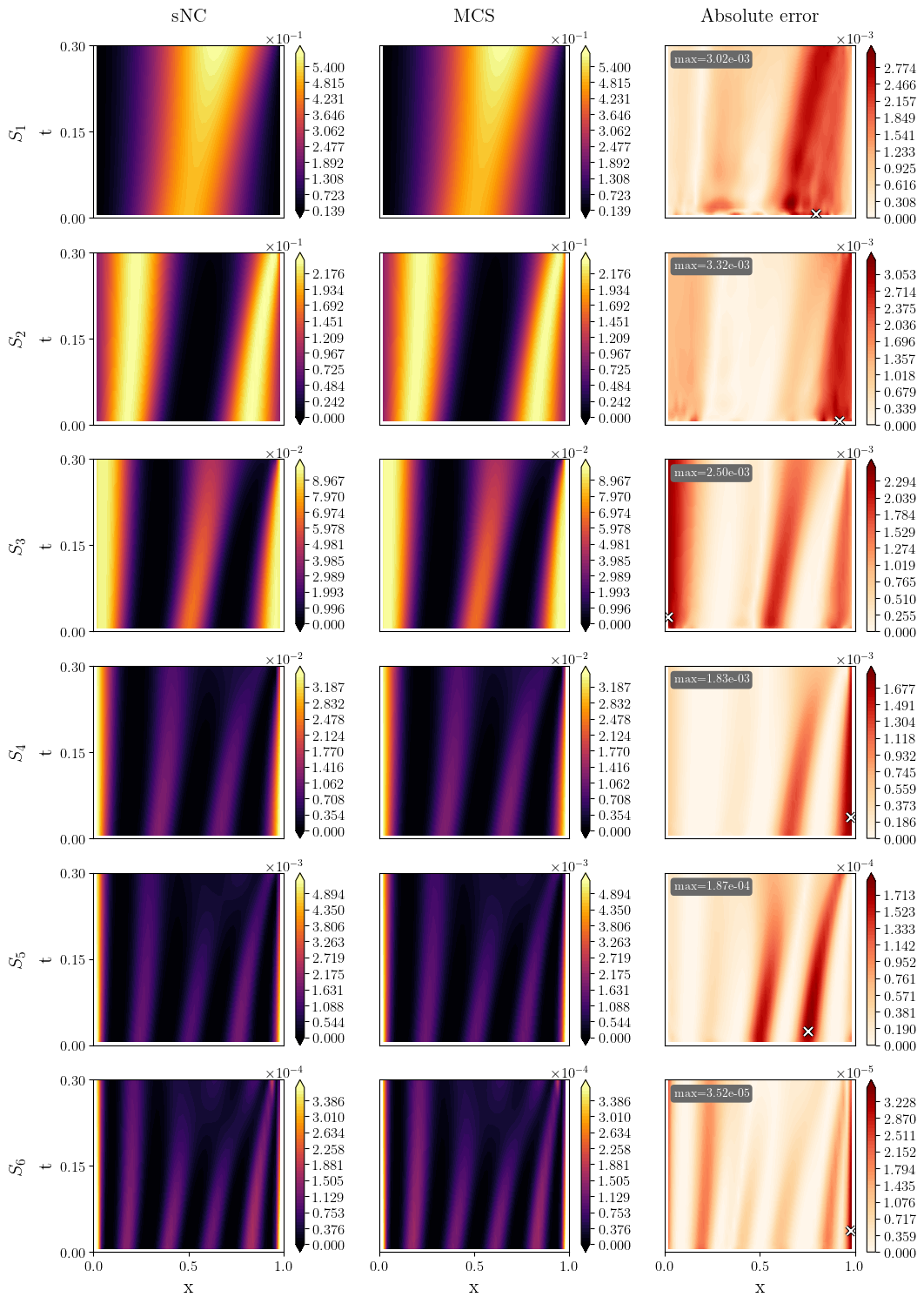}\\
  \vspace{0.2em}
  \caption{One-dimensional Burgers’ equation. First-order Sobol' indices computed directly from the coefficient functions of the sNC expansion, compared with reference values estimated via MCS using $2^{18}$ realizations. The corresponding absolute error fields are also shown, with the maximum absolute error across the spatial domain highlighted in each case.}
  \label{fig:1D6RV_burgers_GSA_first}
\end{figure}

The first-order Sobol' indices computed directly from the coefficient functions of the sNC expansion are shown in Fig.~\ref{fig:1D6RV_burgers_GSA_first}, together with the corresponding MCS estimates and absolute error fields. Excellent agreement between the sNC and MCS estimates is observed across all first-order indices; the proposed framework accurately recovers both the magnitude and spatial-temporal structure of the dominant sensitivity contributions. In comparison with the previous advection-diffusion example, the sensitivity fields exhibit substantially sharper and more localized transport-dominated structures, reflecting the stronger nonlinear dynamics induced by the viscous Burgers system and squared stochastic forcing. The corresponding absolute error fields remain uniformly small throughout the spatial-temporal domain, with the largest discrepancies localized near the sharply varying sensitivity features associated with the nonlinear transport dynamics.

\begin{figure}
  \center
  \includegraphics[width=4.5in]{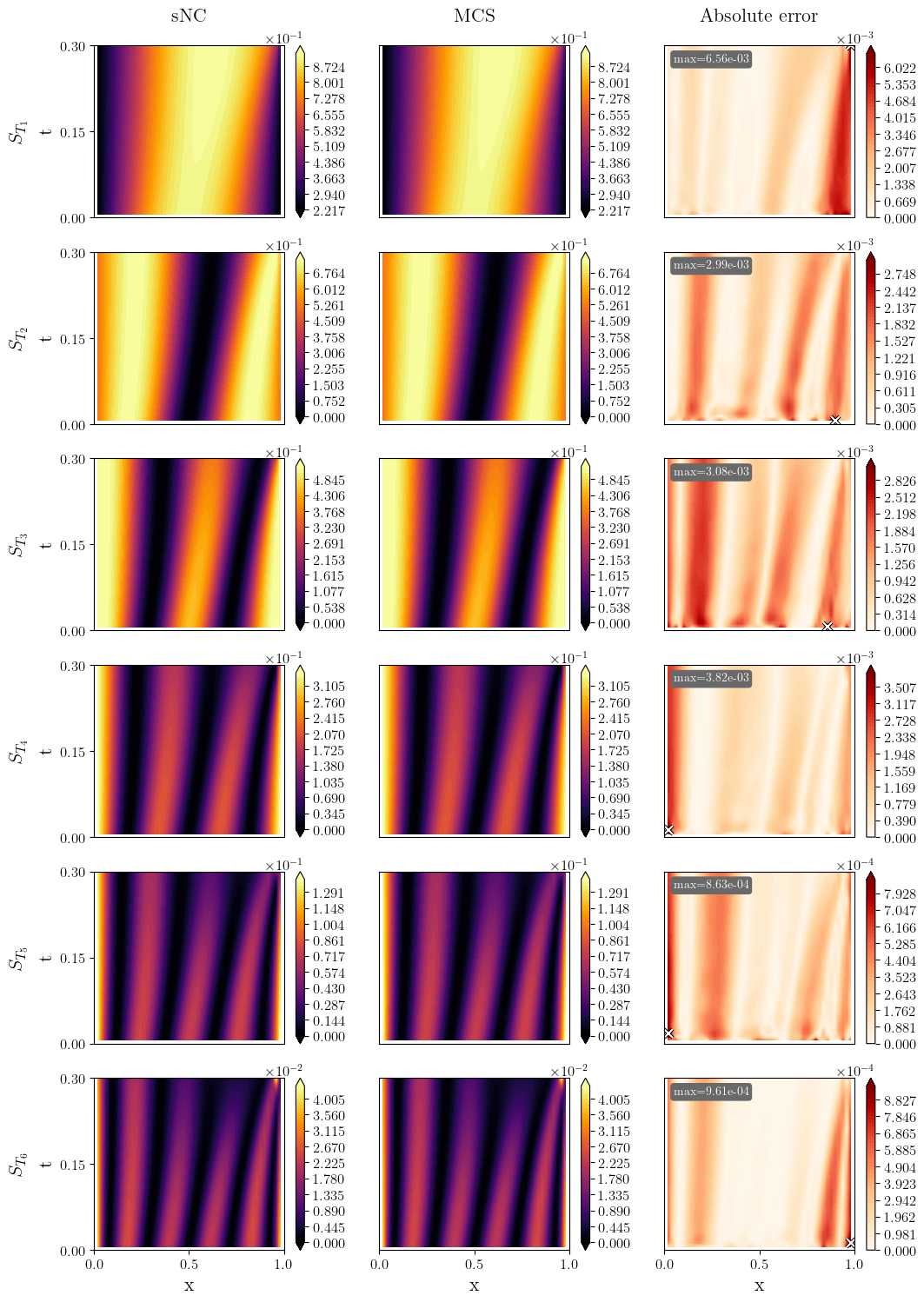}\\
  \vspace{0.2em}
  \caption{One-dimensional Burgers’ equation. Total-order Sobol' indices computed directly from the coefficient functions of the sNC expansion, compared with reference values estimated via MCS using $2^{18}$ realizations. The corresponding absolute error fields are also shown, with the maximum absolute error across the spatial domain highlighted in each case.}
  \label{fig:1D6RV_burgers_GSA_total}
\end{figure}

The total-order Sobol' indices computed directly from the coefficients of the sNC expansion are shown in Fig.~\ref{fig:1D6RV_burgers_GSA_total}; the figures also show the reference MCS estimates and the corresponding absolute errors. Excellent agreement between the sNC and MCS estimates is again observed across all variables. In contrast to the previous advection-diffusion example, the total-order indices are substantially larger than their corresponding first-order indices, clearly indicating the presence of strong higher-order interaction effects throughout the stochastic system. Nevertheless, the proposed framework accurately captures both the magnitude and spatial-temporal structure of the total-order sensitivity fields, with the corresponding absolute error fields remaining uniformly small despite the increased interaction complexity of the problem.

\subsection{Comparison with polynomial chaos expansion}
\label{subsec:pce_comparison}

Classical PCE can be used in a non-intrusive manner to estimate statistical and sensitivity quantities. For functional stochastic responses of the form $u(\bm{x}, \bm{\xi})$, one possible approach is to construct an independent PCE surrogate at each deterministic grid location. However, this strategy can become computationally expensive due to the need to build and store a large number of independent expansions, while also restricting the resulting statistical and sensitivity estimates to the underlying computational grid. 

To enable a more direct comparison with the proposed sNC framework, we instead adopt the strategy used in \cite{novak2024physics, sharma2024physics}, where the deterministic variables are treated as additional random variables with uniform distribution. Following the classical PCE formulation described in Section~\ref{subsec:PCE}, multivariate orthonormal basis functions are constructed through tensor products of univariate orthonormal polynomial bases. The coefficients in Equation~\eqref{eq:PCE} are then determined using least-squares regression on the training data. In this formulation, similarly to the proposed sNC framework, the resulting PCE surrogate provides statistical quantities and Sobol' indices as continuous functions of the deterministic variables rather than only at fixed grid locations.

We consider the one-dimensional Euler-Bernoulli beam problem introduced in Section~\ref{subsec:1D7RV}, using the same training dataset (4096 stochastic samples) to ensure a fair comparison between the proposed sNC framework and the PCE-based approach. Since the functional PCE formulation treats the deterministic coordinate as an additional random variable, the resulting approximation corresponds to an eight-dimensional PCE problem. Consequently, the number of terms in the PCE expansions with $p=5,6$, and $7$ are $1287, 3003$, and $6435$, respectively. For $p\geq7$, the number of coefficients exceeds the training size, which may lead to ill-conditioning in the associated regression problem. Nevertheless, results for $p=7$ are included in the present study since increasing the polynomial degree to $p=7$ still improves the approximation accuracy of the statistical and sensitivity quantities, whereas higher polynomial degrees produce unstable estimates. 

\begin{figure}
  \center
  \includegraphics[width=\textwidth]{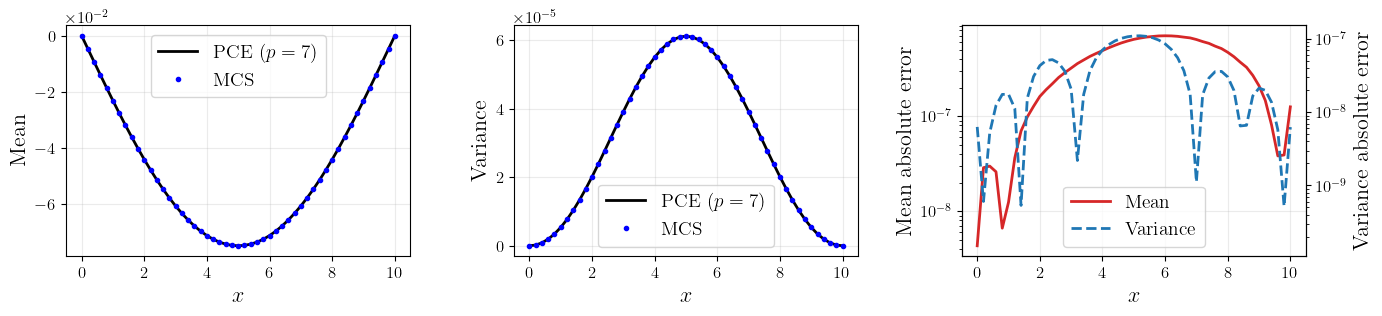} \\
  \makebox[\linewidth][l]{\hspace{1.1in} \small (a) \hspace{1.7in} (b) \hspace{1.7in} (c)}\\
  \vspace{0.2em}
  \caption{One-dimensional Euler-Bernoulli beam. (a)~Mean and (b)~variance estimates computed directly from the functional PCE coefficients with maximum total polynomial degree $p=7$, compared with reference values estimated via MCS using $2^{18}$ realizations. (c)~Absolute errors in the mean and variance.}
  \label{fig:1D7RV_mean_var_pce}
\end{figure}

Fig.~\ref{fig:1D7RV_mean_var_pce} compares the mean and variance computed directly from the coefficients of the functional PCE with maximum total polynomial degree $p=7$ against reference values estimated via MCS with $2^18$ realizations; the corresponding absolute errors across the spatial domain are also shown. Results for $p=7$ are presented since this choice provides the most accurate estimates among the considered polynomial degrees $p=1,\dots,7$. Overall, the PCE estimates show close agreement with the MCS results, with small absolute errors observed throughout the domain for both the mean and variance. Slightly smaller errors are observed for the mean estimates obtained with the PCE approach compared to the sNC results, whereas the sNC framework yields slightly smaller errors for the variance estimates (see Fig.~\ref{fig:1D7RV_mean_var}). Nevertheless, both approaches provide highly accurate approximations of the statistical quantities, and for this problem the functional PCE formulation does not exhibit significant deficiencies in the estimation of the mean and variance.

\begin{figure}
  \center
  \includegraphics[width=\textwidth]{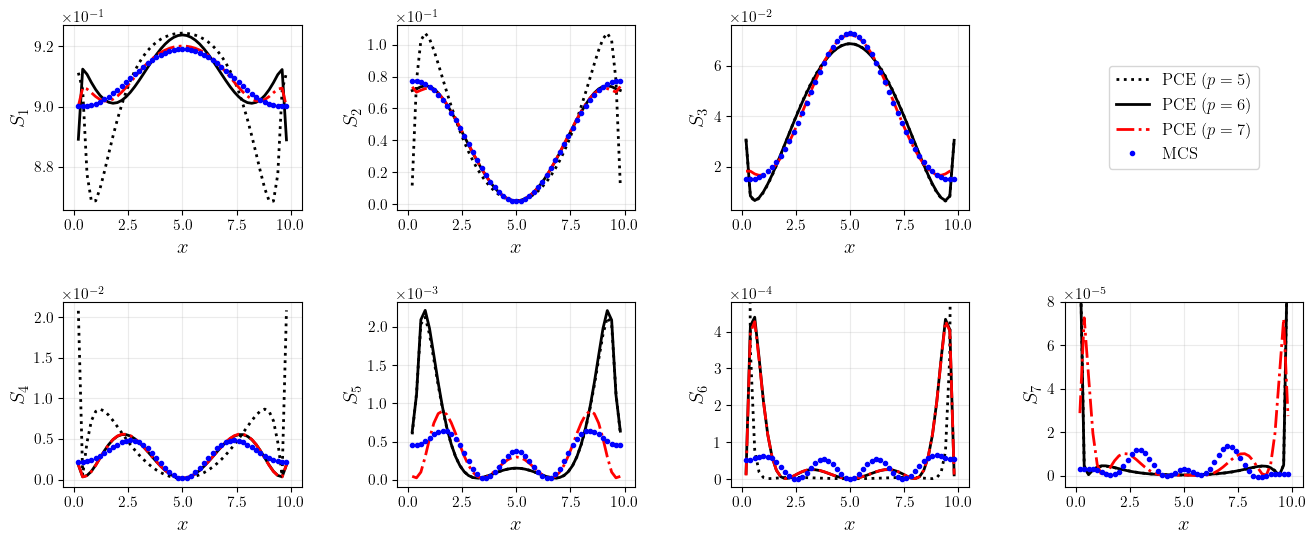}\\
  \makebox[\linewidth][c]{\small (a)} \\
  \vspace{1em}
  \includegraphics[width=\textwidth]{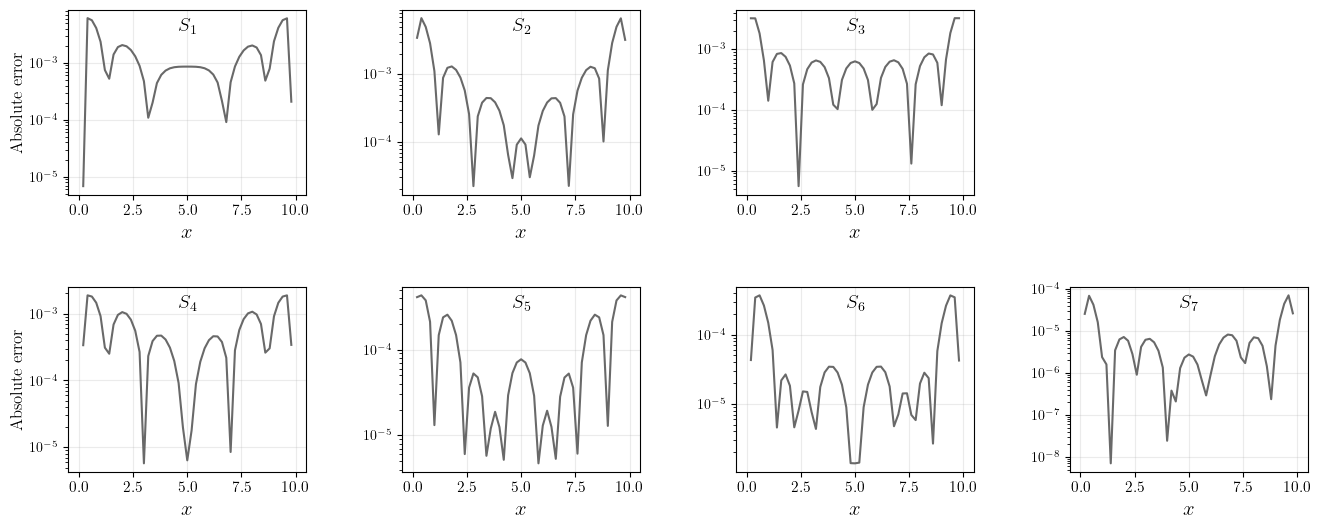}
  \makebox[\linewidth][c]{\small (b)} \\
  \vspace{0.2em}
  \caption{One-dimensional Euler-Bernoulli beam. (a)~First-order Sobol' indices computed directly from the functional PCE coefficients for different values of the maximum total polynomial degree $p$, compared with reference values estimated via MCS using $2^{18}$ realizations. (b)~Absolute error of the PCE ($p=7$) Sobol' index estimates relative to the MCS estimates. Values at locations where the total variance vanishes are omitted, as the Sobol' indices are undefined there.}
  \label{fig:1D7RV_GSA_first_pce}
\end{figure}

The first-order Sobol' indices computed directly from the coefficients of the functional PCE for maximum total polynomial degrees $p=5,6$ and $7$ are shown in Fig.~\ref{fig:1D7RV_GSA_first_pce}(a), together with the reference MCS estimates. The panels corresponding to $S_6$ and $S_7$ are slightly zoomed in to improve visibility of the results. The corresponding absolute errors for the $p=7$ approximation, which provides the most accurate estimates among the considered polynomial degrees, are shown in Fig.~\ref{fig:1D7RV_GSA_first_pce}(b). As expected, the Sobol' index estimates improve as the maximum total polynomial degree $p$ increases. Overall, the PCE estimates capture the general trends and orders of magnitude of the reference Sobol' estimates; however, noticeable discrepancies remain in several of the sensitivity estimates, even for $p=7$. These discrepancies are particularly evident near the domain boundaries, where the variance becomes small and the Sobol' indices become increasingly sensitive to approximation errors.

While the estimates improve as $p$ is increased, the value of p remains restricted by the amount of training data available. Further improvements through larger polynomial degrees would require additional training data, thereby increasing the computational cost. Although similar localized discrepancies may initially arise in NN-based approximations, the flexibility of the NN parameterization in the proposed sNC framework enables these regions to be progressively refined through continued training and increased network expressivity; the NN parameterization provides greater flexibility and adaptive local approximation capability than fixed global polynomial bases. In contrast, the functional PCE results exhibit more persistent discrepancies near the boundaries even as the polynomial degree is increased. As demonstrated by the sNC results in Fig.~\ref{fig:1D7RV_GSA_first}, the proposed framework achieves substantially closer agreement with the reference MCS sensitivity estimates across the spatial domain.
\section{Conclusion}
\label{sec:conclusion}

In this work, we propose the Structured Neural Chaos (sNC) as a non-intrusive surrogate modeling framework for constructing scalable and interpretable approximations of high-dimensional functional stochastic systems. The proposed framework combines a truncated functional ANOVA decomposition with separable low-rank NN parameterizations to retain the interpretability of structured decompositions while leveraging the expressive power of NNs. This structure enables the extraction of statistical quantities and Sobol' sensitivity indices directly from the learned coefficients of the sNC expansion, mirroring several attractive features of PCE while additionally representing these quantities as functions over the underlying physical domain.

The proposed framework accurately recovers the mean, variance, and Sobol' sensitivity fields across benchmark problems of increasing stochastic and physical complexity. Accurate approximations can be obtained using relatively modest stochastic training samples, highlighting the potential of the proposed framework in settings where high-fidelity model evaluations are limited. The proposed framework provides accurate uncertainty quantification and sensitivity analysis at substantially lower computational cost relative to MCS. In addition, the results indicate robustness with respect to random initialization and optimization variability, suggesting that the learned representations are stable across independent runs. Furthermore, the numerical examples demonstrate the ability of the method to adaptively capture the dominant structure of the underlying stochastic system, increasing the complexity of the expansion as required by the underlying problem.

The comparison with functional PCE highlights the benefits of the NN parameterizations employed in sNC, which provide a flexible alternative to polynomial approximations. The proposed framework enables the accurate recovery of Sobol' sensitivity fields, particularly in localized low-variance regions where sensitivity estimation becomes increasingly challenging. In this sense, the proposed framework provides a favorable alternative to functional PCE by combining interpretable low-rank functional decompositions with the adaptive approximation capabilities of NNs. 

Several extensions of the proposed framework warrant further investigation. Adaptive sampling strategies could further improve efficiency by selectively allocating high-fidelity model evaluations to regions of the stochastic space associated with the greatest uncertainty. Extensions to higher-order interaction representations may prove beneficial for problems in which higher-order stochastic interactions play a significant role, such as systems exhibiting strongly coupled responses or rare-event behavior. Additional directions include the investigation of adaptive network architectures, applications to higher-dimensional physical domains, and theoretical analyses of approximation and convergence properties.
\section*{Acknowledgments}

This research was performed under an appointment to the NNSA-IMPACT internship program administered by the Oak Ridge Institute for Science and Education (ORISE) for the National Nuclear Security Administration (NNSA) and the U.S. Department of Energy (DOE). This work was performed under the auspices of the U.S. Department of Energy by Lawrence Livermore National Laboratory under Contract DE-AC52-07NA27344. This work is reviewed and released under LLNL-JRNL-2021069.

\appendix

\section{Ablation study: data dependency}

The quality of the statistical and sensitivity estimates produced by the proposed framework depends on both the quality and quantity of the available training data. This dependence arises from the non-intrusive and fully data-driven nature of the approach, particularly since the sequential orthogonal projection procedure requires sufficiently informative data to accurately identify the corresponding interaction subspaces and avoid projection leakage between ANOVA components. 

To investigate the effect of training data size on the resulting surrogate and sensitivity estimates, we consider the one-dimensional Euler-Bernoulli beam problem introduced in Section~\ref{subsec:1D7RV}. The dependence of the proposed framework on the amount of available training data is investigated using stochastic training datasets consisting of $N=2^8, 2^9,\dots, 2^{13}$ realizations of the random field with the same uniform grid of 51 equidistant points over the spatial domain $x\in[0,10]$. For each training size, the relative $L^2$ errors associated with the surrogate response and the estimates of the mean, variance, first-order indices, and total-order indices are evaluated relative to the reference MCS estimates computed using $2^{18}$ realizations. To assess the robustness of the proposed framework with respect to random initialization and optimization variability, each experiment is repeated over 10 independent runs and the corresponding mean and standard deviation of the resulting errors are reported. 

\begin{figure}
  \center
  \includegraphics[width=\textwidth]{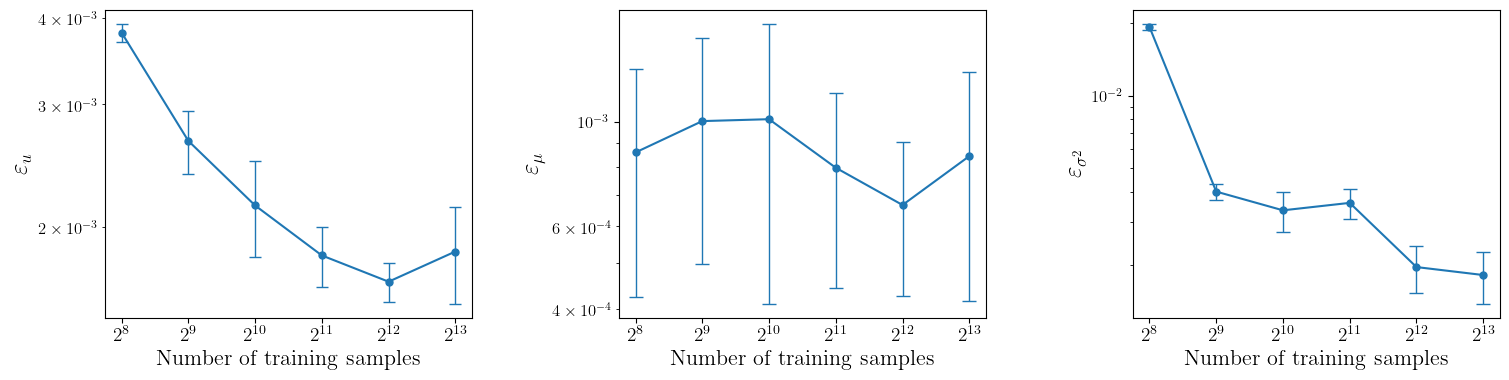}\\
  \makebox[\linewidth][l]{\hspace{1in} \small (a) \hspace{1.95in} (b) \hspace{1.95in} (c)} \\
  \vspace{1em}
  \includegraphics[width=\textwidth]{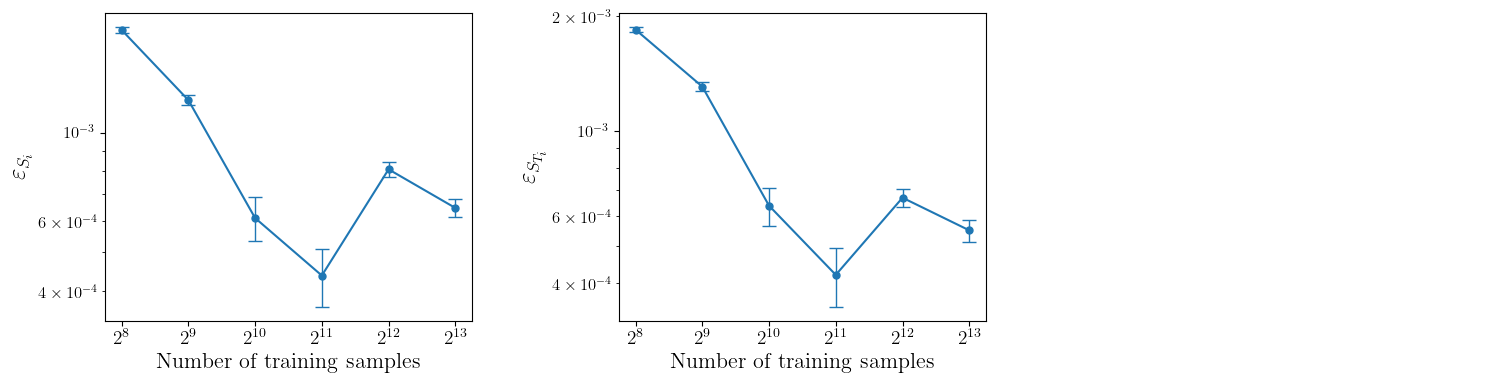}
  \makebox[\linewidth][l]{\hspace{1in} \small (d) \hspace{1.95in} (e)} \\
  \vspace{0.2em}
  \caption{One-dimensional Euler-Bernoulli beam. Dependence of the proposed sNC framework on the size of the stochastic training dataset. Relative $L^2$ errors for the (a)~surrogate response, (b)~mean, (c)~variance, (d)~first-order Sobol' indices, and (e)~total-order Sobol' indices as functions of the number of stochastic training samples. Results are averaged over 10 independent runs, with error bars indicating one standard deviation. Reference quantities are estimated using MCS with $2^{18}$ realizations. }
  \label{fig:1D7RV_ablation}
\end{figure}

Fig.~\ref{fig:1D7RV_ablation} shows the dependence of the proposed sNC framework on the size of the stochastic training dataset. Fig.~\ref{fig:1D7RV_ablation}(a) shows the relative $L^2$ error of the surrogate predictions, denoted by $\varepsilon_u$ defined in Eq.~\eqref{eq:varepsilon_u}. The error decreases as the number of training samples increases and then begins to saturate for larger training datasets. Moreover, the differences in the errors are relatively small, suggesting that the surrogate representation is capable of achieving high accuracy even with a moderate number of training samples.

The relative $L^2$ error of the mean and variance estimated directly from the sNC expansion coefficients are shown in Fig.~\ref{fig:1D7RV_ablation}(b) and Fig.~\ref{fig:1D7RV_ablation}(c), respectively. The error associated with the mean estimates is already very small, even for the smallest training datasets considered, and appears to saturate rapidly, indicating a comparatively weak dependence on the training data size. These results suggest that the mean estimation problem is relatively stable and easier to approximate than the remaining quantities considered. While the variability between runs is larger for the mean estimates compared to the other quantities, the overall magnitude of the variability remains small. The variance estimates, however, show a more pronounced improvement as the number of training samples increases before gradually stabilizing for larger datasets. In contrast to the mean estimates, the variability between independent runs remains consistently small, indicating limited sensitivity of the variance estimates to random initialization and optimization effects.

Fig.~\ref{fig:1D7RV_ablation}(d) and Fig.~\ref{fig:1D7RV_ablation}(e) show the relative $L^2$ errors of the estimated first-order and total-order Sobol' indices, respectively. The errors decrease substantially as the number of training samples increases up to approximately $N=2^{11}$, after which mild non-monotonic fluctuations are observed. Since these fluctuations occur after the errors have already become small, they are likely associated with optimization variability arising from random initialization and stochastic training effects rather than insufficient training data. Nevertheless, the overall variability between independent runs remains small, indicating that the sensitivity estimates are reasonably robust with respect to random initialization. Compared to the mean and variance estimates, the Sobol' index errors exhibit a stronger dependence on the amount of training data available. This behavior is expected, since the sensitivity indices depend on accurate variance decomposition and are therefore more demanding quantities to estimate accurately.

Overall, the results demonstrate that the proposed sNC framework is capable of producing accurate surrogate, statistical, and sensitivity estimates even when trained using relatively small stochastic datasets. While increasing the number of training samples generally improves the resulting approximations, the observed saturation of the errors and the consistently small variability between independent runs indicate that the proposed framework remains stable and data-efficient for the considered problem.


\section{Neural Network Architecture and Training}
\label{sec:Appendix_NN_hyperparameters}

Across all numerical examples, the basis NNs employed in the proposed sNC framework share a common architecture consisting of three hidden layers of width 32 with SiLU activation functions. The learning rate is fixed to $10^{-3}$, while the batch size is set to 64. Each basis network is trained for 300 epochs in all examples, which was found sufficient for the training loss to reach a plateau. In all numerical examples, the AdamW optimizer from the PyTorch package is employed.

The coefficient NNs were parameterized using a single multi-output multilayer perceptron (MLP). In contrast to the basis networks, the coefficient networks are tuned separately through a two-step hyperparameter optimization procedure targeting accurate approximation of the Sobol' sensitivity indices. Across all problems, the batch size is fixed to either 64 or the full dataset size, whichever is smaller. During the first stage, the learning rate, number of hidden layers, network width, and activation function are optimized while the sparsity parameter is fixed at $\tau=-\infty$ (i.e., no sparsity is introduced) and only rank-1 approximations are considered. In the second stage, the NN hyperparameters are fixed and the sparsity parameter $\tau$ is optimized.

Since accurate approximation of Sobol' indices is a primary objective of the proposed framework, particular emphasis is placed on accurately capturing localized low-variance regions, which strongly influence the sensitivity estimates. Consequently, the coefficient networks are trained for a large number of epochs (up to 30,000 epochs). During training, first-order and total-order Sobol' indices are computed pointwise from the discrete coefficient data generated through the orthogonal projection-based ALS step and compared against the corresponding sensitivity estimates obtained from the NN coefficient approximations. To quantify this discrepancy, the maximum absolute pointwise Sobol' error is defined as
\begin{align}
    \varepsilon_{\max} =\max\left\{\max_{i=1,\dots,d}\|S_i^{\text{ALS}}-\hat{S}_i\|_\infty, \quad \max_{i=1,\dots,d}\|S_{T_i}^{\text{ALS}}-\hat{S}_{T_i}\|_\infty\right\}
\end{align}
where $S_i^{\text{ALS}}$ and $S_{T_i}^{\text{ALS}}$ denote the first-order and total-order Sobol' indices computed from the discrete coefficient data, respectively, and $\hat{S}_i$ and $\hat{S}_{T_i}$ denote the corresponding NN approximations. During training, $\epsilon_{\max}$ is periodically evaluated, and the final network parameters are selected from the epoch attaining the minimum recorded value, 
\begin{align}
    \epsilon^*_{\max}=\min_k \epsilon^{(k)}_{\max
    }
\end{align}
where $\epsilon^{(k)}_{\max}$ denotes the value of $\epsilon_{\max}$ at epoch $k$. For the first stage of the hyperparameter optimization procedure, the configuration producing the overall smallest value of $\epsilon^*_{\max}$ across all considered architectures and training settings is selected. When multiple hyperparameter configurations produce comparable values of $\varepsilon^*_{\max}$, preference is given to simpler architectures. 

Since this criterion emphasizes the largest local discrepancies, the optimization procedure naturally biases the training toward accurately resolving regions associated with larger Sobol' indices. This behavior is desirable in practice, as the largest sensitivity values correspond to the most influential variables and interaction structures governing the stochastic response. This procedure was found to improve both the stability and accuracy of the approximations in localized low-variance regions. Once the optimal NN hyperparameters have been identified, they are held fixed and the sparsity parameter $\tau$ is optimized by minimizing the validation MSE over a set of 300 stochastic samples. When multiple values of $\tau$ yield comparable errors, preference is given to sparser representations.

\begin{table}
\centering
\caption{Coefficient-network hyper-parameters used across the numerical examples.}
\label{tab:coeff_hyperparams}
\begin{tabular}{l ccccc}
\toprule
Problem & Hidden layers & Width  & Learning rate & Batch size & $\tau$  \\
\midrule
1D nonlinear ODE & 2 & 32  & $10^{-3}$ & 20 & $10^{-2}$  \\

1D Euler-Bernoulli beam &  3 & 32 & $10^{-3}$ & 51 & $10^{-2}$ \\

2D nonlinear elliptic equation &  4 & 512 & $10^{-3}$  & 64 & $10^{-3}$ \\

1D advection--diffusion equation &  4 & 512 & $10^{-4}$  & 64  & $10^{-2}$\\

1D Burgers' equation & 4 & 512 &  $10^{-4}$ & 64 & $10^{-2}$ \\
\bottomrule
\end{tabular}
\end{table}

Although several activation functions were considered during hyperparameter optimization, GELU activations were selected for all problems. The remaining hyperparameters used in the coefficient-network across all five numerical examples are shown in Table~\ref{tab:coeff_hyperparams}. The selected hyperparameters reflect the increasing complexity of the underlying stochastic systems across the numerical examples. Simpler low-dimensional problems were accurately approximated using relatively shallow and narrow coefficient networks, while the higher-dimensional examples required substantially wider architectures to accurately resolve localized low-variance regions and the associated Sobol' sensitivity fields. Among the considered hyperparameters, the network width was observed to have the strongest influence on accurately capturing these localized sensitivity structures.

\bibliographystyle{unsrt}
\bibliography{refs}

\end{document}